\documentclass[11pt,letterpaper]{article}

\usepackage[margin=1in]{geometry}
\usepackage[utf8]{inputenc}
\usepackage[T1]{fontenc}
\usepackage{lmodern}
\usepackage{amsmath,amssymb,amsfonts, bbm}
\usepackage{amsthm}
\usepackage{booktabs}
\usepackage{array}
\usepackage[table]{xcolor}
\usepackage{longtable}
\usepackage{graphicx}
\usepackage{float}
\usepackage{enumitem}
\usepackage{url}
\usepackage{microtype}
\usepackage{placeins}

\usepackage[authoryear,round]{natbib}
\usepackage{authblk}
\usepackage[hidelinks]{hyperref}

\hypersetup{
  pdftitle={SOTA or Luck? The Winner's Curse in LLM Leaderboards},
  pdfauthor={Sikun Xu, Raphael Thomadsen, Dennis J. Zhang, Heng Zhang}
}

\graphicspath{{figures/}}
\newcommand{\floatnote}[1]{\par\vspace{4pt}\noindent\begin{minipage}{\linewidth}\footnotesize\raggedright\textit{Notes.} #1\end{minipage}\par}

\newcommand{\E}{\mathbb{E}}
\newcommand{\R}{\mathbb{R}}

\theoremstyle{definition}
\newtheorem{assumption}{Assumption}[section]
\theoremstyle{plain}
\newtheorem{lemma}[assumption]{Lemma}
\newtheorem{proposition}[assumption]{Proposition}
\newtheorem{theorem}[assumption]{Theorem}

\theoremstyle{remark}

\title{SOTA or Luck? \\ The Winner’s Curse in LLM Leaderboards}

\author[1]{Sikun Xu}
\author[1]{Raphael Thomadsen}
\author[1]{Dennis J.~Zhang}
\author[2]{Heng Zhang}
\affil[1]{Olin Business School, Washington University in St.~Louis\protect\\
  St.~Louis, MO 63130\protect\\
  \texttt{sikun@wustl.edu}, \texttt{thomadsen@wustl.edu}, \texttt{denniszhang@wustl.edu}}
\affil[2]{W.~P.~Carey School of Business, Arizona State University\protect\\
  Tempe, AZ 85287\protect\\
  \texttt{hzhan388@asu.edu}}
\date{}

\begin{document}
\maketitle

\begin{abstract}
Public LLM leaderboards guide model evaluation and adoption, yet the reported performance of top-ranked models can be systematically overstated. This leaderboard winner's curse arises because ranking noisy benchmark scores rewards sampling luck alongside genuine ability. The same selection effect can distort scores throughout the leaderboard, a phenomenon we call the \textit{rank-induced bias}. We develop an $m$-out-of-$N$ bootstrap to correct for this bias. It assumes a fixed model pool with independent, identically distributed bounded tasks while allowing arbitrary within-task correlation among models. Under exact population tiers separated by positive gaps, we prove that the bootstrap reduces $\Theta(N^{-1/2})$ ranking bias to expected residual bias $o(N^{-1/2})$. Across 14 benchmarks from 5 benchmark families, the rank-1 adjustments range from approximately 0.09 to 3.86 percentage points. In our SWE-bench Verified snapshot, the tied leaders receive an adjustment of 0.90 percentage points. Such overstatement is large enough to make the apparent winner statistically fragile. Thus, we use the same bootstrap to further quantify the intervals of plausible rank positions for each model and group models into tiers. We establish sufficient conditions for consistent recovery of population tiers. On SWE-bench Verified, this procedure places 13 of 173 models in the leading tier. We combine the score adjustment and rank intervals in the Rank Uncertainty and Leaderboard Evaluation Report (RULER). We validate its finite-sample performance using sample splitting and calibrated simulations, and advocate publishing RULER alongside raw rankings so readers can assess both reported performance and the precision of the ordering. RULER also provides guidance for benchmark design that accounts for selection effects and ranking uncertainty.
\end{abstract}

\section{Introduction}
\label{sec:intro}

Public leaderboards are central to the evaluation of large language models. Organizations use benchmark rankings to choose models, researchers to claim state-of-the-art performance, and funding bodies to assess progress. Yet benchmark scores are no more than sample averages over a small set of tasks, and sample averages inherit the randomness of the underlying data. Thus, selecting the model with the highest observed score \textit{rewards sampling luck as well as genuine ability}. As a result, the true performance of the winning model tends to be overestimated. This overestimation is the Winner's Curse (WC), a phenomenon well understood in auction theory \citep{capen1971competitive}, genome-wide association studies \citep{zollner2007overcoming}, and clinical trials \citep{zhong2008bias}, but largely unexamined in LLM evaluation. 
The consequences extend beyond statistical reporting: leaderboard rankings influence model procurement, deployment spending, and research investment, yet these decisions often rely on point rankings without accounting for the uncertainty behind small score differences. As a result, economic resources can be directed toward apparent winners whose performance advantages are not reliably established.

We also stress that LLM leaderboards can inherit the WC dynamics at a much larger scale: the number of competing models has exploded, capability gaps at the top have narrowed, and new State-Of-The-Art (SOTA) claims arrive weekly based on fractional-percentage-point improvements. These are precisely the conditions under which the WC is at its worst \citep{xu2025winners}. Additionally, the same selection effect extends throughout a leaderboard: among models with the same mean performance, scores near the top tend to be inflated and those near the bottom deflated. We call this \emph{rank-induced bias}. We develop an $m$-out-of-$N$ bootstrap \citep{bickel1997resampling, xu2025winners} to (1) estimate and correct this bias and (2) quantify ranking uncertainty by resampling tasks and re-ranking models altogether. We package these results in a \textit{Rank Uncertainty and Leaderboard Evaluation Report (RULER)}: score adjustments for selection, a \emph{rank probability matrix} (RPM) showing how often each model occupies each rank, and rank intervals used to group models into tiers. We advocate publishing RULER alongside raw rankings so readers can assess both the reported performance and the precision of model ranking.

Existing work on leaderboard reliability does not correct rank-induced bias. The literature has studied adaptive test-set reuse \citep{blum2015ladder, dwork2015generalization, roelofs2019meta}, benchmark lottery and prompt sensitivity \citep{dehghani2021benchmark, alzahrani2024benchmarks}, run-to-run variance \citep{bouthillier2021accounting}, item-level and pairwise uncertainty \citep{polo2024efficient, ye2024benchmarking, chiang2024chatbot}, and data contamination \citep{deng2024investigating}. There is a separate research direction concerning the vendors tuning models to public benchmarks, which has motivated private evaluation suites \citep{liang2022holistic}. Because rank-induced bias comes from sampling noise, it persists on private benchmarks too. Our work also relates to the post-selection inference literature. Empirical Bayes methods shrink estimates toward a common prior, often under parametric assumptions \citep{efron2011tweedie, gu2023invidious}. Selective inference conditions on the selection event and requires a separate derivation for each selection procedure \citep{lee2016exact, andrews2024inference}. The $m$-out-of-$N$ bootstrap \citep{bickel1997resampling} needs neither. It requires only i.i.d. bounded tasks, allows arbitrary correlation among models within a task, and yields the score correction and the RPM from a single set of resamples. \citet{xu2025winners} use it to correct the score of a single selected winner with independent alternatives in inference-then-optimize settings. Appendix~\ref{app:related} reviews further related work.

\begin{figure}[!t]
  \centering
  \includegraphics[width=\textwidth]{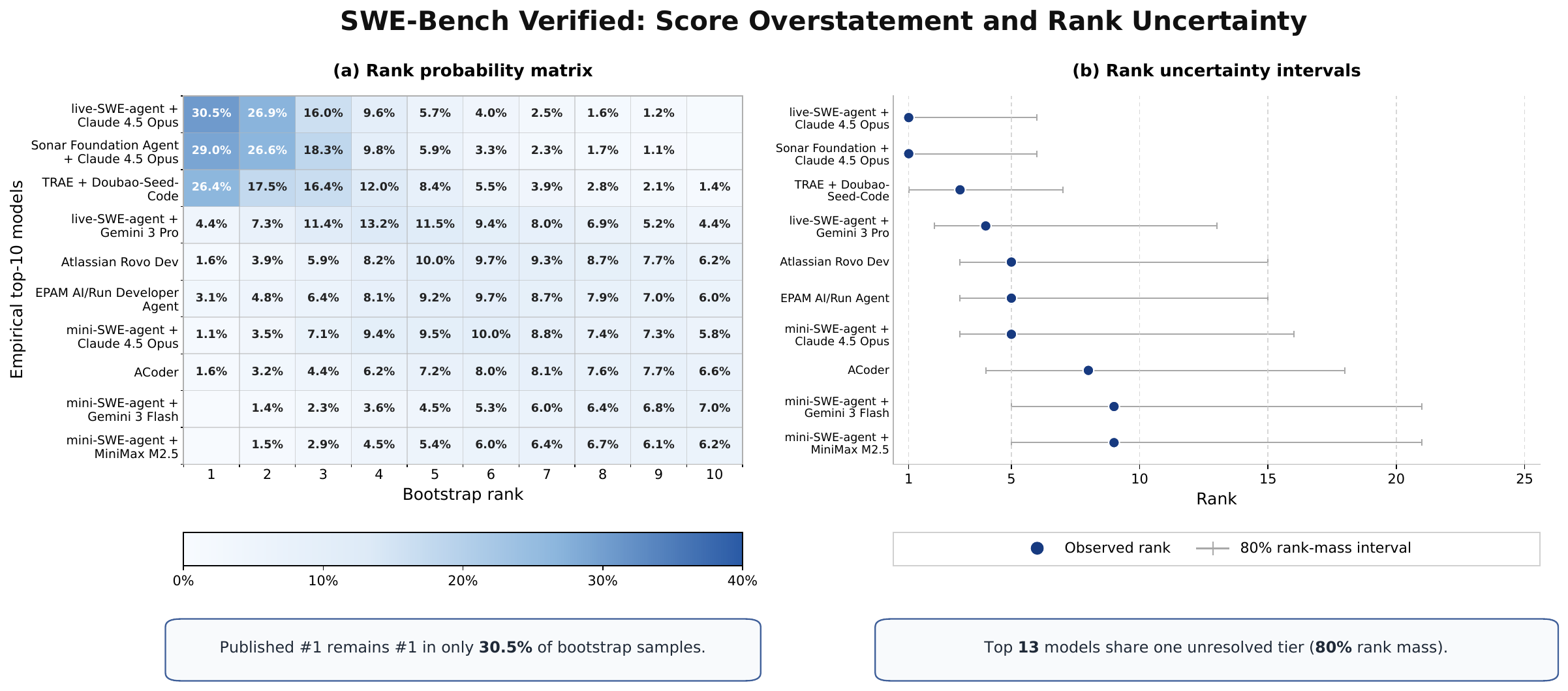}
  \caption{The winner on SWE-bench Verified remains statistically unresolved.}
  \label{fig:swebench_verified_unresolved}

\floatnote{The snapshot contains 173 models and 500 tasks, with $m=\lfloor N^{0.9}\rfloor$ and $B=5{,}000$ bootstrap resamples. Panel (a) shows the top-10 RPM, whose entries are bootstrap rank probabilities. Panel (b) shows 80\% rank-mass intervals, with blue points marking observed ranks. Tiers use the common-intersection rule. Appendix~\ref{app:tables} shows the leading $30\times30$ RPM.}
\end{figure}

We apply RULER to SWE-Bench Verified, which evaluates 173 models on 500 software engineering tasks at the time of our snapshot, the 2 tied leaders score 79.2\%. RULER estimates that the score is inflated by 0.9 percentage points, giving adjusted scores of 78.3\%. This inflation exceeds the typical margin by which a new entry takes the lead. In 2025, the top score on SWE-bench Verified rose 13 times. The median increase was 0.8 percentage points, and 7 of the 13 increases were smaller than the estimated inflation. RULER also reports a significant ranking uncertainty, as shown in Figure~\ref{fig:swebench_verified_unresolved}. Panel (a) shows that the 2 tied leaders receive only 30.5\% and 29.0\% probability of remaining as leaders, while the model at rank 3 receives 26.4\%. In addition, panel (b) shows that 80\% rank intervals for the 2 leaders both span ranks 1--6, and 13 models have overlapping rank intervals, suggesting an indistinguishable top tier. In addition to SWE-Bench Verified, we also apply RULER to 14 benchmarks from 5 families (SWE-bench, Tau2-Bench, LiveBench, AlpacaEval~2, and WildBench), spanning software engineering, agentic customer service, instruction following, and judged response quality. The same patterns frequently recur: rank-1 adjustments exceed 1 percentage point in 7 benchmarks, and the leading tier contains multiple models in 12 benchmarks out of a total of 14 (Table~\ref{tab:cross_variant_summary}). Because organizations, researchers, and funding bodies act on these rankings, overstated winning scores and unresolved orderings can misdirect model adoption, research credit, and investment.

We validate RULER using sample splitting and a calibrated simulation. In the split study, one half of each task pool selects the ranking and the other half evaluates the selected models. We average the resulting score errors over 200 splits within each benchmark and then average their magnitudes across benchmarks. At rank~1, adjustment reduces score errors from 2.38 to 0.73 pp. A simulation calibrated to SWE-bench Verified supplies a second check with known model means: the correction reduces mean winner bias from 1.41 to 0.21 pp (Section~\ref{sec:verification}).

\textbf{Contributions.}
Our paper puts forth a conceptual shift: While people typically treat the numbers on LLM leaderboards as collections of model means that truthfully reflect the models' performance, they are in fact post-selection objects because of ranking. A model might be ranked at the top due to noise in the data, inflating the performance of the top model(s). We correct for this rank-induced inflation and quantify rank uncertainty. Specifically, we make the following contributions.

\begin{itemize}[leftmargin=*,nosep]
  \item \textbf{Full-leaderboard bootstrap inference.} We apply an $m$-out-of-$N$ rank-induced bias correction across the entire leaderboard, allowing arbitrary model correlation within each task. RULER combines position-level score adjustments, RPMs that report how often each model occupies each rank under task resampling, and rank-mass intervals that group models into tiers. Under exact population tiers separated by positive gaps, we derive an all-position bias expansion and prove that the bootstrap reduces $\Theta(N^{-1/2})$ rank-induced bias to an expected residual of $o(N^{-1/2})$. We also show that the RPM converges to a block-diagonal matrix of within-tier Gaussian rank probabilities, and we establish sufficient conditions for recovering the population tiers.

  \item \textbf{Validation and empirical audit.} We validate the correction via sample splitting---one half of the tasks selects the rank order while the held-out half rescores the selected models---and a calibrated simulation with known model means. Both exercises show substantial reductions in average rank-1 selection error. Auditing 14 benchmarks from 5 families, we find position-1 adjustments from 0.09 to 3.86 pp and multiple models sharing the leading tier in most benchmarks (Sections~\ref{sec:verification}--\ref{sec:results} and Appendix~\ref{app:robustness}).

  \item \textbf{Benchmark design guidance.} RULER also translates the analysis into a reporting standard and an asymptotic task-pool sizing rule. Under the exact-tier approximation, $N \gtrsim (\delta/\varepsilon)^2$ tasks are required to decrease the expected rank-1 inflation below $\varepsilon$, where $\delta$ is the within-tier Gaussian winning gap. Designers can use this calculation to plan larger task pools and publish RULER's score adjustments, RPMs, rank intervals, and tiers alongside raw rankings to show the uncertainty remaining at the available sample size (Section~\ref{sec:conclusion}).
\end{itemize}

\section{Model, Bootstrap, and Theory}\label{sec:model}

We develop a statistical framework for benchmark outcomes that enables the bootstrap inference for LLM leaderboards. We state the model and theory for bounded task-level scores, with binary pass/fail outcomes as the primary application to SWE-bench. Throughout, an LLM ``model'' denotes an evaluated leaderboard entry, including its agent or scaffold when applicable. All proofs appear in Appendix~\ref{app:proofs}.

Let $Y_{ij} \in [0,1]$ denote the score for model $j \in \{1,\ldots,J\}$ on task $i \in \{1,\ldots,N\}$, with binary success or failure as a special case. The observed data form an $N \times J$ outcome matrix $\mathbf{Y} = (Y_{ij})$, the transpose of whose $i$-th row we denote $\boldsymbol{Y}_{i\cdot} := (Y_{i1},\ldots,Y_{iJ})^{\top}$. We assume that these task rows $\boldsymbol{Y}_{1\cdot},\ldots,\boldsymbol{Y}_{N\cdot}$ are i.i.d.\ draws from a common distribution $P$ on $[0,1]^J$. We impose no parametric form on $P$ and allow arbitrary correlation and dependence among different models' outcomes on the same task.
Our primary quantity of interest is the true mean score $\tau_j = \E[Y_{ij}]$ for each model $j$, along with the induced true ranking. We collect these into the score vector $\boldsymbol{\tau} := (\tau_1,\ldots,\tau_J)^{\top}$ and write $\boldsymbol{\Sigma} := \mathrm{Cov}(\boldsymbol{Y}_{i\cdot})$ for the within-task covariance across models. The true ranking is a permutation $\pi^*\colon \{1,\ldots,J\} \to \{1,\ldots,J\}$ such that $\tau_{\pi^*(1)} \geq \tau_{\pi^*(2)} \geq \cdots \geq \tau_{\pi^*(J)}$ with a declared arbitrary tie breaking, where $\pi^*(k)$ is the index of the model whose true ability places it at rank $k$.

In practice, we observe only the empirical mean scores $\hat{\tau}_j = N^{-1}\sum_i Y_{ij}$, which we collect into $\hat{\boldsymbol{\tau}} := (\hat{\tau}_1,\ldots,\hat{\tau}_J)^{\top}$, and the corresponding empirical ranking $\hat{\pi}$, defined analogously by $\hat{\tau}_{\hat{\pi}(1)} \geq \cdots \geq \hat{\tau}_{\hat{\pi}(J)}$. 
We assume that there is an arbitrary but fixed tie-breaking rule that is being used throughout in computation.
Because leaderboards are read off sorted score vectors, our inferential targets are not the individual $\hat{\tau}_j$ but functionals of the sorted $\hat{\boldsymbol{\tau}}$. Each $\hat{\tau}_j$ is unbiased for $\tau_j$, but sorting introduces a selection effect that no single-model standard error can capture. We formalize two targets. The first is the \emph{rank-induced bias at position $k$},
\begin{equation*}
  b_N(k) \;:=\; \E\!\left[\hat{\tau}_{\hat{\pi}(k)} - \tau_{\hat{\pi}(k)}\right],
\end{equation*}
which is the bias induced by sorting. The quantity $b_N(1)$ measures the classical \emph{WC}: the expected score error of the winning model. The second target is the population RPM $\mathbf{R} \in [0,1]^{J \times J}$, with entry $R_{jk} = \mathbb{P}(\hat{\pi}(k) = j)$ giving the probability that model $j$ would be placed at rank $k$ on an independent draw of $N$ tasks from the same task distribution. A concentrated row signals a stable ranking, while a diffuse row signals that the model's position is poorly resolved by the available task pool. The bootstrap RPM defined below summarizes rankings across $m$-task resamples. Under the conditions of Proposition~\ref{prop:rank_probs}, it shares the population RPM's limiting rank probabilities. Ranking is nonsmooth at tied mean scores, so standard plug-in approximations can fail where selection effects are most pronounced.

\subsection{Bootstrap correction and components of RULER}\label{sec:bootstrap}

Both the rank-induced bias and the leaderboard uncertainty are functionals of the task distribution: if we knew the population distribution, we could repeatedly draw new leaderboards, rank the models, and directly measure the rank-induced bias. Although we do not observe the population, the leaderboard we already observe, i.e., the empirical distribution of tasks, is a good approximation of the population distribution. Thus, we can bootstrap from the observed tasks in the leaderboard, which preserves the dependence among models evaluated on the same task, and estimate the rank-induced bias from the observed leaderboard itself. The difficulty is that ranking is nonsmooth at ties, so the usual smooth-functional justification for the standard bootstrap fails \citep{fang2019inference}. The $m$-out-of-$N$ bootstrap \citep{bickel1997resampling} fixes this by resampling fewer tasks with $m = o(N)$. We follow the literature standards and use $m = \lfloor N^{\gamma}\rfloor$ for choosing $m$. To the best of our knowledge, there is no universally applicable way to tune the hyperparameter $\gamma$ with theoretical guarantees. Thus, we provide an empirical guidance under Appendix~\ref{app:gamma_guidance} and show that RULER performs relatively robust under different $\gamma$ in Appendix~\ref{app:mn_exponent_sensitivity}. 

\paragraph{Score Adjustment.}
The $m$-out-of-$N$ bootstrap works by drawing $B$ independent resamples of $m < N$ tasks with replacement. Letting $\hat{\tau}^{(b)}_j$ and $\hat{\pi}^{(b)}$ denote their means and rankings, the adjustment for position $k$ is
\begin{equation}
\label{eq:bias_estimator}
\hat b_N(k)=\sqrt{\frac mN}\frac1B \sum_{b=1}^B \left(\hat\tau^{(b)}_{\hat\pi^{(b)}(k)}-\hat\tau_{\hat\pi^{(b)}(k)}\right).
\end{equation}
Intuitively, it measures how much a bootstrap-selected model's resample score exceeds its original score, then rescales this excess to the full-sample noise level. The adjustment belongs to a \emph{rank position}, rather than to a particular model. 

\paragraph{RPM and rank interval.}
We can use the same $m$-task resamples to measure the uncertainty in the original leaderboard. The first uncertainty measure is the probability of each model occupying a specific rank position, called the Rank Probability Matrix (RPM): 
\begin{equation}
\label{eq:rank_prob}
\hat R_{jk}=B^{-1}\sum_{b=1}^B \mathbbm{1}[\text{model $j$ ranks at position $k$ in the $b$-th bootstrap sample}], 
\end{equation}
where $\mathbbm{1}[\cdot]$ is the indicator function.\footnote{If $q$ models tie in a bootstrap resample, each model receives probability mass $1/q$ at every occupied position. This averages over all orderings within the tie and keeps every row and column of the RPM summing to one.} The RPM shows how uncertain a leaderboard ranking can be. Thus, we develop a second metric, the rank interval, that measures the possible rank positions of each model. Let $F_j(k)=\sum_{s\leq k}\hat R_{js}$ denote the probability that model $j$ ranks at or above position $k$, and $Q_j(u)=\min\{k:F_j(k)\geq u\}$ the smallest rank position at which this cumulative probability reaches $u$. Then, the rank interval for model $j$ is computed as:
\begin{equation}
\label{eq:rank_interval}
I_j(\alpha)=[Q_j(\alpha/2),Q_j(1-\alpha/2)].
\end{equation}
These positions contain at least $1-\alpha$ of model $j$'s bootstrap rank mass. We default to $\alpha=0.20$ throughout this paper.

\paragraph{Model tiers.}
Finally, small benchmark samples often leave the ranking of closely performing models unresolved, with no clear winner. We quantify this uncertainty using tiers of models whose rank intervals have a common intersection. Specifically, we start with the first model on the leaderboard for the first tier, and use its rank interval to represent all possible rank positions for the first tier. Then, we add the next model on the board if its rank interval contains a common rank position as the first tier. If yes, we take an intersection of the rank intervals of all models in the tier. Otherwise, close the tier and start a new one. As a result, the rank intervals of all models in a tier share at least one rank. Models with equal observed scores are always placed in the same tier (Appendix~\ref{app:implementation}).

\paragraph{RULER.}
Together, the score adjustment, RPM, and rank intervals with model tiers form the \textit{Rank Uncertainty and Leaderboard Evaluation Report (RULER)}. 

\subsection{Theoretical properties}
\label{sec:theory}

We now state the theoretical results that justify the $m$-out-of-$N$ bootstrap behind RULER. We show that RULER's score adjustment reduces $\Theta(N^{-1/2})$ rank-induced bias to $o(N^{-1/2})$ and that the reported tiers can consistently recover the population model tiers. For a fixed number of models $J$, we make the following assumptions.

\begin{assumption}[Exact population tiers with positive gaps]\label{assmp:tier}
The models partition into $S_1,\ldots,S_L$, with $\tau_j=\tau^{(\ell)}$ for $j\in S_\ell$ and $\tau^{(1)}>\cdots>\tau^{(L)}$. If $L>1$, the smallest adjacent gap is bounded below by a fixed $\Delta>0$.
\end{assumption}

Assumption \ref{assmp:tier} is a deliberate modeling idealization rather than a literal claim that in reality there are always different models with identical population performance. In practice, models may differ in their true mean scores, but often these differences are very small and become statistically indistinguishable with the finite available data. We therefore use exact ties as a tractable approximation for groups of closely performing models whose ordering is difficult to resolve. This idealization captures the selection effects that are particularly pronounced among closely competing models. Under this assumption, we derive theoretical results to provide an informative benchmark for understanding rank-induced bias, and evaluate the practical usefulness of the proposed correction through the finite-sample validation in Section 3. The tiers reported by RULER similarly represent unresolved rankings rather than assertions of exact equality in population performance.

In addition to Assumption \ref{assmp:tier}, 
note that we have assumed that tasks are i.i.d. to facilitate a clean theoretical analysis, as discussed at the beginning of Section \ref{sec:model}.
However, our sensitivity analysis shows that RULER's score adjustments are broadly robust across alternative resampling schemes for non-i.i.d. tasks on SWE-bench, as discussed in Section~\ref{subsec:dependence} and Appendix~\ref{app:dependence}. 

Next, we formalize the magnitude of the rank-induced bias in a leaderboard. Let $K_\ell=\sum_{q<\ell}|S_q|$ denote the number of models sitting above tier $\ell$, $\boldsymbol{Z}\sim\mathcal N(\boldsymbol{0},\boldsymbol{\Sigma})$ a mean-zero Gaussian vector with covariance $\boldsymbol{\Sigma}$, and $Z_{\ell,(r)}$ the $r$-th largest coordinate in $S_\ell$. 
Equivalently, $Z_{\ell,(r)}$ is the $|S_\ell|-r +1$-th ordered statistics of $\{Z_j:~j \in S_{\ell}\}$. 

\begin{proposition}[First-order rank-induced bias at every fixed position]\label{prop:bias_expansion}
At position $k=K_\ell+r$,
\begin{equation}
 b_N(k)=\frac{\delta_{\ell r}}{\sqrt N}+o(N^{-1/2}),\qquad
 \delta_{\ell r}=\E[Z_{\ell,(r)}].\label{eq:bias_expansion}
\end{equation}
\end{proposition}
Therefore, $\delta_{\ell r}$ is the position-induced effect. 
The winner's curse bias \citep{xu2025winners} is a special case of Eq.~\ref{eq:bias_expansion} with $b_N(1)$. The proposition shows that rank-induced bias can be $\Theta(N^{-1/2})$. Next, we show that the $m$-out-of-$N$ bootstrap score adjustment in Eq.~\ref{eq:bias_estimator} reduces this bias to $o(N^{-1/2})$.

\begin{theorem}[Bootstrap score adjustment]\label{thm:moon_validity}
Assume that as $N \to \infty$, $m\to\infty$, $m/N\to0$, and $B\to\infty$. At every fixed position $k=K_\ell+r$,
\begin{equation}
 \hat b_N(k)=\delta_{\ell r}/\sqrt N+o_P(N^{-1/2}),\qquad
 \E[\hat\tau_{\hat\pi(k)}-\hat b_N(k)] = \E[\tau_{\hat\pi(k)}] + o(N^{-1/2}).
\label{eq:all_position_correction}
\end{equation}
\end{theorem}

The second equation in \eqref{eq:all_position_correction} states that, at every fixed position, the adjusted score is unbiased for the true score of the model placed there, up to an error of order $o(N^{-1/2})$. The argument has two steps. First, the bootstrap reproduces the selection effect on the $m$-task scale. In a resample of $m$ tasks, each model's resample score differs from its observed score by noise of order $m^{-1/2}$. Within a tier, observed score gaps are of order $N^{-1/2}$, which is negligible relative to this noise because $m/N\to0$. Between tiers, observed gaps remain near the population gaps and dominate the noise. Resample rankings therefore keep the tiers in order and rank models within a tier by noise alone, as in the population. The average in Eq.~\eqref{eq:bias_estimator} before rescaling thus equals $\delta_{\ell r}/\sqrt m+o_P(m^{-1/2})$. Second, the factor $\sqrt{m/N}$ converts this to $\delta_{\ell r}/\sqrt N$, the leading term of $b_N(k)$ in Proposition~\ref{prop:bias_expansion}. Subtracting $\hat b_N(k)$ therefore removes the first-order bias.

Finally, we show that the RPM and model tiers estimated using the same $m$-out-of-$N$ bootstrap are consistent estimators. 
For $j\in S_\ell$, define $p_{jr}^{(\ell)}:=\Pr\{Z_j\text{ is }r\text{th largest in }S_\ell\}$ and note that 
$
\sum_{r=1}^{|S_{\ell}|}p_{jr} = 1
$.
Further, we define the population rank probability matrix (RPM) with entries $R_{jk}:=\Pr\{\hat\pi(k)=j\}$.
These are probabilities over repeated samples of $N$ tasks. They depend on $N$,
although that dependence is suppressed in the notation.

\begin{proposition}[Convergence of RPM and consistent tier recovery]\label{prop:rank_probs}
Assume the fixed-$J$ bounded iid task model and the exact population tiers with positive gaps. Let $m=m_N$ and $B=B_N$ be deterministic positive integer sequences with $m\to\infty$, $m/N\to0$, and $B\to\infty$ as $N\to\infty$. Suppose $\operatorname{Var}(Z_j-Z_h)>0$ for distinct $j,h$ in each non-singleton tier. 

For every tier $S_\ell$, model $j\in S_\ell$, and $r\in\{1,\ldots,q_\ell\}$,
\begin{equation}\label{eq:rpm_block}
  R_{j,K_\ell+r}\longrightarrow p_{jr}^{(\ell)},\qquad
  \hat R_{j,K_\ell+r}\xrightarrow{P}p_{jr}^{(\ell)}.
\end{equation}
For $j\in S_\ell$ and $k\notin \{K_\ell+1, \dots, K_\ell + q_\ell\}$,
\begin{equation}\label{eq:rpm-outside-block}
  R_{jk}\longrightarrow0,\qquad\hat R_{jk}\xrightarrow{P}0.
\end{equation}
Consequently, the bootstrap RPM consistently estimates the population RPM:
\begin{equation}\label{eq:rpm-consistency}
  \max_{1\leq j,k\leq J}|\hat R_{jk}-R_{jk}|\xrightarrow{P}0.
\end{equation}
For every fixed $\alpha\in(0,1)$, the common-intersection rule returns exactly
$(S_1,\ldots,S_L)$, in that order, with probability tending to one.
\end{proposition}
Proposition~\ref{prop:rank_probs} shows that RULER separates genuine performance differences between tiers while retaining ranking uncertainty within them. As the number of tasks grows, the RPM assigns vanishing probability to ranks outside a model's population tier. Within a tier, the models have equal true scores, so their relative ranks remain sensitive to sampling noise. The resulting rank intervals share a common position within each tier and separate across tiers with probability tending to one, allowing RULER to recover the population partition. The full proof appears in Appendix~\ref{app:proof_rpm}.

\section{Validation}
\label{sec:verification}

Having established RULER's theoretical guarantees, we now validate its performance. Specifically, we use a held-out test and calibrated simulations to evaluate how much its score adjustment reduces selection error in finite samples. Both studies show substantial reductions in average selection error.

\subsection{Held-out Test}

The held-out test separates rank selection from evaluation by assigning them to non-overlapping subsets of the task pool. For each benchmark, we randomly split the tasks into a selection half $S$ with $N_S=\lfloor N/2\rfloor$ tasks and an evaluation half $E$ with the remaining $N_E=N-N_S$ tasks. We treat the selection half as if it were the entire benchmark. Using only its $N_S$ tasks, we rank the models by their scores $\hat\tau^{S}_j$ to obtain the ranking $\hat\pi_S$, and apply the bootstrap in Equation~\ref{eq:bias_estimator} with $m=\lfloor N_S^{0.9}\rfloor$ to obtain the adjustment $\hat b^{S}(k)$ at each position $k$. The selected models are then scored on $E$. Because the evaluation half plays no role in ranking or adjustment, its score $\hat\tau^{E}_j$ provides an unbiased estimate of the selected model's mean under independent task sampling.\footnote{Note that $\hat\tau^{E}_j$ is itself subject to considerable sampling noise because he evaluation half contains only half of an already limited task pool. Therefore, the held-out test checks whether the corrected scores agree with this unbiased reference on average, rather than providing a precise comparison at each position.}

For each position $k$, we compare the raw and adjusted errors
\begin{equation}
 d_k=\hat\tau^{S}_{\hat\pi_S(k)}-\hat\tau^{E}_{\hat\pi_S(k)},\qquad
 e_k=d_k-\hat b^{S}(k).\label{eq:split_reference}
\end{equation}
The expectation of the raw error $d_k$ is the rank-induced bias $b_{N_S}(k)$ of a leaderboard with $N_S$ tasks. If the bootstrap removes the bias present in the selection half, the adjusted error $e_k$ should be centered near zero.\footnote{Proposition~\ref{prop:bias_expansion} shows that the leading term of the rank-induced bias is proportional to $N^{-1/2}$. A leaderboard built from half the tasks therefore has approximately $\sqrt2$ times the bias of the full leaderboard.} We repeat the test over 200 random partitions per benchmark. Appendix~\ref{app:split_protocol} documents the implementation details. 

\begin{table}[!t]
\centering\small
\caption{The correction reduces average selection error across 14 benchmarks.}\label{tab:split_aggregate}
\begin{tabular}{@{}cccc@{}}
\toprule
& \multicolumn{3}{c}{Magnitude of Mean Error} \\
\cmidrule(l){2-4}
Rank & Before RULER & After RULER & Reduction (\%) \\
\midrule
1 & 2.38 & 0.73 & 69.5 \\
2 & 1.43 & 0.74 & 48.0 \\
3 & 1.34 & 0.71 & 46.9 \\
4 & 1.10 & 0.50 & 55.0 \\
5 & 1.23 & 0.58 & 52.7 \\
\bottomrule
\end{tabular}

\floatnote{Each benchmark uses 200 random half-splits, with $m=\lfloor N_S^{0.9}\rfloor$ and $B=2{,}000$ resamples on the selection half. Before RULER uses the selected model's score on selection half $S$ minus its score on evaluation half $E$. After RULER subtracts the estimated bias from that difference. Magnitude of mean error first averages these signed differences over splits within each benchmark, then takes the absolute value and averages across the 14 benchmarks. Its reduction is reported as a percentage. Errors are in percentage points (pp)}
\end{table}

At rank~1, RULER's score adjustment reduces the magnitude of mean split error from 2.38 to 0.73 percentage points (pp), a reduction of 69.5\% across all 14 benchmarks (Table~\ref{tab:split_aggregate}). Positions 2--5 in the same table also show similar improvement, ranging from 46.9\% to 55.0\%. Appendix~\ref{app:split_protocol} reports the complete results. Thus, the held-out test broadly demonstrates that the score adjust can correct the selection effects generally across the leaderboard. 

\subsection{Calibrated Simulation}

While the held-out test uses an independently-estimated score as its reference, we provide a second validation using simulated benchmarks where the models' true scores are known. To mimic realistic settings, we calibrate the simulations to SWE-bench Verified. Specifically, we fit a logistic model combining model-specific abilities with a shared task difficulty, so outcomes for different models on the same task are dependent (see Appendix~\ref{app:alternatives} for more details). From this fitted model, we generate 100 synthetic benchmarks and apply the $m$-out-of-$N$ bootstrap with 2,000 resamples to each. Table~\ref{tab:calib_sim} compares the raw and corrected scores of the models at positions 1--5 with their known means.

\begin{table}[!t]
\centering\small
\caption{The $m$-out-of-$N$ correction removes most of the winner's bias and lowers its root mean squared error.}\label{tab:calib_sim}
\begin{tabular}{@{}ccccc@{}}
\toprule
& \multicolumn{2}{c}{Mean Signed Error (pp)} & \multicolumn{2}{c}{RMSE (pp)} \\
\cmidrule(lr){2-3}\cmidrule(l){4-5}
Rank & No Correction & \shortstack{$m$-out-of-$N$\\bootstrap} & No Correction & \shortstack{$m$-out-of-$N$\\bootstrap} \\
\midrule
1 & $+1.41\pm0.14$ & $+0.21\pm0.15$ & $1.97\pm0.13$ & $1.55\pm0.10$ \\
2 & $+0.80\pm0.17$ & $-0.06\pm0.17$ & $1.85\pm0.18$ & $1.74\pm0.14$ \\
3 & $+0.66\pm0.18$ & $-0.12\pm0.18$ & $1.87\pm0.16$ & $1.78\pm0.14$ \\
4 & $+0.98\pm0.16$ & $+0.23\pm0.16$ & $1.85\pm0.11$ & $1.64\pm0.13$ \\
5 & $+0.96\pm0.17$ & $+0.24\pm0.18$ & $1.97\pm0.15$ & $1.79\pm0.13$ \\
\bottomrule
\end{tabular}

\floatnote{Entries are the mean signed error and root mean squared error (RMSE) of the selected model's score relative to its known mean, $\pm$ standard errors, in pp. The $m$-out-of-$N$ bootstrap uses $m=\lfloor N^{0.9}\rfloor $ and 2,000 resamples.}
\end{table}

The $m$-out-of-$N$ correction provides more accurate evaluation to model performance. The raw winner overstates its true mean by 1.41 pp on average, and the $m$-out-of-$N$ correction reduces it to 0.21 pp. The correction also brings the mean errors at positions 2--5 closer to zero. It lowers the winner's RMSE from 1.97 to 1.55 pp and reduces the RMSE at positions 4 and 5 by about 0.2 pp. Appendix~\ref{app:alternatives} also compares the $m$-out-of-$N$ correction with the standard bootstrap ($m=N$), a beta-binomial empirical-Bayes posterior mean, and a correction built from sample splitting. The $m$-out-of-$N$ also performs well generally among the 4 corrections.

\section{Empirical Study Across Benchmark Families}\label{sec:results}

The validation shows that RULER can substantially reduce average selection error. We now use it to ask how much public leaderboard scores are overstated and how precisely their task pools resolve the model ordering. Across the 14 benchmarks, rank-1 adjustments exceed 1 pp in 7, and 12 have multiple models in the leading tier. We begin with SWE-bench Verified to show how the score and rank summaries work together, then assess how broadly the findings hold across benchmark families and alternative reporting choices. Appendices~\ref{app:implementation} documents the implementation details

\subsection{Leaderboard Results}

SWE-bench Verified evaluates 173 models on 500 tasks extracted from 12 GitHub repositories in our snapshot.\footnote{The snapshot combines the \href{https://github.com/SWE-bench/swe-bench.github.io/blob/7c4289f30aa1a1c63c2e2a25aae30c16d92b5114/data/leaderboards.json}{official SWE-bench leaderboard} dated February 27, 2026 (commit~\texttt{7c4289f}) with task-level outcomes from \href{https://github.com/SWE-bench/experiments/tree/82be2b5ac4967af9559484e7bffef227e97348af/evaluation/verified}{SWE-bench/experiments} (commit~\texttt{82be2b5}).} Its 2 tied leaders score 79.2\%. RULER estimates a downward adjustment of 0.90 pp for each, giving adjusted scores of 78.30\% (Table~\ref{tab:bias_top10}). Thus the winning score includes a measurable selection advantage.

\begin{table}[!t]
\centering\small
\caption{RULER reveals score overstatement and unresolved ranks among SWE-bench Verified's leaders.}\label{tab:bias_top10}
\begin{tabular}{@{}c>{\raggedright\arraybackslash}p{4.7cm}cccc@{}}
\toprule
\shortstack{Raw\\Rank} & Model & \shortstack{Raw Score\\(\%)} & \shortstack{Score Correction\\(pp)} & \shortstack{80\% Rank\\Interval} & Tier \\
\midrule
1--2 & live-SWE + Claude 4.5 Opus med. & 79.20 & 0.90 & [1,6] & 1 \\
1--2 & Sonar + Claude 4.5 Opus & 79.20 & 0.90 & [1,6] & 1 \\
3 & TRAE + Doubao-Seed-Code & 78.80 & 0.60 & [1,7] & 1 \\
4 & live-SWE + Gemini 3 Pro & 77.40 & 0.69 & [2,13] & 1 \\
5--7 & Atlassian Rovo Dev & 76.80 & 0.63 & [3,15] & 1 \\
5--7 & EPAM AI/Run v20250719 + Claude 4 Sonnet & 76.80 & 0.63 & [3,15] & 1 \\
5--7 & mini-SWE-agent + Claude 4.5 Opus (high) & 76.80 & 0.63 & [3,16] & 1 \\
\bottomrule
\end{tabular}

\floatnote{The table shows the top 7 models from SWE-bench Verified. RULER uses $m=\lfloor N^{0.9}\rfloor$ and $B=5{,}000$. Score Correction is the amount subtracted from the raw score. Rank intervals contain at least 80\% of each model's bootstrap rank mass.}
\end{table}

In addition to score adjustment, RULER also suggests that the ranking is much less decisive than the raw scores suggest. The 2 leaders receive 30.5\% and 29.0\% of the bootstrap rank-1 mass, while the model at rank 3 receives 26.4\%. Both leaders' 80\% rank intervals span positions 1--6, and the rank-3 model's interval spans 1--7. Uncertainty extends farther down the leaderboard: the models tied at positions 9--10 each have an interval of [5,21] (Figure~\ref{fig:swebench_verified_unresolved}). The common-intersection rule groups 13 models into the leading tier, with rank~6 contained in every member's interval, and the full leaderboard has 15 tiers in total.

\begin{table}[!t]
\centering\small
\setlength{\tabcolsep}{3pt}
\caption{Score overstatement and unresolved leaders occur across benchmark families.}\label{tab:cross_variant_summary}
\begin{tabular}{@{}lcccccc@{}}
\toprule
Benchmark & \shortstack{Number\\of Tasks} & \shortstack{Number\\of Models} & \shortstack{Position 1\\Score (Raw)} & \shortstack{Position 1\\Score (Adjusted)} & \shortstack{Number\\of Tiers} & \shortstack{Number of\\Models in Tier 1} \\
\midrule
SWE Bash-only & 500 & 41 & 76.80 & 75.71 & 10 & 6 \\
SWE Verified & 500 & 173 & 79.20 & 78.30 & 15 & 13 \\
SWE Multimodal & 510 & 12 & 36.47 & 35.84 & 3 & 5 \\
SWE Multilingual & 300 & 14 & 72.67 & 71.17 & 4 & 9 \\
SWE Pro & 395 & 14 & 44.30 & 43.43 & 6 & 4 \\
Tau2 Airline & 50 & 7 & 69.00 & 65.93 & 2 & 5 \\
Tau2 Retail & 114 & 7 & 84.21 & 82.01 & 2 & 5 \\
Tau2 Telecom & 114 & 7 & 98.25 & 98.07 & 6 & 1 \\
AlpacaEval 2 & 805 & 208 & 77.83 & 77.29 & 21 & 3 \\
WildBench v2 & 988 & 63 & 79.81 & 79.71 & 15 & 2 \\
LiveBench pooled & 318 & 81 & 77.56 & 77.44 & 9 & 1 \\
LiveBench Coding & 128 & 81 & 86.72 & 85.61 & 8 & 2 \\
LiveBench Language & 140 & 81 & 69.81 & 67.91 & 7 & 4 \\
LiveBench Instr. & 50 & 81 & 86.78 & 82.93 & 5 & 19 \\
\bottomrule
\end{tabular}

\floatnote{The raw and adjusted scores are expressed as percentages. Tier counts and Tier 1 sizes use 80\% rank mass and the common-intersection rule. All 14 benchmarks use $m=\lfloor N^{0.9}\rfloor$ and $B=5{,}000$. Appendix~\ref{app:implementation} documents the implementation details.}
\end{table}

These patterns extend beyond SWE-bench Verified: score overstatement and unresolved leaders appear across benchmark families (Table~\ref{tab:cross_variant_summary}). On Tau2 Airline, the leader's score falls by 3.07 pp after adjustment and 5 of 7 models share the leading tier. LiveBench instruction following has the largest adjustment, 3.86 pp, with 19 of 81 models in the leading tier. Both benchmarks contain only 50 tasks. Even a small correction can accompany an unresolved leader, as WildBench's 0.09-pp adjustment and 2-model leading tier demonstrate. Across the audit, RULER therefore reveals both how much the winning score is overstated and how much ambiguity remains in the model ordering. Appendix~\ref{app:implementation} explains how we apply the same procedure to each benchmark's task-level scores.

\subsection{Robustness Study}\label{subsec:dependence}

The score adjustment for SWE-bench Verified is robust to the i.i.d. task assumptions. Appendix~\ref{app:dependence} examines alternative sampling schemes for different assumptions on task distributions. Specifically, we compare pooled resampling with repository-stratified resampling, which approximately fixes the mix of repositories, and a cluster stress test, which allows tasks from the same repository to move together. Verified's position-1 adjustment is 1.12 pp under pooled resampling, 1.20 pp under stratification, and 1.16 pp under clustering. Other benchmarks, such as Telecom, could be more sensitive to alternative assumptions on task distributions, as detailed in Appendix~\ref{app:dependence}. We also examine the RULER's robustness against different resample size for the $m$-out-of-$N$ bootstrap and the retained rank mass used for model tiering. Across resample-size exponents from 0.4 to 0.9, score adjustments are stable and keep their direction across most rank positions. Increasing rank mass broadens the reported uncertainty, expanding Verified's leading tier from 13 models at 80\% to 15 at 90\%, 24 at 95\%, and 29 at 99\%. Appendix~\ref{app:robustness} gives the procedures and full results.

\section{Conclusion}\label{sec:conclusion}

Public LLM leaderboards rank noisy estimates of model performance, thus the ranking rewards sampling luck alongside genuine ability. As a result, the winner's score is overstated (the winner's curse), scores throughout the leaderboard are distorted (rank-induced bias), and the ordering itself is uncertain. RULER addresses both the score distortion and the rank uncertainty. Its score adjustments correct each reported score for the selection effect of its position, and its RPM, rank intervals, and tiers show how firmly each model holds that position. Theorem~\ref{thm:moon_validity} shows that the adjustment reduces rank-induced bias from $\Theta(N^{-1/2})$ to $o(N^{-1/2})$, and Proposition~\ref{prop:rank_probs} shows that RULER's tiering can recover the population model tiers. Across 14 benchmarks, rank-1 adjustments exceed 1 pp in 7 and the leading tier contains multiple models in 12 of the benchmarks. Score distortion and rank uncertainty are therefore common across the audited benchmark families. Because both arise from ranking noisy scores, they are likely to affect other leaderboards that compare closely matched models on limited task pools.

These findings have implications for model adopters, benchmark designers, and model developers. First, firms choosing a model should account for both the uncertainty in the ranking and the overstatement of top scores. A choice based on raw scores and ranks overstates the selected model's expected performance and treats an unresolved ordering as settled, which can lead to suboptimal adoption decisions. Second, benchmark designers can reduce both problems by enlarging the task pool. Under the exact-tier approximation, halving expected winner inflation requires approximately 4 times as many tasks when the model pool and task distribution remain stable. Appendix~\ref{app:tables} provides a detailed sizing calculation. Finally, model developers who claim state-of-the-art performance should lead by a margin wider than the winner's curse. On SWE-bench Verified, 7 of the 13 increases in the top score during 2025 were smaller than the estimated inflation of the current leaders. A confident claim needs a lead larger than the estimated inflation, and RULER's tiers show whether the new leader is separated from its closest competitors.

\paragraph{Scope and Limitations.}
RULER describes what a specified benchmark can resolve about the evaluated models. Its score adjustments and rank summaries concern sampling tasks from that benchmark's target population. Their relevance to deployment depends on the benchmark's representativeness and the meaning of its score. The audit conditions on released responses and judge annotations, so variation from generating new responses or rerunning judges requires a separate assessment. Contamination, adaptive submissions, and selective reporting can introduce additional biases beyond task-sampling selection. RULER's finite-sample performance also depends on the data-generating process. When mean differences are comparable to sampling noise, the correction can leave residual bias and may overcorrect. 

\bibliographystyle{plainnat}
\bibliography{references}

\clearpage
\appendix

\noindent\textbf{Guide to the appendices.}
Appendix~\ref{app:toy} uses a two-model example to build intuition for how the ability gap and resample size affect score correction and ranking uncertainty, and Appendix~\ref{app:gamma_guidance} uses simulations with known model means to guide the choice of resample size. Appendix~\ref{app:proofs} proves the main results. Appendix~\ref{app:implementation} documents data and implementation. Appendix~\ref{app:validation} reports supplementary validation results, and Appendix~\ref{app:robustness} reports robustness studies. Appendix~\ref{app:tables} provides supplementary RULER reports and task-pool planning, followed by related work in Appendix~\ref{app:related}.

\section{Intuition from a Two-Model Example}\label{app:toy}
With two models, the relation between the ability gap and sampling noise is easy to see: a near tie produces frequent changes in the winner, while a large gap makes the better model's position more stable. The same example shows why very small resamples can obscure a real gap.

We use independent Bernoulli outcomes for two models on $N=500$ tasks, with probabilities $(0.500,0.505)$ or $(0.500,0.600)$. We generate 3,000 benchmark draws and use $m=\lfloor N^{0.9}\rfloor$, with 2,000 bootstrap replicates per draw and seed 20260502. Multinomial sampling of the four joint-outcome cells is exactly equivalent in distribution to resampling complete binary task rows. Observed score ties are broken uniformly. Bootstrap RPM ties receive fractional mass. Bootstrap selection gaps are averaged over tied winners.

\begin{figure}[!t]
\centering\includegraphics[width=\textwidth]{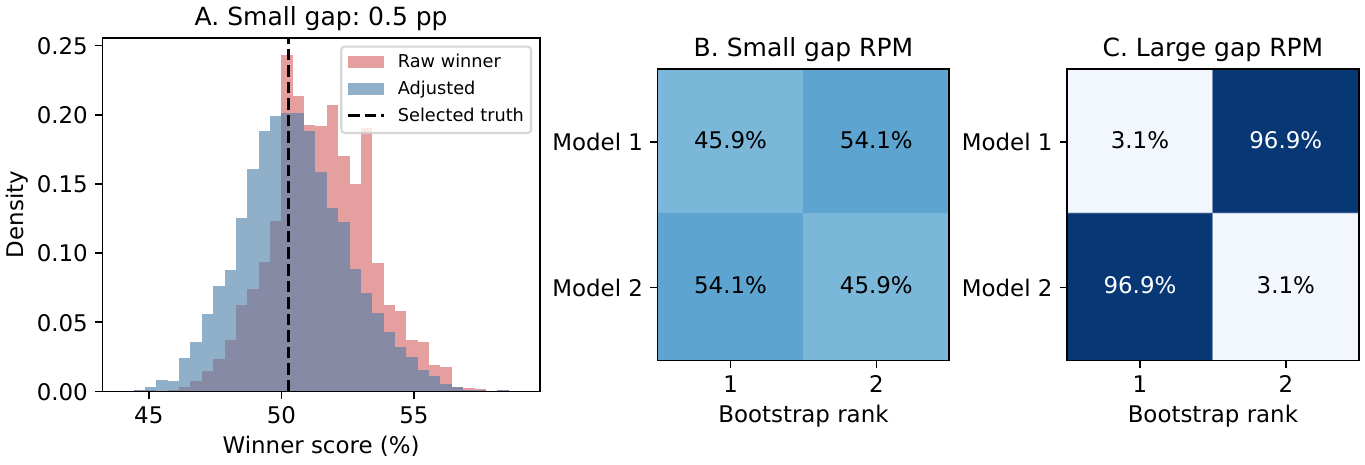}
\caption{Score correction reduces inflation while near-tied models remain hard to rank.}\label{fig:two_model_toy}

\floatnote{Two models have independent Bernoulli outcomes with probabilities $(0.500,0.505)$ or $(0.500,0.600)$ on $N=500$ tasks. Each case uses 3,000 benchmark draws, $m=\lfloor N^{0.9}\rfloor$, and 2,000 resamples per draw. Left: raw and adjusted winning scores for the 0.5-pp gap, with the selected model's mean true ability marked by the dashed line. Middle and right: fractional-tie RPMs averaged over benchmark draws for the 0.5-pp and 10-pp gaps.}
\end{figure}

With a 0.5-pp ability gap, selection produces substantial score inflation and leaves the winner uncertain. The raw winning score averages 51.51\%, while the selected model's true ability averages 50.28\% (Figure~\ref{fig:two_model_toy}). Mean bias falls from 1.23 pp (Monte Carlo SE 0.034) to a residual of 0.22 pp (SE 0.037) after correction. Exact summation over the two independent binomial count distributions gives population raw bias 1.246 pp. The worse model wins in 43.4\% of draws, and the better model receives only 54.1\% rank-1 mass in the averaged bootstrap RPM. Thus the correction reduces inflation while the RPM continues to show an unresolved ordering.

A 10-pp ability gap largely resolves the winner: the better model's averaged rank-1 mass rises to 96.9\%, and exact population winner bias falls to 0.0078 pp. The bootstrap nevertheless applies a mean adjustment of 0.176 pp, leaving a simulated mean residual of $-0.238$ pp (SE 0.043).\footnote{The simulated raw bias is $-0.062$ pp (SE 0.041). Its difference from the positive exact bias reflects Monte Carlo variation.} This example shows that the finite-sample correction can overshoot even when the winning model is well separated.

Smaller resamples make that separation harder to detect. An additional experiment uses $m=\lfloor N^{0.6}\rfloor$, 3,000 draws, and 2,000 resamples per draw. For the 10-pp gap, the better model's averaged rank-1 mass falls to 80.8\%, compared with 96.9\% at $m=\lfloor N^{0.9}\rfloor$, and the mean adjustment rises to approximately 0.822 pp. Raw bias of 0.033 pp then becomes a residual of $-0.789$ pp. For the 0.5-pp gap, however, raw bias of 1.257 pp becomes a smaller residual of 0.048 pp. The examples expose the tradeoff: smaller resamples suppress apparent differences near a tie but can also obscure genuine ability gaps. The experiments use different random streams, so the comparison across resample sizes is illustrative rather than paired.

\section{Guidance on Choosing the Resample Size for the $m$-out-of-$N$ Bootstrap}\label{app:gamma_guidance}
The theory requires $m=o(N)$ but does not fix the resample size for a given benchmark. We write $m=\max\{2,\lfloor N^\gamma\rfloor\}$ with exponent $\gamma\in(0,1]$, so $\gamma=1$ is the standard bootstrap. The two-model example in Appendix~\ref{app:toy} shows what the exponent trades off: smaller resamples suppress apparent differences near a tie but can also obscure genuine ability gaps. The true means of the audited models are unknown, so the audit cannot show which exponent is most accurate. A simulation with known model means measures that accuracy directly. The most accurate exponent depends mainly on how closely model abilities are spaced and on the number of models. When abilities are closely spaced, every exponent from 0.4 to 0.9 is nearly as accurate as the best, and the default $\gamma=0.9$ is within 8\% of the most accurate exponent unless all models are tied.

Each synthetic benchmark has $J$ models and $N$ tasks with independent Bernoulli outcomes. The true model means are evenly spaced over an interval of width $w\sigma$ centered at $p$, where $\sigma=\sqrt{p(1-p)/N}$ is the standard error of one model's score. Thus $p$ is both the mean and the median ability. In the tied design ($w=0$), every model has mean $p$. In the close design ($w=2$), the means span 2 standard errors. In the spread design ($w=15$), they span 15. The width does not change with $J$, so larger leaderboards pack their models more densely. The grid crosses $N\in\{500,2{,}000\}$, $p\in\{0.5,0.75\}$, and $J\in\{2,5,10,20,50,100\}$ with 11 exponents $\gamma\in\{0.1,0.2,\ldots,0.9,0.95,1\}$. Each design uses 1,000 synthetic benchmarks and $B=2{,}000$ resamples. Every exponent is applied to the same benchmarks, so comparisons across exponents are paired.

For each synthetic benchmark, the error at position $k$ is the corrected score at that position minus the true mean of the model observed there. We average its square over all $J$ positions and over benchmarks and take the square root to obtain the root-mean-square error (RMSE). Tables~\ref{tab:gamma_table_tied}--\ref{tab:gamma_table_spread} report it for every exponent. In each column, the 3 smallest displayed values are shaded from dark to light, and equal displayed values share a shade, so a long shaded block marks exponents with nearly equal accuracy. We call the exponents tied for the smallest displayed RMSE in a column the most accurate.

When all models are tied, every observed difference is noise, and exponents well below the default give lower error (Table~\ref{tab:gamma_table_tied}). The most accurate exponents lie between 0.1 and 0.7 and rise with $J$, reaching 0.4--0.6 at $J=100$. The default $\gamma=0.9$ has an RMSE 8--18\% above that of the most accurate exponent for $J\le10$ and 47--57\% above it at $J=100$. The standard bootstrap is 12--110\% above it.

\begin{table}[!t]\centering\small
\setlength{\tabcolsep}{4pt}
\caption{When all models are tied, exponents below the default minimize error, and their advantage over the default grows with the number of models.}\label{tab:gamma_table_tied}
\begin{tabular}{@{}c*{12}{c}@{}}
\toprule
 & \multicolumn{6}{c}{$p=0.5$} & \multicolumn{6}{c}{$p=0.75$} \\
\cmidrule(lr){2-7}\cmidrule(lr){8-13}
$\gamma\,/\,J$ & 2 & 5 & 10 & 20 & 50 & 100 & 2 & 5 & 10 & 20 & 50 & 100 \\
\midrule
\multicolumn{13}{@{}l}{$N=500$ tasks} \\
0.1 & \cellcolor[HTML]{0B5D1E}\color{white}\bfseries 1.87 & \cellcolor[HTML]{0B5D1E}\color{white}\bfseries 1.35 & \cellcolor[HTML]{2E8B3E}\color{white}\bfseries 1.05 & 0.87 & 0.81 & 0.83 & \cellcolor[HTML]{0B5D1E}\color{white}\bfseries 1.64 & 1.23 & 1.05 & 0.94 & 0.91 & 0.94 \\
0.2 & \cellcolor[HTML]{0B5D1E}\color{white}\bfseries 1.87 & \cellcolor[HTML]{0B5D1E}\color{white}\bfseries 1.35 & \cellcolor[HTML]{0B5D1E}\color{white}\bfseries 1.03 & \cellcolor[HTML]{CDEFD4}\bfseries 0.79 & 0.64 & 0.63 & \cellcolor[HTML]{0B5D1E}\color{white}\bfseries 1.64 & \cellcolor[HTML]{CDEFD4}\bfseries 1.20 & 0.97 & 0.82 & 0.75 & 0.76 \\
0.3 & \cellcolor[HTML]{2E8B3E}\color{white}\bfseries 1.88 & \cellcolor[HTML]{0B5D1E}\color{white}\bfseries 1.35 & \cellcolor[HTML]{0B5D1E}\color{white}\bfseries 1.03 & \cellcolor[HTML]{0B5D1E}\color{white}\bfseries 0.77 & \cellcolor[HTML]{CDEFD4}\bfseries 0.53 & 0.45 & \cellcolor[HTML]{0B5D1E}\color{white}\bfseries 1.64 & \cellcolor[HTML]{0B5D1E}\color{white}\bfseries 1.18 & \cellcolor[HTML]{2E8B3E}\color{white}\bfseries 0.92 & \cellcolor[HTML]{CDEFD4}\bfseries 0.70 & 0.55 & 0.51 \\
0.4 & \cellcolor[HTML]{2E8B3E}\color{white}\bfseries 1.88 & \cellcolor[HTML]{0B5D1E}\color{white}\bfseries 1.35 & \cellcolor[HTML]{0B5D1E}\color{white}\bfseries 1.03 & \cellcolor[HTML]{0B5D1E}\color{white}\bfseries 0.77 & \cellcolor[HTML]{0B5D1E}\color{white}\bfseries 0.51 & \cellcolor[HTML]{2E8B3E}\color{white}\bfseries 0.39 & \cellcolor[HTML]{2E8B3E}\color{white}\bfseries 1.65 & \cellcolor[HTML]{0B5D1E}\color{white}\bfseries 1.18 & \cellcolor[HTML]{0B5D1E}\color{white}\bfseries 0.91 & \cellcolor[HTML]{2E8B3E}\color{white}\bfseries 0.68 & \cellcolor[HTML]{2E8B3E}\color{white}\bfseries 0.47 & 0.39 \\
0.5 & \cellcolor[HTML]{CDEFD4}\bfseries 1.89 & \cellcolor[HTML]{2E8B3E}\color{white}\bfseries 1.36 & \cellcolor[HTML]{0B5D1E}\color{white}\bfseries 1.03 & \cellcolor[HTML]{0B5D1E}\color{white}\bfseries 0.77 & \cellcolor[HTML]{0B5D1E}\color{white}\bfseries 0.51 & \cellcolor[HTML]{0B5D1E}\color{white}\bfseries 0.38 & \cellcolor[HTML]{2E8B3E}\color{white}\bfseries 1.65 & \cellcolor[HTML]{0B5D1E}\color{white}\bfseries 1.18 & \cellcolor[HTML]{2E8B3E}\color{white}\bfseries 0.92 & \cellcolor[HTML]{0B5D1E}\color{white}\bfseries 0.67 & \cellcolor[HTML]{0B5D1E}\color{white}\bfseries 0.46 & \cellcolor[HTML]{2E8B3E}\color{white}\bfseries 0.36 \\
0.6 & 1.91 & \cellcolor[HTML]{CDEFD4}\bfseries 1.37 & \cellcolor[HTML]{2E8B3E}\color{white}\bfseries 1.05 & \cellcolor[HTML]{2E8B3E}\color{white}\bfseries 0.78 & \cellcolor[HTML]{2E8B3E}\color{white}\bfseries 0.52 & \cellcolor[HTML]{2E8B3E}\color{white}\bfseries 0.39 & \cellcolor[HTML]{CDEFD4}\bfseries 1.67 & \cellcolor[HTML]{2E8B3E}\color{white}\bfseries 1.19 & \cellcolor[HTML]{2E8B3E}\color{white}\bfseries 0.92 & \cellcolor[HTML]{2E8B3E}\color{white}\bfseries 0.68 & \cellcolor[HTML]{0B5D1E}\color{white}\bfseries 0.46 & \cellcolor[HTML]{0B5D1E}\color{white}\bfseries 0.35 \\
0.7 & 1.93 & 1.40 & \cellcolor[HTML]{CDEFD4}\bfseries 1.07 & 0.80 & 0.55 & \cellcolor[HTML]{CDEFD4}\bfseries 0.42 & 1.69 & 1.22 & \cellcolor[HTML]{CDEFD4}\bfseries 0.94 & \cellcolor[HTML]{CDEFD4}\bfseries 0.70 & \cellcolor[HTML]{CDEFD4}\bfseries 0.48 & \cellcolor[HTML]{CDEFD4}\bfseries 0.37 \\
0.8 & 1.98 & 1.46 & 1.12 & 0.85 & 0.60 & 0.48 & 1.72 & 1.27 & 0.98 & 0.74 & 0.53 & 0.43 \\
0.9 & 2.04 & 1.55 & 1.21 & 0.95 & 0.71 & 0.60 & 1.77 & 1.35 & 1.06 & 0.83 & 0.63 & 0.53 \\
0.95 & 2.08 & 1.62 & 1.27 & 1.02 & 0.79 & 0.69 & 1.80 & 1.41 & 1.11 & 0.89 & 0.70 & 0.61 \\
1 & 2.12 & 1.69 & 1.35 & 1.11 & 0.89 & 0.79 & 1.83 & 1.47 & 1.18 & 0.96 & 0.78 & 0.70 \\
\midrule
\multicolumn{13}{@{}l}{$N=2{,}000$ tasks} \\
0.1 & \cellcolor[HTML]{0B5D1E}\color{white}\bfseries 0.90 & \cellcolor[HTML]{2E8B3E}\color{white}\bfseries 0.67 & \cellcolor[HTML]{CDEFD4}\bfseries 0.53 & 0.44 & 0.41 & 0.41 & \cellcolor[HTML]{2E8B3E}\color{white}\bfseries 0.80 & 0.59 & 0.52 & 0.47 & 0.46 & 0.47 \\
0.2 & \cellcolor[HTML]{0B5D1E}\color{white}\bfseries 0.90 & \cellcolor[HTML]{0B5D1E}\color{white}\bfseries 0.66 & \cellcolor[HTML]{2E8B3E}\color{white}\bfseries 0.52 & \cellcolor[HTML]{0B5D1E}\color{white}\bfseries 0.39 & \cellcolor[HTML]{CDEFD4}\bfseries 0.29 & 0.26 & \cellcolor[HTML]{0B5D1E}\color{white}\bfseries 0.79 & \cellcolor[HTML]{2E8B3E}\color{white}\bfseries 0.57 & \cellcolor[HTML]{2E8B3E}\color{white}\bfseries 0.46 & 0.38 & 0.33 & 0.32 \\
0.3 & \cellcolor[HTML]{0B5D1E}\color{white}\bfseries 0.90 & \cellcolor[HTML]{0B5D1E}\color{white}\bfseries 0.66 & \cellcolor[HTML]{0B5D1E}\color{white}\bfseries 0.51 & \cellcolor[HTML]{0B5D1E}\color{white}\bfseries 0.39 & \cellcolor[HTML]{0B5D1E}\color{white}\bfseries 0.26 & \cellcolor[HTML]{2E8B3E}\color{white}\bfseries 0.20 & \cellcolor[HTML]{0B5D1E}\color{white}\bfseries 0.79 & \cellcolor[HTML]{0B5D1E}\color{white}\bfseries 0.56 & \cellcolor[HTML]{0B5D1E}\color{white}\bfseries 0.45 & \cellcolor[HTML]{0B5D1E}\color{white}\bfseries 0.34 & \cellcolor[HTML]{2E8B3E}\color{white}\bfseries 0.25 & 0.21 \\
0.4 & \cellcolor[HTML]{0B5D1E}\color{white}\bfseries 0.90 & \cellcolor[HTML]{0B5D1E}\color{white}\bfseries 0.66 & \cellcolor[HTML]{2E8B3E}\color{white}\bfseries 0.52 & \cellcolor[HTML]{0B5D1E}\color{white}\bfseries 0.39 & \cellcolor[HTML]{0B5D1E}\color{white}\bfseries 0.26 & \cellcolor[HTML]{0B5D1E}\color{white}\bfseries 0.19 & \cellcolor[HTML]{0B5D1E}\color{white}\bfseries 0.79 & \cellcolor[HTML]{0B5D1E}\color{white}\bfseries 0.56 & \cellcolor[HTML]{0B5D1E}\color{white}\bfseries 0.45 & \cellcolor[HTML]{0B5D1E}\color{white}\bfseries 0.34 & \cellcolor[HTML]{0B5D1E}\color{white}\bfseries 0.23 & \cellcolor[HTML]{2E8B3E}\color{white}\bfseries 0.18 \\
0.5 & \cellcolor[HTML]{0B5D1E}\color{white}\bfseries 0.90 & \cellcolor[HTML]{2E8B3E}\color{white}\bfseries 0.67 & \cellcolor[HTML]{2E8B3E}\color{white}\bfseries 0.52 & \cellcolor[HTML]{0B5D1E}\color{white}\bfseries 0.39 & \cellcolor[HTML]{0B5D1E}\color{white}\bfseries 0.26 & \cellcolor[HTML]{0B5D1E}\color{white}\bfseries 0.19 & \cellcolor[HTML]{2E8B3E}\color{white}\bfseries 0.80 & \cellcolor[HTML]{2E8B3E}\color{white}\bfseries 0.57 & \cellcolor[HTML]{0B5D1E}\color{white}\bfseries 0.45 & \cellcolor[HTML]{0B5D1E}\color{white}\bfseries 0.34 & \cellcolor[HTML]{0B5D1E}\color{white}\bfseries 0.23 & \cellcolor[HTML]{0B5D1E}\color{white}\bfseries 0.17 \\
0.6 & \cellcolor[HTML]{2E8B3E}\color{white}\bfseries 0.91 & \cellcolor[HTML]{2E8B3E}\color{white}\bfseries 0.67 & \cellcolor[HTML]{2E8B3E}\color{white}\bfseries 0.52 & \cellcolor[HTML]{0B5D1E}\color{white}\bfseries 0.39 & \cellcolor[HTML]{0B5D1E}\color{white}\bfseries 0.26 & \cellcolor[HTML]{0B5D1E}\color{white}\bfseries 0.19 & \cellcolor[HTML]{2E8B3E}\color{white}\bfseries 0.80 & \cellcolor[HTML]{2E8B3E}\color{white}\bfseries 0.57 & \cellcolor[HTML]{0B5D1E}\color{white}\bfseries 0.45 & \cellcolor[HTML]{0B5D1E}\color{white}\bfseries 0.34 & \cellcolor[HTML]{0B5D1E}\color{white}\bfseries 0.23 & \cellcolor[HTML]{0B5D1E}\color{white}\bfseries 0.17 \\
0.7 & \cellcolor[HTML]{CDEFD4}\bfseries 0.92 & \cellcolor[HTML]{CDEFD4}\bfseries 0.68 & \cellcolor[HTML]{CDEFD4}\bfseries 0.53 & \cellcolor[HTML]{2E8B3E}\color{white}\bfseries 0.40 & \cellcolor[HTML]{2E8B3E}\color{white}\bfseries 0.27 & \cellcolor[HTML]{2E8B3E}\color{white}\bfseries 0.20 & \cellcolor[HTML]{CDEFD4}\bfseries 0.81 & \cellcolor[HTML]{CDEFD4}\bfseries 0.58 & \cellcolor[HTML]{2E8B3E}\color{white}\bfseries 0.46 & \cellcolor[HTML]{2E8B3E}\color{white}\bfseries 0.35 & \cellcolor[HTML]{0B5D1E}\color{white}\bfseries 0.23 & \cellcolor[HTML]{2E8B3E}\color{white}\bfseries 0.18 \\
0.8 & 0.94 & 0.71 & 0.55 & \cellcolor[HTML]{CDEFD4}\bfseries 0.42 & \cellcolor[HTML]{CDEFD4}\bfseries 0.29 & \cellcolor[HTML]{CDEFD4}\bfseries 0.22 & 0.83 & 0.60 & \cellcolor[HTML]{CDEFD4}\bfseries 0.47 & \cellcolor[HTML]{CDEFD4}\bfseries 0.36 & \cellcolor[HTML]{2E8B3E}\color{white}\bfseries 0.25 & \cellcolor[HTML]{CDEFD4}\bfseries 0.20 \\
0.9 & 0.97 & 0.76 & 0.59 & 0.47 & 0.35 & 0.28 & 0.87 & 0.64 & 0.51 & 0.41 & \cellcolor[HTML]{CDEFD4}\bfseries 0.30 & 0.25 \\
0.95 & 1.00 & 0.80 & 0.63 & 0.51 & 0.39 & 0.33 & 0.89 & 0.67 & 0.54 & 0.44 & 0.34 & 0.29 \\
1 & 1.02 & 0.84 & 0.68 & 0.56 & 0.45 & 0.39 & 0.91 & 0.71 & 0.58 & 0.49 & 0.39 & 0.34 \\
\bottomrule
\end{tabular}

\floatnote{Entries are the RMSE in pp of the corrected scores over all $J$ positions and 1,000 synthetic benchmarks with independent Bernoulli outcomes. Every model has true mean $p$. Each correction uses $m=\max\{2,\lfloor N^\gamma\rfloor\}$ and $B=2{,}000$, and $\gamma=1$ is the standard bootstrap. All exponents use the same benchmarks (seed 20261103). In each column, bold type with the darkest, medium, and lightest shading marks the smallest, second-smallest, and third-smallest displayed values. Monte Carlo standard errors are at most 0.04 pp.}
\end{table}

With closely spaced models, error changes little across a wide range of exponents (Table~\ref{tab:gamma_table_close}). In every column, each exponent from 0.4 to 0.9 has an RMSE within 8\% of the most accurate exponent. The default is within 5\% of the most accurate exponent for $J\ge20$ and within 8\% for every $J$, while the standard bootstrap is 4--13\% above it.

\begin{table}[!t]\centering\small
\setlength{\tabcolsep}{4pt}
\caption{With closely spaced models, error is nearly flat across a wide range of exponents.}\label{tab:gamma_table_close}
\begin{tabular}{@{}c*{12}{c}@{}}
\toprule
 & \multicolumn{6}{c}{$p=0.5$} & \multicolumn{6}{c}{$p=0.75$} \\
\cmidrule(lr){2-7}\cmidrule(lr){8-13}
$\gamma\,/\,J$ & 2 & 5 & 10 & 20 & 50 & 100 & 2 & 5 & 10 & 20 & 50 & 100 \\
\midrule
\multicolumn{13}{@{}l}{$N=500$ tasks} \\
0.1 & \cellcolor[HTML]{0B5D1E}\color{white}\bfseries 2.30 & \cellcolor[HTML]{0B5D1E}\color{white}\bfseries 1.78 & \cellcolor[HTML]{0B5D1E}\color{white}\bfseries 1.54 & \cellcolor[HTML]{CDEFD4}\bfseries 1.42 & 1.39 & 1.40 & \cellcolor[HTML]{0B5D1E}\color{white}\bfseries 1.98 & \cellcolor[HTML]{0B5D1E}\color{white}\bfseries 1.56 & 1.39 & 1.33 & 1.31 & 1.32 \\
0.2 & \cellcolor[HTML]{2E8B3E}\color{white}\bfseries 2.31 & \cellcolor[HTML]{2E8B3E}\color{white}\bfseries 1.79 & \cellcolor[HTML]{0B5D1E}\color{white}\bfseries 1.54 & \cellcolor[HTML]{0B5D1E}\color{white}\bfseries 1.38 & 1.31 & 1.30 & \cellcolor[HTML]{2E8B3E}\color{white}\bfseries 1.99 & \cellcolor[HTML]{0B5D1E}\color{white}\bfseries 1.56 & \cellcolor[HTML]{CDEFD4}\bfseries 1.36 & 1.27 & 1.22 & 1.22 \\
0.3 & \cellcolor[HTML]{CDEFD4}\bfseries 2.32 & \cellcolor[HTML]{CDEFD4}\bfseries 1.80 & \cellcolor[HTML]{0B5D1E}\color{white}\bfseries 1.54 & \cellcolor[HTML]{0B5D1E}\color{white}\bfseries 1.38 & \cellcolor[HTML]{CDEFD4}\bfseries 1.27 & 1.23 & \cellcolor[HTML]{CDEFD4}\bfseries 2.01 & \cellcolor[HTML]{0B5D1E}\color{white}\bfseries 1.56 & \cellcolor[HTML]{0B5D1E}\color{white}\bfseries 1.34 & \cellcolor[HTML]{CDEFD4}\bfseries 1.22 & 1.13 & 1.11 \\
0.4 & 2.33 & \cellcolor[HTML]{CDEFD4}\bfseries 1.80 & \cellcolor[HTML]{0B5D1E}\color{white}\bfseries 1.54 & \cellcolor[HTML]{0B5D1E}\color{white}\bfseries 1.38 & \cellcolor[HTML]{2E8B3E}\color{white}\bfseries 1.26 & \cellcolor[HTML]{CDEFD4}\bfseries 1.21 & \cellcolor[HTML]{CDEFD4}\bfseries 2.01 & \cellcolor[HTML]{0B5D1E}\color{white}\bfseries 1.56 & \cellcolor[HTML]{0B5D1E}\color{white}\bfseries 1.34 & \cellcolor[HTML]{2E8B3E}\color{white}\bfseries 1.21 & \cellcolor[HTML]{CDEFD4}\bfseries 1.10 & 1.06 \\
0.5 & 2.34 & 1.81 & \cellcolor[HTML]{2E8B3E}\color{white}\bfseries 1.55 & \cellcolor[HTML]{0B5D1E}\color{white}\bfseries 1.38 & \cellcolor[HTML]{2E8B3E}\color{white}\bfseries 1.26 & \cellcolor[HTML]{2E8B3E}\color{white}\bfseries 1.20 & 2.02 & \cellcolor[HTML]{2E8B3E}\color{white}\bfseries 1.57 & \cellcolor[HTML]{0B5D1E}\color{white}\bfseries 1.34 & \cellcolor[HTML]{0B5D1E}\color{white}\bfseries 1.20 & \cellcolor[HTML]{2E8B3E}\color{white}\bfseries 1.09 & \cellcolor[HTML]{CDEFD4}\bfseries 1.05 \\
0.6 & 2.35 & 1.81 & \cellcolor[HTML]{2E8B3E}\color{white}\bfseries 1.55 & \cellcolor[HTML]{0B5D1E}\color{white}\bfseries 1.38 & \cellcolor[HTML]{0B5D1E}\color{white}\bfseries 1.25 & \cellcolor[HTML]{2E8B3E}\color{white}\bfseries 1.20 & 2.03 & \cellcolor[HTML]{2E8B3E}\color{white}\bfseries 1.57 & \cellcolor[HTML]{0B5D1E}\color{white}\bfseries 1.34 & \cellcolor[HTML]{0B5D1E}\color{white}\bfseries 1.20 & \cellcolor[HTML]{0B5D1E}\color{white}\bfseries 1.08 & \cellcolor[HTML]{2E8B3E}\color{white}\bfseries 1.04 \\
0.7 & 2.36 & 1.82 & \cellcolor[HTML]{2E8B3E}\color{white}\bfseries 1.55 & \cellcolor[HTML]{0B5D1E}\color{white}\bfseries 1.38 & \cellcolor[HTML]{0B5D1E}\color{white}\bfseries 1.25 & \cellcolor[HTML]{0B5D1E}\color{white}\bfseries 1.19 & 2.03 & \cellcolor[HTML]{CDEFD4}\bfseries 1.58 & \cellcolor[HTML]{2E8B3E}\color{white}\bfseries 1.35 & \cellcolor[HTML]{0B5D1E}\color{white}\bfseries 1.20 & \cellcolor[HTML]{0B5D1E}\color{white}\bfseries 1.08 & \cellcolor[HTML]{0B5D1E}\color{white}\bfseries 1.03 \\
0.8 & 2.38 & 1.85 & \cellcolor[HTML]{CDEFD4}\bfseries 1.57 & \cellcolor[HTML]{2E8B3E}\color{white}\bfseries 1.39 & \cellcolor[HTML]{0B5D1E}\color{white}\bfseries 1.25 & \cellcolor[HTML]{0B5D1E}\color{white}\bfseries 1.19 & 2.05 & 1.61 & \cellcolor[HTML]{CDEFD4}\bfseries 1.36 & \cellcolor[HTML]{CDEFD4}\bfseries 1.22 & \cellcolor[HTML]{2E8B3E}\color{white}\bfseries 1.09 & \cellcolor[HTML]{0B5D1E}\color{white}\bfseries 1.03 \\
0.9 & 2.40 & 1.91 & 1.62 & 1.43 & 1.28 & 1.22 & 2.06 & 1.67 & 1.40 & 1.25 & 1.12 & 1.06 \\
0.95 & 2.40 & 1.96 & 1.66 & 1.47 & 1.32 & 1.25 & 2.06 & 1.70 & 1.44 & 1.29 & 1.15 & 1.09 \\
1 & 2.39 & 2.01 & 1.71 & 1.52 & 1.37 & 1.30 & 2.06 & 1.75 & 1.48 & 1.33 & 1.19 & 1.13 \\
\midrule
\multicolumn{13}{@{}l}{$N=2{,}000$ tasks} \\
0.1 & \cellcolor[HTML]{0B5D1E}\color{white}\bfseries 1.13 & \cellcolor[HTML]{0B5D1E}\color{white}\bfseries 0.89 & \cellcolor[HTML]{0B5D1E}\color{white}\bfseries 0.77 & \cellcolor[HTML]{2E8B3E}\color{white}\bfseries 0.72 & 0.69 & 0.70 & \cellcolor[HTML]{0B5D1E}\color{white}\bfseries 0.99 & \cellcolor[HTML]{0B5D1E}\color{white}\bfseries 0.76 & \cellcolor[HTML]{CDEFD4}\bfseries 0.70 & 0.67 & 0.66 & 0.66 \\
0.2 & \cellcolor[HTML]{2E8B3E}\color{white}\bfseries 1.14 & \cellcolor[HTML]{0B5D1E}\color{white}\bfseries 0.89 & \cellcolor[HTML]{0B5D1E}\color{white}\bfseries 0.77 & \cellcolor[HTML]{0B5D1E}\color{white}\bfseries 0.70 & \cellcolor[HTML]{2E8B3E}\color{white}\bfseries 0.64 & 0.63 & \cellcolor[HTML]{2E8B3E}\color{white}\bfseries 1.00 & \cellcolor[HTML]{0B5D1E}\color{white}\bfseries 0.76 & \cellcolor[HTML]{0B5D1E}\color{white}\bfseries 0.67 & \cellcolor[HTML]{2E8B3E}\color{white}\bfseries 0.62 & 0.59 & 0.58 \\
0.3 & \cellcolor[HTML]{2E8B3E}\color{white}\bfseries 1.14 & \cellcolor[HTML]{0B5D1E}\color{white}\bfseries 0.89 & \cellcolor[HTML]{2E8B3E}\color{white}\bfseries 0.78 & \cellcolor[HTML]{0B5D1E}\color{white}\bfseries 0.70 & \cellcolor[HTML]{0B5D1E}\color{white}\bfseries 0.63 & \cellcolor[HTML]{CDEFD4}\bfseries 0.61 & \cellcolor[HTML]{CDEFD4}\bfseries 1.01 & \cellcolor[HTML]{0B5D1E}\color{white}\bfseries 0.76 & \cellcolor[HTML]{0B5D1E}\color{white}\bfseries 0.67 & \cellcolor[HTML]{0B5D1E}\color{white}\bfseries 0.60 & \cellcolor[HTML]{2E8B3E}\color{white}\bfseries 0.55 & \cellcolor[HTML]{CDEFD4}\bfseries 0.54 \\
0.4 & \cellcolor[HTML]{CDEFD4}\bfseries 1.15 & \cellcolor[HTML]{0B5D1E}\color{white}\bfseries 0.89 & \cellcolor[HTML]{2E8B3E}\color{white}\bfseries 0.78 & \cellcolor[HTML]{0B5D1E}\color{white}\bfseries 0.70 & \cellcolor[HTML]{0B5D1E}\color{white}\bfseries 0.63 & \cellcolor[HTML]{CDEFD4}\bfseries 0.61 & \cellcolor[HTML]{CDEFD4}\bfseries 1.01 & \cellcolor[HTML]{0B5D1E}\color{white}\bfseries 0.76 & \cellcolor[HTML]{0B5D1E}\color{white}\bfseries 0.67 & \cellcolor[HTML]{0B5D1E}\color{white}\bfseries 0.60 & \cellcolor[HTML]{2E8B3E}\color{white}\bfseries 0.55 & \cellcolor[HTML]{2E8B3E}\color{white}\bfseries 0.53 \\
0.5 & \cellcolor[HTML]{CDEFD4}\bfseries 1.15 & \cellcolor[HTML]{2E8B3E}\color{white}\bfseries 0.90 & \cellcolor[HTML]{2E8B3E}\color{white}\bfseries 0.78 & \cellcolor[HTML]{0B5D1E}\color{white}\bfseries 0.70 & \cellcolor[HTML]{0B5D1E}\color{white}\bfseries 0.63 & \cellcolor[HTML]{2E8B3E}\color{white}\bfseries 0.60 & \cellcolor[HTML]{CDEFD4}\bfseries 1.01 & \cellcolor[HTML]{0B5D1E}\color{white}\bfseries 0.76 & \cellcolor[HTML]{0B5D1E}\color{white}\bfseries 0.67 & \cellcolor[HTML]{0B5D1E}\color{white}\bfseries 0.60 & \cellcolor[HTML]{2E8B3E}\color{white}\bfseries 0.55 & \cellcolor[HTML]{2E8B3E}\color{white}\bfseries 0.53 \\
0.6 & \cellcolor[HTML]{CDEFD4}\bfseries 1.15 & \cellcolor[HTML]{2E8B3E}\color{white}\bfseries 0.90 & \cellcolor[HTML]{2E8B3E}\color{white}\bfseries 0.78 & \cellcolor[HTML]{0B5D1E}\color{white}\bfseries 0.70 & \cellcolor[HTML]{0B5D1E}\color{white}\bfseries 0.63 & \cellcolor[HTML]{2E8B3E}\color{white}\bfseries 0.60 & 1.02 & \cellcolor[HTML]{2E8B3E}\color{white}\bfseries 0.77 & \cellcolor[HTML]{0B5D1E}\color{white}\bfseries 0.67 & \cellcolor[HTML]{0B5D1E}\color{white}\bfseries 0.60 & \cellcolor[HTML]{2E8B3E}\color{white}\bfseries 0.55 & \cellcolor[HTML]{0B5D1E}\color{white}\bfseries 0.52 \\
0.7 & 1.16 & \cellcolor[HTML]{2E8B3E}\color{white}\bfseries 0.90 & \cellcolor[HTML]{2E8B3E}\color{white}\bfseries 0.78 & \cellcolor[HTML]{0B5D1E}\color{white}\bfseries 0.70 & \cellcolor[HTML]{0B5D1E}\color{white}\bfseries 0.63 & \cellcolor[HTML]{2E8B3E}\color{white}\bfseries 0.60 & 1.02 & \cellcolor[HTML]{2E8B3E}\color{white}\bfseries 0.77 & \cellcolor[HTML]{0B5D1E}\color{white}\bfseries 0.67 & \cellcolor[HTML]{0B5D1E}\color{white}\bfseries 0.60 & \cellcolor[HTML]{0B5D1E}\color{white}\bfseries 0.54 & \cellcolor[HTML]{0B5D1E}\color{white}\bfseries 0.52 \\
0.8 & 1.17 & \cellcolor[HTML]{CDEFD4}\bfseries 0.92 & \cellcolor[HTML]{2E8B3E}\color{white}\bfseries 0.78 & \cellcolor[HTML]{0B5D1E}\color{white}\bfseries 0.70 & \cellcolor[HTML]{0B5D1E}\color{white}\bfseries 0.63 & \cellcolor[HTML]{0B5D1E}\color{white}\bfseries 0.59 & 1.04 & \cellcolor[HTML]{CDEFD4}\bfseries 0.78 & \cellcolor[HTML]{0B5D1E}\color{white}\bfseries 0.67 & \cellcolor[HTML]{0B5D1E}\color{white}\bfseries 0.60 & \cellcolor[HTML]{0B5D1E}\color{white}\bfseries 0.54 & \cellcolor[HTML]{0B5D1E}\color{white}\bfseries 0.52 \\
0.9 & 1.18 & 0.95 & \cellcolor[HTML]{CDEFD4}\bfseries 0.81 & \cellcolor[HTML]{2E8B3E}\color{white}\bfseries 0.72 & \cellcolor[HTML]{2E8B3E}\color{white}\bfseries 0.64 & \cellcolor[HTML]{2E8B3E}\color{white}\bfseries 0.60 & 1.05 & 0.81 & \cellcolor[HTML]{2E8B3E}\color{white}\bfseries 0.69 & \cellcolor[HTML]{2E8B3E}\color{white}\bfseries 0.62 & \cellcolor[HTML]{2E8B3E}\color{white}\bfseries 0.55 & \cellcolor[HTML]{2E8B3E}\color{white}\bfseries 0.53 \\
0.95 & 1.18 & 0.97 & 0.83 & \cellcolor[HTML]{CDEFD4}\bfseries 0.74 & \cellcolor[HTML]{CDEFD4}\bfseries 0.66 & 0.62 & 1.05 & 0.83 & 0.71 & \cellcolor[HTML]{CDEFD4}\bfseries 0.64 & \cellcolor[HTML]{CDEFD4}\bfseries 0.57 & \cellcolor[HTML]{CDEFD4}\bfseries 0.54 \\
1 & 1.17 & 1.00 & 0.86 & 0.77 & 0.69 & 0.65 & 1.05 & 0.86 & 0.74 & 0.67 & 0.59 & 0.56 \\
\bottomrule
\end{tabular}

\floatnote{As in Table~\ref{tab:gamma_table_tied}, except that the true means are evenly spaced over $2\sigma$ around $p$, with $\sigma=\sqrt{p(1-p)/N}$.}
\end{table}

When the means span a wide range, larger exponents are more accurate (Table~\ref{tab:gamma_table_spread}). The most accurate exponents lie between 0.7 and 1, and the default is within 5\% of the best. Small exponents blur genuine gaps between models, as in the two-model example in Appendix~\ref{app:toy}.

\begin{table}[!t]\centering\small
\setlength{\tabcolsep}{4pt}
\caption{With well-separated models, large exponents minimize error.}\label{tab:gamma_table_spread}
\begin{tabular}{@{}c*{12}{c}@{}}
\toprule
 & \multicolumn{6}{c}{$p=0.5$} & \multicolumn{6}{c}{$p=0.75$} \\
\cmidrule(lr){2-7}\cmidrule(lr){8-13}
$\gamma\,/\,J$ & 2 & 5 & 10 & 20 & 50 & 100 & 2 & 5 & 10 & 20 & 50 & 100 \\
\midrule
\multicolumn{13}{@{}l}{$N=500$ tasks} \\
0.1 & 2.47 & 2.81 & 2.81 & 2.64 & 2.67 & 2.69 & 2.11 & 2.44 & 2.38 & 2.23 & 2.23 & 2.26 \\
0.2 & 2.43 & 2.80 & 2.80 & 2.61 & 2.61 & 2.62 & 2.08 & 2.44 & 2.39 & 2.22 & 2.22 & 2.24 \\
0.3 & 2.32 & 2.73 & 2.74 & 2.57 & 2.55 & 2.55 & 1.97 & 2.39 & 2.35 & 2.20 & 2.18 & 2.19 \\
0.4 & 2.20 & 2.62 & 2.63 & 2.49 & 2.47 & 2.46 & 1.88 & 2.29 & 2.25 & 2.13 & 2.11 & 2.12 \\
0.5 & \cellcolor[HTML]{CDEFD4}\bfseries 2.13 & 2.54 & 2.52 & 2.41 & 2.38 & 2.36 & \cellcolor[HTML]{CDEFD4}\bfseries 1.82 & 2.23 & 2.15 & 2.06 & 2.04 & 2.04 \\
0.6 & \cellcolor[HTML]{2E8B3E}\color{white}\bfseries 2.09 & 2.52 & \cellcolor[HTML]{2E8B3E}\color{white}\bfseries 2.44 & \cellcolor[HTML]{2E8B3E}\color{white}\bfseries 2.33 & 2.28 & 2.26 & \cellcolor[HTML]{2E8B3E}\color{white}\bfseries 1.79 & 2.21 & \cellcolor[HTML]{2E8B3E}\color{white}\bfseries 2.08 & \cellcolor[HTML]{2E8B3E}\color{white}\bfseries 2.00 & \cellcolor[HTML]{CDEFD4}\bfseries 1.96 & 1.95 \\
0.7 & \cellcolor[HTML]{0B5D1E}\color{white}\bfseries 2.08 & 2.53 & \cellcolor[HTML]{0B5D1E}\color{white}\bfseries 2.42 & \cellcolor[HTML]{0B5D1E}\color{white}\bfseries 2.29 & 2.22 & \cellcolor[HTML]{CDEFD4}\bfseries 2.19 & \cellcolor[HTML]{0B5D1E}\color{white}\bfseries 1.78 & 2.22 & \cellcolor[HTML]{0B5D1E}\color{white}\bfseries 2.06 & \cellcolor[HTML]{0B5D1E}\color{white}\bfseries 1.97 & \cellcolor[HTML]{2E8B3E}\color{white}\bfseries 1.91 & \cellcolor[HTML]{CDEFD4}\bfseries 1.89 \\
0.8 & \cellcolor[HTML]{0B5D1E}\color{white}\bfseries 2.08 & 2.53 & \cellcolor[HTML]{CDEFD4}\bfseries 2.46 & \cellcolor[HTML]{0B5D1E}\color{white}\bfseries 2.29 & \cellcolor[HTML]{0B5D1E}\color{white}\bfseries 2.19 & \cellcolor[HTML]{2E8B3E}\color{white}\bfseries 2.15 & \cellcolor[HTML]{0B5D1E}\color{white}\bfseries 1.78 & 2.22 & \cellcolor[HTML]{CDEFD4}\bfseries 2.09 & \cellcolor[HTML]{0B5D1E}\color{white}\bfseries 1.97 & \cellcolor[HTML]{0B5D1E}\color{white}\bfseries 1.89 & \cellcolor[HTML]{2E8B3E}\color{white}\bfseries 1.86 \\
0.9 & \cellcolor[HTML]{0B5D1E}\color{white}\bfseries 2.08 & \cellcolor[HTML]{CDEFD4}\bfseries 2.48 & 2.50 & \cellcolor[HTML]{2E8B3E}\color{white}\bfseries 2.33 & \cellcolor[HTML]{0B5D1E}\color{white}\bfseries 2.19 & \cellcolor[HTML]{0B5D1E}\color{white}\bfseries 2.14 & \cellcolor[HTML]{0B5D1E}\color{white}\bfseries 1.78 & \cellcolor[HTML]{CDEFD4}\bfseries 2.17 & 2.13 & \cellcolor[HTML]{2E8B3E}\color{white}\bfseries 2.00 & \cellcolor[HTML]{0B5D1E}\color{white}\bfseries 1.89 & \cellcolor[HTML]{0B5D1E}\color{white}\bfseries 1.85 \\
0.95 & \cellcolor[HTML]{0B5D1E}\color{white}\bfseries 2.08 & \cellcolor[HTML]{2E8B3E}\color{white}\bfseries 2.43 & 2.51 & \cellcolor[HTML]{CDEFD4}\bfseries 2.35 & \cellcolor[HTML]{2E8B3E}\color{white}\bfseries 2.20 & \cellcolor[HTML]{0B5D1E}\color{white}\bfseries 2.14 & \cellcolor[HTML]{0B5D1E}\color{white}\bfseries 1.78 & \cellcolor[HTML]{2E8B3E}\color{white}\bfseries 2.13 & 2.14 & \cellcolor[HTML]{CDEFD4}\bfseries 2.02 & \cellcolor[HTML]{0B5D1E}\color{white}\bfseries 1.89 & \cellcolor[HTML]{0B5D1E}\color{white}\bfseries 1.85 \\
1 & \cellcolor[HTML]{0B5D1E}\color{white}\bfseries 2.08 & \cellcolor[HTML]{0B5D1E}\color{white}\bfseries 2.38 & 2.51 & 2.38 & \cellcolor[HTML]{CDEFD4}\bfseries 2.21 & \cellcolor[HTML]{2E8B3E}\color{white}\bfseries 2.15 & \cellcolor[HTML]{0B5D1E}\color{white}\bfseries 1.78 & \cellcolor[HTML]{0B5D1E}\color{white}\bfseries 2.08 & 2.13 & 2.04 & \cellcolor[HTML]{2E8B3E}\color{white}\bfseries 1.91 & \cellcolor[HTML]{2E8B3E}\color{white}\bfseries 1.86 \\
\midrule
\multicolumn{13}{@{}l}{$N=2{,}000$ tasks} \\
0.1 & 1.29 & 1.41 & 1.40 & 1.35 & 1.37 & 1.37 & 1.12 & 1.20 & 1.20 & 1.16 & 1.16 & 1.17 \\
0.2 & 1.29 & 1.42 & 1.41 & 1.35 & 1.34 & 1.34 & 1.13 & 1.22 & 1.21 & 1.17 & 1.16 & 1.16 \\
0.3 & 1.26 & 1.40 & 1.40 & 1.34 & 1.33 & 1.32 & 1.11 & 1.21 & 1.21 & 1.16 & 1.14 & 1.14 \\
0.4 & 1.21 & 1.37 & 1.37 & 1.32 & 1.31 & 1.30 & 1.06 & 1.18 & 1.18 & 1.14 & 1.12 & 1.12 \\
0.5 & \cellcolor[HTML]{CDEFD4}\bfseries 1.16 & 1.32 & 1.32 & 1.27 & 1.26 & 1.25 & \cellcolor[HTML]{CDEFD4}\bfseries 1.01 & 1.14 & 1.14 & 1.10 & 1.08 & 1.08 \\
0.6 & \cellcolor[HTML]{2E8B3E}\color{white}\bfseries 1.12 & \cellcolor[HTML]{CDEFD4}\bfseries 1.28 & \cellcolor[HTML]{CDEFD4}\bfseries 1.26 & 1.21 & 1.20 & 1.19 & \cellcolor[HTML]{2E8B3E}\color{white}\bfseries 0.97 & \cellcolor[HTML]{CDEFD4}\bfseries 1.10 & \cellcolor[HTML]{CDEFD4}\bfseries 1.08 & 1.05 & 1.03 & \cellcolor[HTML]{CDEFD4}\bfseries 1.03 \\
0.7 & \cellcolor[HTML]{0B5D1E}\color{white}\bfseries 1.09 & 1.29 & \cellcolor[HTML]{0B5D1E}\color{white}\bfseries 1.22 & \cellcolor[HTML]{2E8B3E}\color{white}\bfseries 1.17 & \cellcolor[HTML]{CDEFD4}\bfseries 1.14 & \cellcolor[HTML]{CDEFD4}\bfseries 1.13 & \cellcolor[HTML]{0B5D1E}\color{white}\bfseries 0.95 & 1.11 & \cellcolor[HTML]{0B5D1E}\color{white}\bfseries 1.05 & \cellcolor[HTML]{2E8B3E}\color{white}\bfseries 1.01 & \cellcolor[HTML]{CDEFD4}\bfseries 0.98 & \cellcolor[HTML]{2E8B3E}\color{white}\bfseries 0.98 \\
0.8 & \cellcolor[HTML]{0B5D1E}\color{white}\bfseries 1.09 & 1.30 & \cellcolor[HTML]{2E8B3E}\color{white}\bfseries 1.23 & \cellcolor[HTML]{0B5D1E}\color{white}\bfseries 1.16 & \cellcolor[HTML]{0B5D1E}\color{white}\bfseries 1.11 & \cellcolor[HTML]{2E8B3E}\color{white}\bfseries 1.09 & \cellcolor[HTML]{0B5D1E}\color{white}\bfseries 0.95 & 1.12 & \cellcolor[HTML]{2E8B3E}\color{white}\bfseries 1.06 & \cellcolor[HTML]{0B5D1E}\color{white}\bfseries 1.00 & \cellcolor[HTML]{0B5D1E}\color{white}\bfseries 0.96 & \cellcolor[HTML]{0B5D1E}\color{white}\bfseries 0.94 \\
0.9 & \cellcolor[HTML]{0B5D1E}\color{white}\bfseries 1.09 & 1.29 & \cellcolor[HTML]{CDEFD4}\bfseries 1.26 & \cellcolor[HTML]{2E8B3E}\color{white}\bfseries 1.17 & \cellcolor[HTML]{0B5D1E}\color{white}\bfseries 1.11 & \cellcolor[HTML]{0B5D1E}\color{white}\bfseries 1.08 & \cellcolor[HTML]{0B5D1E}\color{white}\bfseries 0.95 & \cellcolor[HTML]{CDEFD4}\bfseries 1.10 & \cellcolor[HTML]{CDEFD4}\bfseries 1.08 & \cellcolor[HTML]{2E8B3E}\color{white}\bfseries 1.01 & \cellcolor[HTML]{0B5D1E}\color{white}\bfseries 0.96 & \cellcolor[HTML]{0B5D1E}\color{white}\bfseries 0.94 \\
0.95 & \cellcolor[HTML]{0B5D1E}\color{white}\bfseries 1.09 & \cellcolor[HTML]{2E8B3E}\color{white}\bfseries 1.26 & 1.27 & \cellcolor[HTML]{CDEFD4}\bfseries 1.19 & \cellcolor[HTML]{0B5D1E}\color{white}\bfseries 1.11 & \cellcolor[HTML]{0B5D1E}\color{white}\bfseries 1.08 & \cellcolor[HTML]{0B5D1E}\color{white}\bfseries 0.95 & \cellcolor[HTML]{2E8B3E}\color{white}\bfseries 1.08 & 1.09 & \cellcolor[HTML]{CDEFD4}\bfseries 1.02 & \cellcolor[HTML]{0B5D1E}\color{white}\bfseries 0.96 & \cellcolor[HTML]{0B5D1E}\color{white}\bfseries 0.94 \\
1 & \cellcolor[HTML]{0B5D1E}\color{white}\bfseries 1.09 & \cellcolor[HTML]{0B5D1E}\color{white}\bfseries 1.23 & \cellcolor[HTML]{CDEFD4}\bfseries 1.26 & 1.20 & \cellcolor[HTML]{2E8B3E}\color{white}\bfseries 1.12 & \cellcolor[HTML]{0B5D1E}\color{white}\bfseries 1.08 & \cellcolor[HTML]{0B5D1E}\color{white}\bfseries 0.95 & \cellcolor[HTML]{0B5D1E}\color{white}\bfseries 1.05 & 1.09 & 1.04 & \cellcolor[HTML]{2E8B3E}\color{white}\bfseries 0.97 & \cellcolor[HTML]{0B5D1E}\color{white}\bfseries 0.94 \\
\bottomrule
\end{tabular}

\floatnote{As in Table~\ref{tab:gamma_table_tied}, except that the true means are evenly spaced over $15\sigma$ around $p$, with $\sigma=\sqrt{p(1-p)/N}$.}
\end{table}

The most accurate exponents depend mainly on $J$ and the spacing of model abilities. For each $J$ and design, the shaded exponents overlap across all 4 combinations of task count and success rate in the grid. The default $\gamma=0.9$ is within 8\% of the most accurate exponent in the close and spread designs. Smaller exponents improve on the default by more than 8\% only in the tied design, where the gain grows with $J$. An exponent near the default therefore suits leaderboards whose models range from closely to widely spaced, and smaller exponents pay off mainly when many models are tied.

The spacing of a real leaderboard can be gauged by dividing the range of observed scores by $\sqrt{\bar p(1-\bar p)/N}$, where $\bar p$ is the mean score. Sampling noise widens observed ranges, so this gauge overstates the true spacing. Even 10 tied models have an expected observed range of about 3 standard errors. The designs draw models independently. When outcomes on a task are positively correlated across models, score differences are less noisy than this gauge assumes, so it understates the separation. Measured this way, the 14 audited benchmarks span 2.7--51.4 standard errors (median 14.8). Their 10 leading models, or all models when there are fewer, span 0.6--14.3 standard errors and lie within 3 standard errors in 6 benchmarks. In these 6 benchmarks, the leading groups are as closely spaced as the tied or close design. Appendix~\ref{app:mn_exponent_sensitivity} examines the audited corrections over the exponents from 0.4 to 0.9.

\section{Proofs of the Main Results}\label{app:proofs}

On a high level, to prove the correctness of the bootstrap procedure for all positions, we use a joint conditional central limit theorem to provide the common approximation for score bias and rank probabilities. 
Since positive gaps separate the population tiers with probabilities approaching one, the key arguments concern the ordering of Gaussian score errors within each tier. 
Below we first remark on the notations and then build the proofs step by step in this section.

Let $\mathcal F_N=\sigma(\boldsymbol Y_{1\cdot},\ldots,\boldsymbol Y_{N\cdot})$
denote the information in the observed sample. Given $\mathcal F_N$, the boostrap draw
$m$ vectors $\boldsymbol Y^*_{1\cdot},\ldots,\boldsymbol Y^*_{m\cdot}$ independently
with replacement from the $N$ observed rows, and set
\[
  \hat{\boldsymbol\tau}^{\,*}
    =\frac1m\sum_{r=1}^m\mathbf Y^*_{r\cdot},\qquad
  \boldsymbol{W}_N=\sqrt N(\hat{\boldsymbol\tau}-\boldsymbol\tau),\qquad
  \boldsymbol{W}_m^*=\sqrt m(\hat{\boldsymbol\tau}^{\,*}-\hat{\boldsymbol\tau}).
\]
Write $\E_*$ and $\Pr_*$
for expectation and probability conditional on
$\mathcal F_N$. 
For a vector
$\boldsymbol{x}\in\R^J$, write ${x}_{\ell,(r)}$ for its $r$-th largest coordinate
among the indices in $S_\ell$. In particular, ${Z}_{\ell,(r)}$ is this
order statistic of the Gaussian vector $\boldsymbol{Z}\sim\mathcal N(\boldsymbol{0},\boldsymbol{\Sigma})$ and $W_{N, \ell, (r)}$ ($W^*_{m, \ell, (r)}$, resp.) is the $r$-th largest component of those coordinates of $\boldsymbol{W}_N$ ($\boldsymbol{W}^*_m$, resp.) in $S_{\ell}$. 
The symbols
$\|\cdot\|_2$ and $\|\cdot\|_\infty$ denote the Euclidean and maximum
norms, respectively.

\subsection{Joint normal approximation and moment bounds}
\begin{lemma}[Joint CLT, bootstrap CLT, and moment bounds]\label{lem:clt_bootstrap_clt}
It holds that $\boldsymbol{W}_N=\sqrt N(\hat{\boldsymbol\tau}-\boldsymbol\tau)\Rightarrow \boldsymbol{Z}\sim\mathcal N(\boldsymbol{0},\boldsymbol{\Sigma})$. For an independent empirical $m$-resample, $\boldsymbol{W}_m^*=\sqrt m(\hat{\boldsymbol\tau}^*-\hat{\boldsymbol\tau})\Rightarrow \boldsymbol{Z}$ in probability as $m,N\to\infty$. Moreover $\E\|\boldsymbol{W}_N\|_\infty^2\leq J/4$ and $\E_*\|\boldsymbol{W}_m^*\|_\infty^2\leq J/4$.
\end{lemma}

\begin{proof}
The proof of original-sample central limit theorem is standard and 
follows from Cram\'er--Wold theorem. 
We next show that $\boldsymbol{W}^*_m$ converges in conditional distribution almost surely, which implies conditional convergence in distribution in probability. 

\vspace{2mm}
\noindent
{\textit{1. Almost sure convergence of distribution.}}
Define the empirical covariance matrix by
\begin{align}
  \hat{\boldsymbol{\Sigma}}
  &:=\frac1N\sum_{i=1}^N
      (\mathbf Y_{i\cdot}-\hat{\boldsymbol\tau})
      (\mathbf Y_{i\cdot}-\hat{\boldsymbol\tau})^\top=\frac1N\sum_{i=1}^N
      \mathbf Y_{i\cdot}\mathbf Y_{i\cdot}^\top
      -\hat{\boldsymbol\tau}\hat{\boldsymbol\tau}^{\top}.
      \label{eq:draft-empirical-covariance}
\end{align}
By the strong law of large numbers, 
$\hat{\boldsymbol{\tau}} \rightarrow \boldsymbol{\tau}$ and 
$\hat{\boldsymbol{\Sigma}}\to\boldsymbol{\Sigma}$ on an $\Omega_0$ which is of probability one.

Fix $\omega\in\Omega_0$. Note that the observed rows are fixed for fixed $N$ and any $\omega$. Write $\E^*_{\omega}$ and
$\operatorname{Cov}^*_{\omega}$ for expectation and covariance under
bootstrap sampling. 
To highlight the dependence on $N$ and $\omega$, below we write $m_N$, $\hat{\boldsymbol\tau}_N(\omega)$ and $\hat{\boldsymbol{\Sigma}}_N(\omega)$ instead of $m$, $\hat{\boldsymbol\tau}$ and $\hat{\boldsymbol{\Sigma}}$. 
For $1\leq r\leq m_N$, define
the centered, normalized bootstrap row
\[
  \boldsymbol{U}^*_{N,r}
    :=\frac{\mathbf Y^*_{r\cdot}
               -\hat{\boldsymbol\tau}_N(\omega)}{\sqrt{m_N}}.
\]
Within each row of this triangular array, the vectors $\boldsymbol{U}^*_{N,r}$ are
independent. Since a bootstrap row is uniform on the observed rows,
\[
  \E^*_{\omega}[\mathbf Y^*_{r\cdot}]
     =\hat{\boldsymbol\tau}(\omega),\qquad
  \operatorname{Cov}^*_{\omega}(\mathbf Y^*_{r\cdot})
     =\hat{\boldsymbol{\Sigma}}_N(\omega).
\]
Consequently,
\begin{equation}\label{eq:draft-array-covariance}
  \E^*_{\omega}[\boldsymbol{U}^*_{N,r}]=\boldsymbol{0},
  \qquad
  \sum_{r=1}^{m_N}\operatorname{Cov}^*_{\omega}(\boldsymbol{U}^*_{N,r})
     =\hat{\boldsymbol{\Sigma}}_N(\omega)\longrightarrow\boldsymbol{\Sigma}.
\end{equation}

Both a bootstrap row and the empirical mean belong to $[0,1]^J$.
Thus every possible bootstrap outcome satisfies
\[
  \|\boldsymbol{U}^*_{N,r}\|_2
     \leq\frac{\sqrt J}{\sqrt{m_N}}.
\]
For any $\varepsilon>0$, the assumption $m_N\to\infty$ implies that
$m_N>J/\varepsilon^2$ for all sufficiently large $N$. For those $N$,
the Lindeberg sum is exactly zero:
\begin{equation}\label{eq:draft-lindeberg}
  \sum_{r=1}^{m_N}\E^*_{\omega}\!\left[
     \|\boldsymbol{U}^*_{N,r}\|_2^2\,
     \mathbbm{1}\{\|\boldsymbol{U}^*_{N,r}\|_2>\varepsilon\}
  \right]=0.
\end{equation}
Equations \eqref{eq:draft-array-covariance} and
\eqref{eq:draft-lindeberg} verify the hypotheses of the ordinary
multivariate Lindeberg--Feller theorem. Applying that theorem under
the bootstrap law for this fixed data sequence gives that on $\Omega_0$: 
\[
  \sum_{r=1}^{m_N}\boldsymbol{U}^*_{N,r}=\boldsymbol{W}_{m_N}^*
    \Rightarrow\mathcal N(\boldsymbol{0},\boldsymbol{\Sigma}).
\]
\noindent
{\textit{2. Second moment bounds.}}
Finally, for any random variable $X\in[0,1]$, the inequality $X^2\leq X$ gives
\[
  \operatorname{Var}(X)
    =\E[X^2]-(\E X)^2
    \leq(\E X)(1-\E X)\leq\frac14.
\]
Independence of the original task rows implies, for each coordinate $j$,
\[
  \E[W_{N,j}^2]
    =\operatorname{Var}(Y_{1j})\leq\frac14.
\]
Conditional independence of the bootstrap rows similarly gives
\begin{align*}
  \E_*[(W_{m,j}^*)^2]
    &=\frac1N\sum_{i=1}^N(Y_{ij}-\hat\tau_j)^2=\frac1N\sum_{i=1}^NY_{ij}^2-\hat\tau_j^2
      \leq\hat\tau_j(1-\hat\tau_j)\leq\frac14.
\end{align*}
Upper bounding the maximum by the sum allows us to conclude the proof. 
\end{proof}

\subsection{Bias at each leaderboard position}

\begin{proof}[Proof of Proposition~\ref{prop:bias_expansion}]
We first control mistakes in the ordering between different population
tiers. We then take expectations of the within-tier order statistic.

Let $M_N=\|\boldsymbol{W}_N\|_\infty$ denote the largest absolute scaled score error.
For $L>1$, define the event
\[
  A_N=\left\{
    \max_{1\leq j\leq J}|\hat\tau_j-\tau_j|<\Delta/4
  \right\}.
\]
On $A_N$, if $j\in S_h$, $j'\in S_{h'}$, and $h<h'$, then
\[
  \hat\tau_j-\hat\tau_{j'}
    >\tau_j-\tau_{j'}-\Delta/2
    \geq\Delta/2>0.
\]
Thus every observed score in a higher tier exceeds every observed score
in a lower tier. By Markov's inequality and Lemma \ref{lem:clt_bootstrap_clt},
\begin{equation}\label{eq:draft-tier-probability}
  \Pr(A_N^c)
    =\Pr\{M_N\geq\sqrt N\,\Delta/4\}
    \leq\frac{16\E M_N^2}{N\Delta^2}
    \leq\frac{4J}{N\Delta^2}\longrightarrow0.
\end{equation}
When $L=1$, take $A_N$ to be the whole sample space. No gap or
cross-tier probability bound is needed in that case.

Let us write
\[
  T_{N,k}
    :=\sqrt N\bigl(\hat\tau_{\hat\pi(k)}
                       -\tau_{\hat\pi(k)}\bigr)
     =W_{N,\hat\pi(k)},
  \qquad V_N:=W_{N,\ell,(r)}.
\]
On $A_N$, the index $\hat\pi(k)$ belongs to $S_\ell$ and occupies
position $r$ within that tier. Every model in $S_\ell$ has the same
population mean. Sorting its empirical means is therefore equivalent to
sorting the coordinates of $W_N$ in that tier. Hence $T_{N,k}=V_N$ on $A_N,$
including when the empirical means tie. Both $|T_{N,k}|$ and $|V_N|$
are bounded by $M_N$. Cauchy--Schwarz gives
\begin{equation}\label{eq:draft-original-selection-error}
  \E|T_{N,k}-V_N|
    \leq2\E[M_N\mathbbm{1}\{A_N^c\}]
    \leq2\sqrt{\E M_N^2\,\Pr(A_N^c)}
    \leq\sqrt J\,\sqrt{\Pr(A_N^c)}\longrightarrow0.
\end{equation}

Taking an order
statistic is continuous in the maximum norm. Lemma~\ref{lem:clt_bootstrap_clt} and the continuous
mapping theorem yield
\[
  V_N\Rightarrow Z_{\ell,(r)}.
\]
Finally, we notice that 
$V_N^2 \leq \|W_N\|_\infty^2$ so $\E[V_N^2] \leq J/4$ for all $N$.
A uniform bound on the second moment implied uniform integrability, so $\E[V_N] \rightarrow \E[Z_{\ell, (r)}] = \delta_{\ell r}$. 
Combining with \eqref{eq:draft-original-selection-error}, we  concludes the proof. 
\end{proof}

\subsection{Bootstrap score correction}
The following deterministic fact will control changes in the score of a selected
model when the ranking is perturbed.

\begin{lemma}[Ordering fact]
Let $q$ be a positive integer, and let $\boldsymbol{x},\boldsymbol{d}\in\R^q$.
Write $x_{(r)}$ for the $r$-th largest coordinate of $\boldsymbol{x}$, for
$1\leq r\leq q$. Put $\boldsymbol{y}=\boldsymbol{x}+\boldsymbol{d}$, and let $\rho$ be any
permutation such that $y_{\rho(1)}\geq\cdots\geq y_{\rho(q)}$. Then
\begin{equation}\label{eq:draft-order-lipschitz}
  |y_{(r)}-x_{(r)}|\leq\|\boldsymbol{d}\|_\infty,
\end{equation}
and
\begin{equation}\label{eq:draft-selected-coordinate}
  |x_{\rho(r)}-x_{(r)}|\leq2\|\boldsymbol{d}\|_\infty.
\end{equation}
Both conclusions hold for every way of ordering tied coordinates.
\end{lemma}

\begin{proof}
Set $\varepsilon=\|\boldsymbol{d}\|_\infty$. At least $r$ coordinates of $\boldsymbol{x}$
are at least $x_{(r)}$. The corresponding coordinates of $\boldsymbol{y}$ are at least
$x_{(r)}-\varepsilon$, so $y_{(r)}\geq x_{(r)}-\varepsilon$. Also, at least
$q-r+1$ coordinates of $\boldsymbol{x}$ are at most $x_{(r)}$. Their coordinates in
$\boldsymbol{y}$ are at most $x_{(r)}+\varepsilon$, so
$y_{(r)}\leq x_{(r)}+\varepsilon$. This proves
\eqref{eq:draft-order-lipschitz}, including in the presence of ties.

Since $\rho$ sorts $\boldsymbol{y}$, its selected value satisfies
$y_{\rho(r)}=y_{(r)}$, regardless of which tied index occupies rank $r$.
Consequently,
\[
  x_{\rho(r)}-x_{(r)}=y_{(r)}-x_{(r)}-d_{\rho(r)},
\]
and the triangle inequality gives
\[
  |x_{\rho(r)}-x_{(r)}|
    \leq|y_{(r)}-x_{(r)}|+|d_{\rho(r)}|\leq2\varepsilon.
\]
This argument compares sorted values directly. Its bound does not depend on how
many coordinates change rank.
\end{proof}

\begin{proof}[Proof of Theorem~\ref{thm:moon_validity}]
We first replace the average over $B$ resamples by its conditional mean.
We then show that this mean estimates the same constant as in
Proposition~\ref{prop:bias_expansion}, and finally bound the error from using
finitely many resamples.
As the first step, let us consider sorting $W_{m,j}^*$. Define $q_N(\ell,r):=\E_*W_{m,\ell,(r)}^*$ and we first study its properties. 
Also for convenience we let $M_m^*:=\|\boldsymbol{W}_m^*\|_\infty$ denote the largest absolute bootstrap scaled
score error and similarly $M_N:=\|\boldsymbol{W}_N\|_\infty$.

\vspace{2mm}
\noindent
{\textit{1. Convergence results for $q_N(\ell,r)$.}} Lemma~\ref{lem:clt_bootstrap_clt} and
continuity of the order-statistic map give
\[
W_{m,\ell,(r)}^* \Rightarrow Z_{\ell,(r)} \qquad\text{in probability}
\]
by the continuous mapping theorem for weak convergence in probability. 
For each such sequence,
$\E_*[(W_{m,\ell,(r)}^*)^2]\leq J/4.$
Therefore, applying the uniform integrability for weak convergence in probability, it holds that 
\begin{equation}
    \label{eqn:P-q}q_N(\ell,r)\longrightarrow_{P}\delta_{\ell r}.
\end{equation}
Further, we note that 
$$
|q_N(\ell,r)| \leq 
\left(\E_*\left[(M_m^*)^2\right]\right)^{1/2}\leq \frac{\sqrt{J}}{2}.
$$
Dominated convergence therefore gives
\begin{equation}
\label{eqn:L-1-q}
      \E|q_N(\ell,r)-\delta_{\ell r}|\longrightarrow0.
\end{equation}

\vspace{2mm}
\noindent\emph{2. Cross-tier mistakes in the bootstrap.}
Let $C_{N,m}^*$ be the event that every bootstrap score in a higher population tier
exceeds every bootstrap score in a lower tier. For $L=1$, define this event to be
the whole bootstrap sample space. For $L>1$, fix an original sample in the event
$A_N$ from the proof of Proposition~\ref{prop:bias_expansion}. Its empirical
cross-tier score gaps exceed $\Delta/2$. If $M_m^*<\sqrt m\,\Delta/8$, every
bootstrap mean differs from its empirical mean by less than $\Delta/8$.
Each bootstrap cross-tier gap then exceeds $\Delta/2-2(\Delta/8)=\Delta/4>0$.
Markov's inequality gives, for these original samples,
\begin{equation}\label{eq:draft-bootstrap-tier-probability}
  {\Pr}_*\{(C_{N,m}^*)^c\}
    \leq{\Pr}_*\{M_m^*\geq\sqrt m\,\Delta/8\}
    \leq\frac{64\E_*[(M_m^*)^2]}{m\Delta^2}
    \leq\frac{16J}{m\Delta^2}.
\end{equation}
Thus the conditional probability of a cross-tier mistake tends to zero on $A_N$ uniformly.

\vspace{2mm}
\noindent\emph{3. Convergence of the bootstraped mean.}
For a single bootstrap resample, define
\[
  \mu_N(k):=\E_*W^*_{m,\hat\pi^*(k)},
\]
where $\hat{\pi}^*(k)$ is index of the $k$-th largest coordinate of $\hat{\boldsymbol\tau}^*$.
Define the empirical mean drift, on the bootstrap scale, by
\[
  d_{N,j}:=\sqrt m(\hat\tau_j-\tau_j) = \sqrt{\frac{m}{N}}W_{N,j},\qquad
  a_N:=\max_{1\leq j\leq J}|d_{N,j}|=\sqrt{m/N}\,M_N.
\]
The moment bound in Lemma~\ref{lem:clt_bootstrap_clt} gives
\begin{equation}\label{eq:draft-drift-moment}
  \E a_N^2\leq\frac{Jm}{4N}\longrightarrow0.
\end{equation}
In particular, $a_N$ tends to zero in $L^2$ and in probability under original
sampling. For $j\in S_\ell$, the bootstrap mean decomposes exactly as
\[
  \hat\tau_j^*=\tau^{(\ell)}+\frac{W_{m,j}^*+d_{N,j}}{\sqrt m}.
\]
Consequently, within $S_\ell$, sorting the bootstrap means is equivalent to sorting
$W_{m,j}^*+d_{N,j}$. Put
\[
q_N(\ell,r):=\E_*W_{m,\ell,(r)}^*.
\]
On $C_{N,m}^*$, the globally selected index $\hat\pi^*(k)$ belongs to $S_\ell$
and $k = K_{\ell} + r$. Apply
\eqref{eq:draft-selected-coordinate} with $\mathbf{x}=(W_{m,j}^*:j\in S_\ell)$
and perturbation $\mathbf{d}=(d_{N,j}:j\in S_\ell)$. It gives
\begin{equation}\label{eq:draft-selected-order-comparison}
  |W^*_{m,\hat\pi^*(k)}-W_{m,\ell,(r)}^*|
    \leq2\max_{j\in S_\ell}|d_{N,j}|\leq2a_N
    \quad\text{on }C_{N,m}^*.
\end{equation}

Off $C_{N,m}^*$, both $|W^*_{m,\hat\pi^*(k)}|$ and $|W_{m,\ell,(r)}^*|$ are at most $M_m^*$.
When $L>1$, for an original sample in $A_N$, conditional Cauchy--Schwarz and
\eqref{eq:draft-bootstrap-tier-probability} yield
\begin{align}
  |\mu_N(k)-q_N(\ell,r)|
    &\leq\E_*|W^*_{m,\hat\pi^*(k)}-W_{m,\ell,(r)}^*|\notag\\
    &\leq2a_N+2\sqrt{\E_*[(M_m^*)^2]\,
                         {\Pr}_*\{(C_{N,m}^*)^c\}}\notag\\
    &\leq2a_N+\frac{4J}{\sqrt m\,\Delta}.
    \label{eq:draft-conditional-comparison}
\end{align}
For an original sample outside $A_N$, the uniform bound
\[
  |\mu_N(k)-q_N(\ell,r)|
    \leq\E_*|W^*_{m,\hat\pi^*(k)}|+\E_*|W_{m,\ell,(r)}^*|\leq\sqrt J
\]
still holds. Taking expectations over the original data, we obtain for $L>1$
\begin{equation}\label{eq:draft-unconditional-comparison}
  \E|\mu_N(k)-q_N(\ell,r)|
    \leq2\E a_N+\frac{4J}{\sqrt m\,\Delta}
       +\sqrt J\,\Pr(A_N^c)\longrightarrow0.
\end{equation}
Here $\E a_N\leq\sqrt{Jm/(4N)}\to0$, by \eqref{eq:draft-drift-moment}.
When $L=1$, the cross-tier events always hold, and the same conclusion follows
from the bound $2\E a_N$ alone.
Combining \eqref{eq:draft-unconditional-comparison} with \eqref{eqn:L-1-q},
we see that 
\begin{equation}
    \label{eq:draft-conditional-mean-l1}
\mu_N(k)\longrightarrow \delta_{\ell r}\quad\text{in }L^1.
\end{equation}
Together with Proposition~\ref{prop:bias_expansion} and the identity $\E_*[\sqrt N\,\hat b_N(k)]=\mu_N(k)$ in \eqref{eq:draft-monte-carlo} below, this implies the second identity in \eqref{eq:all_position_correction}. 

\vspace{2mm}
\noindent\emph{4. Finitely many bootstrap resamples.}
As the final step, we prove the first identity in \eqref{eq:all_position_correction}.
Equation~\eqref{eq:bias_estimator} can be rewritten exactly as
 \[
        \sqrt N\,\hat b_N(k)
        =\frac1B\sum_{b=1}^B\sqrt m
        \left(
        \hat\tau^{(b)}_{\hat\pi^{(b)}(k)}
        -\hat\tau_{\hat\pi^{(b)}(k)}
        \right).
        \]
     These summands are conditionally i.i.d. with the conditional law of $W^*_{m, \hat{\pi}^*(k)}$. Their conditional mean is
$\mu_N(k)$, and each conditional variance is at most $J/4$. Consequently,
\begin{align}
  \E_*[\sqrt N\,\hat b_N(k)]&=\mu_N(k),\qquad
  \E_*\!\left[\bigl(\sqrt N\,\hat b_N(k)-\mu_N(k)\bigr)^2\right]
    \leq\frac{J}{4B}.
    \label{eq:draft-monte-carlo}
\end{align}
In particular,
\[
  \E\bigl|\sqrt N\,\hat b_N(k)-\mu_N(k)\bigr|
    \leq\sqrt{\frac{J}{4B}}\longrightarrow0.
\]
Together with \eqref{eq:draft-conditional-mean-l1}, this proves the stronger conclusion
\[
  \sqrt N\,\hat b_N(k)\longrightarrow\delta_{\ell r}\quad\text{in }L^1,
\]
and hence also in probability. 
We conclude the proof. 
\end{proof}

\subsection{Rank probabilities and tier recovery}\label{app:proof_rpm}
\begin{proof}[Proof of Proposition~\ref{prop:rank_probs}]
We first remark on a few facts.
Recall that as argued in previous proofs, 
\begin{equation}\label{eq:rpm-original-separation}
  \Pr(A_N^c)\leq\frac{4J}{N\Delta^2}\longrightarrow0, \qquad
  {\Pr}_*\{(C_{N,m}^*)^c\}
    \leq\frac{16J}{m\Delta^2}\qquad\text{on }A_N
  .
\end{equation}
Together, they imply that 
${\Pr}_*\{(C_{N,m}^*)^c\} \rightarrow_P 0$.
For a fixed tier $\ell$, define its tie set by
\[
  \mathcal H_\ell:=\{\boldsymbol{x}\in\R^J:
    x_j=x_h\text{ for some distinct }j,h\in S_\ell\}.
\]
For a singleton tier this set is empty. For a nonsingleton tier, each
$Z_j-Z_h$ is a centered normal variable with strictly positive variance, so
\begin{equation}\label{eq:rpm-no-gaussian-ties}
  \Pr\{\boldsymbol{Z}\in\mathcal H_\ell\}
    \leq\sum_{\substack{j<h\\j,h\in S_\ell}}\Pr\{Z_j=Z_h\}=0.
\end{equation}
Suppose that $j \in S_{\ell}$.
Let $g_{jr}^{(\ell)}(\boldsymbol{x})$ be the indicator that model $j$ occupies
position $r$ when the coordinates in $S_\ell$ are sorted in decreasing order of $\boldsymbol{x}$
with any fixed tie rule. Let $\tilde g_{jr}^{(\ell)}(\boldsymbol{x})$ be the
tie-averaged weight used by the RPM: if $x_j$ ties with $q-1$ other coordinates in
$S_\ell$, the tied block occupies $q$ consecutive positions and
$\tilde g_{jr}^{(\ell)}(\boldsymbol{x})=1/q$ at each of them.
Both functions take values in $[0,1]$. They coincide and are locally constant
outside $\mathcal H_\ell$. Thus their discontinuity sets have Gaussian
probability zero, and $\E g_{jr}^{(\ell)}(\boldsymbol{Z})=\E \tilde g_{jr}^{(\ell)}(\boldsymbol{Z})
    =p_{jr}^{(\ell)}.$
Write $\mathcal R_j^{(\ell)}$ for the within-tier rank of $Z_j$ among
$\{Z_h:h\in S_\ell\}$, so that $p_{jr}^{(\ell)}=\Pr\{\mathcal R_j^{(\ell)}=r\}$, and
$\hat\pi_\ell(r)$ for the model at position $r$ when the original empirical means
in $S_\ell$ are sorted with the fixed tie rule. For the $b$-th resample, let
$w_{jk}^{(b)}\in[0,1]$ be the rank weight of model $j$ at global position $k$,
with the tie averaging of Equation~\eqref{eq:rank_prob}, so that
$\hat R_{jk}=B^{-1}\sum_{b=1}^B w_{jk}^{(b)}$. Its conditional mean
$\hat R_{jk}^{\infty}:=\E_* w_{jk}^{(1)}$ is the RPM with infinitely many resamples.

\vspace{2mm}
\noindent\emph{1. Original-sample rank probabilities.}
Within $S_\ell$, all population means equal $\tau^{(\ell)}$. Sorting the original
empirical means in that tier is therefore exactly the same as sorting the
coordinates of $\boldsymbol{W}_N$ in that tier. Hence
\[
  \Pr\{\hat\pi_\ell(r)=j\}
    =\E g_{jr}^{(\ell)}(\boldsymbol{W}_N).
\]
Lemma~\ref{lem:clt_bootstrap_clt} gives
$\boldsymbol{W}_N\Rightarrow\boldsymbol{Z}$. Boundedness and almost-sure
continuity of $g_{jr}^{(\ell)}$ under the Gaussian law then imply
\[
  \Pr\{\hat\pi_\ell(r)=j\}\longrightarrow p_{jr}^{(\ell)}.
\]
On $A_N$, the global and within-tier rankings agree after the position shift
$K_\ell$. Consequently,
\[
  \left|R_{j,K_\ell+r}-\Pr\{\hat\pi_\ell(r)=j\}\right|
    \leq\Pr(A_N^c)\longrightarrow0.
\]
This proves the first convergence in \eqref{eq:rpm_block}. Also, for $k\notin\mathcal B_\ell := \{K_{\ell}+1,\dots, K_{\ell}+|S_{\ell}|\}$,
\[
  R_{jk}\leq\Pr(A_N^c)\longrightarrow0.
\]

\noindent\emph{2. The conditional bootstrap rank probabilities.}
Define the empirical drift vector and the bootstrap errors centered at the
population means by
\[
  \boldsymbol{d}_N:=\sqrt m(\hat{\boldsymbol{\tau}}-\boldsymbol{\tau})
       =\sqrt{m/N}\,\boldsymbol{W}_N,\qquad
  \boldsymbol{V}_{N,m}^*
       :=\sqrt m(\hat{\boldsymbol{\tau}}^{\,*}-\boldsymbol{\tau})
        =\boldsymbol{W}_m^*+\boldsymbol{d}_N.
\]
By the original-sample moment bound,
\[
  \E\|\boldsymbol{d}_N\|_\infty^2\leq\frac{Jm}{4N}\longrightarrow0.
\]
Thus $\boldsymbol{d}_N\to\boldsymbol{0}$ in probability. Together with the
conditional bootstrap limit in Lemma~\ref{lem:clt_bootstrap_clt}, this gives
\begin{equation}
\label{eq:rpm-shifted-bootstrap-clt}
  \boldsymbol{V}_{N,m}^*
    \Rightarrow  \boldsymbol{Z} \qquad\text{in probability}.
\end{equation}
This follows from a subsequence argument. 
From any subsequence, select a further subsequence along which
$\boldsymbol{d}_N\to\boldsymbol{0}$ almost surely. On almost every such original
data sequence, the conditional CLT holds by Lemma~\ref{lem:clt_bootstrap_clt},
and ordinary Slutsky convergence applies under that fixed conditional law.
This establishes \eqref{eq:rpm-shifted-bootstrap-clt} in probability by the
subsequence principle.

Within $S_\ell$, sorting bootstrap means is equivalent to sorting
$\boldsymbol{V}_{N,m}^*$. The same Gaussian-null discontinuity set in
\eqref{eq:rpm-no-gaussian-ties} therefore yields
\[
  \E_*\tilde g_{jr}^{(\ell)}(\boldsymbol{V}_{N,m}^*)
    \xrightarrow{P}p_{jr}^{(\ell)}.
\]
On $C_{N,m}^*$, the global bootstrap rank weight at $K_\ell+r$ equals this
within-tier weight. Since all weights lie in $[0,1]$,
\[
  \left|\hat R_{j,K_\ell+r}^{\infty}
     -\E_* \tilde g_{jr}^{(\ell)}(\boldsymbol{V}_{N,m}^*)\right|
    \leq{\Pr}_*\{(C_{N,m}^*)^c\}\xrightarrow{P}0.
\]
It follows that $\hat R_{j,K_\ell+r}^{\infty}\xrightarrow{P}p_{jr}^{(\ell)}$.
For $j\in S_\ell$ and $k\notin\mathcal B_\ell$,
\[
  0\leq\hat R_{jk}^{\infty}
     \leq{\Pr}_*\{(C_{N,m}^*)^c\}\xrightarrow{P}0.
\]

\vspace{2mm}
\noindent\emph{3. A finite number of bootstrap resamples.}
Write $\operatorname{Var}_*$ for variance conditional on $\mathcal F_N$.
The weights $w_{jk}^{(1)},\ldots,w_{jk}^{(B)}$ are conditionally independent
and identically distributed in $[0,1]$. Thus
\begin{equation}\label{eq:rpm-monte-carlo}
  \E_*\hat R_{jk}=\hat R_{jk}^{\infty},\qquad
  \E_*\!\left[(\hat R_{jk}-\hat R_{jk}^{\infty})^2\right]
    =\frac{\operatorname{Var}_*(w_{jk}^{(1)})}{B}\leq\frac1{4B}.
\end{equation}
Conditional Chebyshev's inequality and $B\to\infty$ show that
$\hat R_{jk}-\hat R_{jk}^{\infty}\xrightarrow{P}0$. Combining this with steps 1
and 2 proves all the limits in \eqref{eq:rpm_block}--\eqref{eq:rpm-outside-block}.
For each $(j,k)$, the deterministic quantity $R_{jk}$ and the random quantity
$\hat R_{jk}$ therefore converge to the same constant. Since $J$ is fixed,
taking a maximum over the $J^2$ entries proves \eqref{eq:rpm-consistency}.

\vspace{2mm}
\noindent\emph{4. Rank intervals stay inside their tier and contain its center.}
Fix $\alpha\in(0,1)$ and put $u:=\alpha/2$ and $v:=1-\alpha/2$.
For $j\in S_\ell$, define the limiting cumulative rank probabilities by
\[
  F_j^{\infty}(s):=
  \begin{cases}
    0, & s\leq K_\ell,\\
    \displaystyle\sum_{r=1}^{s-K_\ell}p_{jr}^{(\ell)},
       & K_\ell<s<K_\ell+q_\ell,\\
    1, & s\geq K_\ell+q_\ell.
  \end{cases}
\]
The RPM limits and finiteness of $J$ imply
\begin{equation}\label{eq:rpm-cdf-convergence}
  \max_{1\leq j\leq J}\max_{0\leq s\leq J}
    |F_j(s)-F_j^{\infty}(s)|\xrightarrow{P}0.
\end{equation}
Since $F_j^{\infty}(K_\ell)=0$ and
$F_j^{\infty}(K_\ell+q_\ell)=1$, with probability tending to one,
\[
  F_j(K_\ell)<u,\qquad F_j(K_\ell+q_\ell)>v.
\]
By the definition of $Q_j$, these inequalities give
\begin{equation}\label{eq:rpm-interval-block}
  I_j(\alpha)\subseteq[K_\ell+1,K_\ell+q_\ell].
\end{equation}

The Gaussian vector is centrally symmetric:
$\boldsymbol{Z}\overset d=-\boldsymbol{Z}$. Since within-tier Gaussian ties have
probability zero, negating all coordinates reverses their ranks. Therefore
\begin{equation}\label{eq:rpm-rank-symmetry}
  \mathcal R_j^{(\ell)}\overset d=q_\ell+1-\mathcal R_j^{(\ell)}.
\end{equation}
Define the middle global rank or ranks of the tier by
\[
  c_\ell^-:=K_\ell+\left\lfloor\frac{q_\ell+1}{2}\right\rfloor,\qquad
  c_\ell^+:=K_\ell+\left\lceil\frac{q_\ell+1}{2}\right\rceil.
\]
When $q_\ell=2h-1$ is odd, symmetry gives
\[
  \Pr\{\mathcal R_j^{(\ell)}\leq h-1\}
    =\frac{1-\Pr\{\mathcal R_j^{(\ell)}=h\}}2\leq\frac12,
\]
and $\Pr\{\mathcal R_j^{(\ell)}\leq h\}\geq1/2$.
When $q_\ell=2h$ is even, symmetry gives
$\Pr\{\mathcal R_j^{(\ell)}\leq h\}=1/2$.
Consequently, in both cases,
\[
  F_j^{\infty}(c_\ell^-)\geq\frac12,\qquad
  F_j^{\infty}(c_\ell^+-1)\leq\frac12.
\]
This includes a singleton tier, for which $c_\ell^-=c_\ell^+=K_\ell+1$.
Because $u<1/2<v$, convergence in \eqref{eq:rpm-cdf-convergence} implies,
with probability tending to one,
\[
  F_j(c_\ell^-)>u,\qquad F_j(c_\ell^+-1)<v.
\]
The quantile definition then gives
\begin{equation}\label{eq:rpm-common-center}
  Q_j(u)\leq c_\ell^-\leq c_\ell^+\leq Q_j(v),
  \qquad [c_\ell^-,c_\ell^+]\subseteq I_j(\alpha).
\end{equation}
There are finitely many models and tiers, so the block inclusions in
\eqref{eq:rpm-interval-block} and the center inclusions in
\eqref{eq:rpm-common-center} hold for all models simultaneously with
probability tending to one. The argument allows zero probability at a middle
rank. It uses the strict gap between the quantile thresholds and $1/2$, rather
than positivity of every Gaussian ordering probability.

\noindent\emph{5. The common-intersection rule recovers the partition.}
Consider the event on which $A_N$ and all the interval inclusions from step 4
hold. Its probability tends to one. On this event, the observed leaderboard
visits $S_1,\ldots,S_L$ as consecutive blocks, and no observed score tie contains
models from two different population tiers.

Every interval belonging to $S_\ell$ contains
$[c_\ell^-,c_\ell^+]$. Hence every observed tie block within $S_\ell$ is compatible,
and the running intersection remains nonempty while all models in that tier
are added. The procedure cannot split a population tier. Each such interval
also lies inside $[K_\ell+1,K_\ell+q_\ell]$, while intervals from the next tier
lie inside its disjoint rank block. Their intersection is empty, so the
procedure closes the current tier at exactly the population boundary.
It cannot merge two population tiers. Thus the reported ordered partition is
exactly $(S_1,\ldots,S_L)$ on this event, proving the claim.
\end{proof}

\section{Data, Implementation, and Reproducibility}\label{app:implementation}

This appendix explains how to construct a RULER report from task-level benchmark scores. It gives the reporting conventions, score definitions, snapshot sources, and computational details needed to reproduce the audit.

\subsection{Resampling and reporting conventions}\label{app:reporting}

The full audit draws $B=5{,}000$ resamples of $m=\lfloor N^{0.9}\rfloor$ complete task rows with replacement. Every row contains the outcomes of all models on the same task, preserving their within-task dependence. The validation experiments use $B=2{,}000$. The exponent is a predeclared, outcome-independent heuristic. Appendix~\ref{app:gamma_guidance} gives guidance on choosing it, and Appendix~\ref{app:mn_exponent_sensitivity} examines the sensitivity of the audit to it.

\paragraph{Score adjustments and observed ties.}
For the score correction in Equation~\ref{eq:bias_estimator}, each replicate ranks models by decreasing resample score. To give models with equal observed scores a common displayed adjustment, we average the adjustments over the positions occupied by the tie. Verified's displayed gap between positions 1 and 5 consequently falls from 2.40 to 2.13 pp after tie-averaged adjustments of approximately 0.90 and 0.63 pp.\footnote{Applying the individual position adjustments instead gives a gap of 1.94 pp, a reduction of about 19\%. These are alternative summaries of the same position-wise estimates. The main report consistently uses the tie-averaged display.}

\paragraph{Rank probabilities and intervals.}
For the RPM, we average uniformly over tied ranks. If $q$ models tie at positions $a,\ldots,a+q-1$ in a resample, each receives mass $1/q$ at each of those positions. The indicator in Equation~\ref{eq:rank_prob} is averaged over these tie-breaking permutations before averaging over resamples. Each row and column of the full RPM sums to one. Models with equal observed scores can still have different RPM rows because they may solve different tasks and have different dependence with their competitors.

Rank intervals use the smallest integer rank whose cumulative mass reaches each quantile threshold. The implementation permits $10^{-12}$ roundoff when comparing cumulative masses and identifying score ties. We report 80\% central rank mass by default and 90\%, 95\%, and 99\% in the sensitivity analysis. These intervals summarize the $m$-resample distribution. Proposition~\ref{prop:rank_probs} establishes its connection to population rank probabilities under exact tiers.

\paragraph{Tiers and leaderboard order.}
To form tiers, traverse models in raw-score order while keeping each observed score tie intact. A new block joins the current tier only if all its intervals and the current tier's intersection share a rank. Requiring a common rank prevents a chain of pairwise overlaps from joining models whose intervals have no overall intersection. If the intervals inside an observed score tie have no common rank, the implementation stops with an error, because that block is incompatible with the tier summary. This did not occur in the audit. The report retains the raw ordering and displays the raw score, tie-averaged adjustment, rank interval, tier, and RPM. Adjusted scores are reported in that original order, even when their values would imply a different ordering.

\paragraph{Complete reports.}
The interactive RULER report supplies model identities, complete RPMs, all 3 interval levels, and tiers for every benchmark.

\subsection{Benchmark scores and snapshot sources}\label{app:design}

For each benchmark, we form a matrix with tasks as rows and models as columns, using the task-level score defined in Table~\ref{tab:design}. A model's benchmark score is the mean of its column. RULER resamples whole rows, so all models are evaluated on the same sampled tasks and the procedure preserves their dependence within each task. The resulting score adjustments and rank summaries describe uncertainty from sampling tasks. For judged outcomes, they condition on the released responses and judge annotations.

\begin{table}[!t]
\centering
\caption{Each RULER report refers to a specified score and task population.}\label{tab:design}
\begin{tabular}{@{}>{\raggedright\arraybackslash}p{2.0cm}>{\raggedright\arraybackslash}p{3.6cm}>{\raggedright\arraybackslash}p{7.5cm}@{}}
\toprule Family & Score and sampling unit & Target population and interpretation \\\midrule
SWE-bench (5 benchmarks) & Binary issue resolution. Unit: issue. & Software issues under each variant's task and scaffold rules, evaluated by the retained model-plus-agent systems. The analysis conditions on submitted trajectories. Model-plus-agent systems can differ across entries. \\
Tau2-Bench (3 domains) & Released task reward in $[0,1]$. Unit: customer-service scenario. & Airline, retail, or telecom scenarios under the released agents and environment. Frozen outcomes retain the realized agent randomness and environment. \\
AlpacaEval 2 & Raw weighted rate: soft preference minus 1. Unit: instruction. & Instructions compared against a common GPT-4-Turbo reference, conditional on released responses and judge annotations. Instructions are equally weighted. The raw score differs from the length-controlled leaderboard metric. \\
WildBench v2 & Checklist-guided judge score divided by 10, in $[0.1,1]$. Unit: prompt/session. & Real-user-style prompts under the frozen scoring protocol. Uncertainty concerns sampling prompts, conditional on the released judge scores. \\
LiveBench (4 benchmarks) & Released task score in $[0,1]$. Unit: task/question. & Tasks in the retained model-coverage subset. We report Coding, Language, Instruction Following, and their task-weighted pooled mean. This subset and pooled score differ from the full live leaderboard. \\
\bottomrule
\end{tabular}

\floatnote{The listed populations are modeling targets. Released tasks need not be random samples from them. Pooled analyses assume independent task rows from the target and resample each row's complete model vector. Judged scores condition on the released responses and annotations.}
\end{table}

\paragraph{SWE-bench and Tau2-Bench.}
For SWE-bench, a task is a software issue and its score is 1 if the evaluated model-plus-agent system resolves it and 0 otherwise. The column mean is therefore the issue-resolution rate. For SWE-bench Pro, we use the 395 tasks for which every system has a released record. For Tau2-Bench, a task is a customer-service scenario and its released reward lies in $[0,1]$. RULER applies to these rewards in the same way, separately for Airline, Retail, and Telecom.

\paragraph{AlpacaEval 2.}
We analyze the raw weighted win rate against the common GPT-4-Turbo reference, rather than the length-controlled leaderboard score \citep{dubois2024length}. Each instruction has a soft judge preference in $[1,2]$, from which we subtract 1 to obtain a score in $[0,1]$. The ``weighted'' rate averages these soft preferences, with all valid instructions receiving equal weight. The official AlpacaEval scorer computes the same average. Each resampled row contains all models' scores for the same instruction, preserving the shared reference and prompt.

\paragraph{WildBench.}
The task-level outcome is the released checklist-guided judge score for a prompt or session \citep{lin2024wildbench}. Dividing its 1--10 score by 10 puts it in $[0.1,1]$. We average these normalized scores across the 988 prompts and use each prompt's complete vector of model scores as a resampling row.

\paragraph{LiveBench.}
We use released task-level accuracy scores in $[0,1]$ \citep{white2024livebench}. The analyzed matrix contains 81 models on 318 tasks: 128 Coding, 140 Language, and 50 Instruction Following tasks. RULER reports each category separately and a pooled score that weights all 318 tasks equally. The pooled score therefore weights categories by their task counts and describes this analyzed subset, rather than the full LiveBench leaderboard metric.

\paragraph{Snapshot sources.}
Table~\ref{tab:benchmark_sources} records the date and source of every benchmark snapshot. Table~\ref{tab:cross_variant_summary} gives the corresponding numbers of tasks and models.

\begin{table}[!t]
\centering\small
\caption{Each benchmark audit is tied to an explicitly dated data source.}\label{tab:benchmark_sources}
\begin{tabular}{@{}>{\raggedright\arraybackslash}p{2.5cm}>{\raggedright\arraybackslash}p{2.1cm}>{\raggedright\arraybackslash}p{9.8cm}@{}}
\toprule Benchmark & Snapshot date & Frozen source \\\midrule
SWE Verified & 2026-02-27 & \href{https://github.com/SWE-bench/swe-bench.github.io/blob/7c4289f30aa1a1c63c2e2a25aae30c16d92b5114/data/leaderboards.json}{Official leaderboard, \texttt{7c4289f}}, and \href{https://github.com/SWE-bench/experiments/tree/82be2b5ac4967af9559484e7bffef227e97348af/evaluation/verified}{task outcomes, \texttt{82be2b5}} (2026-01-26). \\
SWE Bash-only & 2026-02-27 & \href{https://github.com/SWE-bench/swe-bench.github.io/blob/7c4289f30aa1a1c63c2e2a25aae30c16d92b5114/data/leaderboards.json}{Official leaderboard, \texttt{7c4289f}}, and \href{https://github.com/SWE-bench/experiments/tree/2f95aad/evaluation/bash-only}{task outcomes, \texttt{2f95aad}} (2026-02-27). \\
SWE Multimodal & 2026-02-27 & \href{https://github.com/SWE-bench/swe-bench.github.io/blob/7c4289f30aa1a1c63c2e2a25aae30c16d92b5114/data/leaderboards.json}{Official leaderboard, \texttt{7c4289f}}, and \href{https://github.com/SWE-bench/experiments/tree/cff22d0/evaluation/multimodal}{task outcomes, \texttt{cff22d0}} (2025-11-21). \\
SWE Multilingual & 2026-02-27 & \href{https://github.com/SWE-bench/swe-bench.github.io/blob/7c4289f30aa1a1c63c2e2a25aae30c16d92b5114/data/leaderboards.json}{Official leaderboard, \texttt{7c4289f}}, and \href{https://github.com/SWE-bench/experiments/tree/6083d31/evaluation/multilingual}{task outcomes, \texttt{6083d31}} (2026-02-24). \\
SWE Pro & 2026-03-05 & \href{https://docent.transluce.org}{Docent} per-task resolution records, collection \path{032fb63d-4992-4bfc-911d-3b7dafcb931f}. \\
Tau2-Bench & 2026-04-10 & Released leaderboard outcomes from \href{https://github.com/sierra-research/tau2-bench}{\texttt{sierra-research/tau2-bench}} for Airline, Retail, and Telecom. \\
AlpacaEval 2 & 2025-08-09 & \href{https://github.com/tatsu-lab/alpaca_eval/tree/cd543a149df89434d8a54582c0151c0b945c3d20}{\texttt{tatsu-lab/alpaca\_eval}, \texttt{cd543a1}}, using \texttt{weighted\_alpaca\_eval\_gpt4\_turbo} annotations. \\
WildBench v2 & 2024-08-01 & \href{https://github.com/allenai/WildBench/tree/d6b8dcaf377d173d031980f97c16e1a82618c03d}{\texttt{allenai/WildBench}, \texttt{d6b8dca}}, release \texttt{v2.0625}, with judge \texttt{gpt-4o-2024-05-13}. \\
LiveBench & 2025-04-07 & Hugging Face \href{https://huggingface.co/datasets/livebench/model_judgment/tree/9704e5da7bfbefe75ac1482a13de827127295993}{\texttt{livebench/model\_judgment}, \texttt{9704e5d}}, using the released leaderboard split. \\
\bottomrule
\end{tabular}
\floatnote{Dates for the 4 classic SWE-bench variants refer to the leaderboard revision, with outcome-revision dates shown separately. Pro and Tau2 dates identify the local snapshot retrievals. AlpacaEval, WildBench, and LiveBench dates identify the pinned upstream revisions. Revision timestamps are interpreted in UTC.}
\end{table}

\paragraph{Computation.}
RULER reuses released task outcomes, so producing the report requires no language-model training, inference, or judge calls. Computation uses CPU-based Python with NumPy, SciPy, pandas, statsmodels, and Matplotlib. The full audit uses 5,000 resamples per benchmark with seed 42. The sensitivity sweep comprises 154 such runs, the split experiment uses 200 partitions per benchmark with 2,000 resamples each, and the calibrated comparison uses 100 synthetic benchmarks with 2,000 resamples per bootstrap method. Complete runtime and peak-memory measurements are not reported.
\section{Supplementary Results for Validation}\label{app:validation}

The held-out test and calibrated simulation in Section~\ref{sec:verification} assess the correction using different reference scores. We describe their protocols and report the complete results below.

\subsection{Sample-splitting protocol and results}\label{app:split_protocol}

Each of the 14 benchmarks uses 200 random partitions into a selection half of $N_S=\lfloor N/2\rfloor$ tasks and an evaluation half of $N_E=N-N_S$ tasks. SWE-bench splits balance repository representation. AlpacaEval uses 5 source-dataset strata and pooled LiveBench uses 3 categories. WildBench, individual LiveBench categories, and Tau2 use unstratified splits. When strata have odd sizes, the remaining assignments are randomized to reach the exact overall half-size. This approximately fixes the population mix but does not remove dependence among issues from the same repository. Every selection half uses $m=\lfloor N_S^{0.9}\rfloor$ and $B=2{,}000$ bootstrap resamples.

For benchmark $h$, the signed mean errors $\bar d_{hk}$ and $\bar e_{hk}$ are the averages of $d_k$ and $e_k$ in Equation~\ref{eq:split_reference} over the 200 random partitions. Table~\ref{tab:split_aggregate} reports $14^{-1}\sum_h|\bar d_{hk}|$ and $14^{-1}\sum_h|\bar e_{hk}|$. Averaging signed errors first isolates the mean selection error within each benchmark. Taking its magnitude before combining benchmarks prevents overstatement in one benchmark from canceling understatement in another.

On Verified, where each half contains 250 tasks, the mean winner error falls from 1.96 to 0.18 pp (Table~\ref{tab:sample_split_gaps}). Table~\ref{tab:all_splits} gives all 70 benchmark--position combinations. Each residual is computed within its split before summarization. The split SE is the standard deviation across partitions divided by $\sqrt{200}$ and measures Monte Carlo precision conditional on the frozen task pool.

\begin{table}[!t]
\centering\small
\caption{The correction substantially reduces SWE-bench Verified's mean winner error.}\label{tab:sample_split_gaps}
\begin{tabular}{@{}cccc@{}}
\toprule
Rank & \shortstack{Mean Error\\Before RULER} & \shortstack{Mean Score\\Correction} & \shortstack{Mean Error\\After RULER} \\
\midrule
1 & 1.96 & 1.79 & +0.18 \\
2 & 0.66 & 1.38 & -0.72 \\
3 & 0.72 & 1.24 & -0.52 \\
4 & 1.41 & 1.16 & +0.26 \\
5 & 1.52 & 1.08 & +0.44 \\
\bottomrule
\end{tabular}

\floatnote{Results average 200 random half-splits with $N_S=N_E=250$, $m=\lfloor N_S^{0.9}\rfloor$, and $B=2{,}000$. Error Before RULER is the selected model's selection-half score minus its evaluation-half score. Score Correction is the estimated bias. Error After RULER subtracts that correction within each split. Entries are means across splits in percentage points (pp).}
\end{table}

\begingroup\small\setlength{\tabcolsep}{3pt}\setlength{\LTpre}{0pt}
\begin{longtable}{@{}lcccc@{}}
\caption{The reduction in selection error varies across benchmarks and ranks.}\label{tab:all_splits}\\
\toprule
& & \multicolumn{3}{c}{Mean $\pm$ SE Across Splits} \\
\cmidrule(l){3-5}
Benchmark & Rank & \shortstack{Error Before\\RULER} & \shortstack{Score\\Correction} & \shortstack{Error After\\RULER} \\
\midrule
\endfirsthead
\toprule
& & \multicolumn{3}{c}{Mean $\pm$ SE Across Splits} \\
\cmidrule(l){3-5}
Benchmark & Rank & \shortstack{Error Before\\RULER} & \shortstack{Score\\Correction} & \shortstack{Error After\\RULER} \\
\midrule
\endhead
\bottomrule
\endfoot
SWE Bash-only & 1 & $+2.20\pm0.25$ & $+1.60\pm0.02$ & $+0.60\pm0.26$ \\
SWE Bash-only & 2 & $+1.24\pm0.26$ & $+1.07\pm0.01$ & $+0.17\pm0.26$ \\
SWE Bash-only & 3 & $+0.62\pm0.25$ & $+0.79\pm0.01$ & $-0.18\pm0.25$ \\
SWE Bash-only & 4 & $+0.20\pm0.27$ & $+0.63\pm0.01$ & $-0.43\pm0.27$ \\
SWE Bash-only & 5 & $+0.56\pm0.26$ & $+0.52\pm0.01$ & $+0.04\pm0.27$ \\
SWE Verified & 1 & $+1.96\pm0.24$ & $+1.79\pm0.02$ & $+0.18\pm0.25$ \\
SWE Verified & 2 & $+0.66\pm0.25$ & $+1.38\pm0.02$ & $-0.72\pm0.26$ \\
SWE Verified & 3 & $+0.72\pm0.27$ & $+1.24\pm0.01$ & $-0.52\pm0.27$ \\
SWE Verified & 4 & $+1.41\pm0.28$ & $+1.16\pm0.01$ & $+0.26\pm0.29$ \\
SWE Verified & 5 & $+1.52\pm0.25$ & $+1.08\pm0.01$ & $+0.44\pm0.25$ \\
SWE Multimodal & 1 & $+1.68\pm0.22$ & $+1.29\pm0.01$ & $+0.39\pm0.23$ \\
SWE Multimodal & 2 & $+0.55\pm0.23$ & $+0.57\pm0.01$ & $-0.01\pm0.23$ \\
SWE Multimodal & 3 & $-0.03\pm0.24$ & $+0.14\pm0.01$ & $-0.17\pm0.24$ \\
SWE Multimodal & 4 & $-0.13\pm0.25$ & $-0.11\pm0.01$ & $-0.02\pm0.25$ \\
SWE Multimodal & 5 & $-0.77\pm0.26$ & $-0.09\pm0.02$ & $-0.69\pm0.26$ \\
SWE Multilingual & 1 & $+2.62\pm0.37$ & $+2.39\pm0.03$ & $+0.23\pm0.39$ \\
SWE Multilingual & 2 & $+0.88\pm0.38$ & $+1.70\pm0.02$ & $-0.82\pm0.39$ \\
SWE Multilingual & 3 & $+0.92\pm0.35$ & $+1.26\pm0.02$ & $-0.34\pm0.35$ \\
SWE Multilingual & 4 & $+0.92\pm0.38$ & $+0.96\pm0.02$ & $-0.04\pm0.38$ \\
SWE Multilingual & 5 & $+1.65\pm0.41$ & $+0.68\pm0.02$ & $+0.96\pm0.41$ \\
SWE Pro & 1 & $+1.88\pm0.30$ & $+1.59\pm0.03$ & $+0.28\pm0.32$ \\
SWE Pro & 2 & $+0.45\pm0.32$ & $+0.63\pm0.03$ & $-0.18\pm0.33$ \\
SWE Pro & 3 & $+1.83\pm0.30$ & $+0.23\pm0.03$ & $+1.59\pm0.31$ \\
SWE Pro & 4 & $+0.31\pm0.31$ & $-0.27\pm0.03$ & $+0.58\pm0.32$ \\
SWE Pro & 5 & $-1.07\pm0.31$ & $-0.51\pm0.05$ & $-0.56\pm0.34$ \\
Tau2 Airline & 1 & $+4.35\pm0.85$ & $+5.94\pm0.11$ & $-1.59\pm0.91$ \\
Tau2 Airline & 2 & $+4.31\pm0.90$ & $+3.34\pm0.07$ & $+0.97\pm0.92$ \\
Tau2 Airline & 3 & $+2.07\pm0.87$ & $+1.76\pm0.06$ & $+0.31\pm0.88$ \\
Tau2 Airline & 4 & $-0.00\pm0.83$ & $+0.25\pm0.05$ & $-0.25\pm0.83$ \\
Tau2 Airline & 5 & $-2.02\pm0.77$ & $-1.21\pm0.05$ & $-0.81\pm0.77$ \\
Tau2 Retail & 1 & $+4.58\pm0.40$ & $+3.05\pm0.06$ & $+1.53\pm0.44$ \\
Tau2 Retail & 2 & $+1.32\pm0.37$ & $+1.34\pm0.03$ & $-0.02\pm0.38$ \\
Tau2 Retail & 3 & $+0.48\pm0.39$ & $+0.37\pm0.03$ & $+0.11\pm0.40$ \\
Tau2 Retail & 4 & $-1.73\pm0.42$ & $-0.44\pm0.03$ & $-1.30\pm0.43$ \\
Tau2 Retail & 5 & $-1.09\pm0.45$ & $-0.81\pm0.04$ & $-0.28\pm0.46$ \\
Tau2 Telecom & 1 & $+0.22\pm0.14$ & $+0.63\pm0.02$ & $-0.41\pm0.15$ \\
Tau2 Telecom & 2 & $+0.05\pm0.14$ & $+0.07\pm0.02$ & $-0.02\pm0.15$ \\
Tau2 Telecom & 3 & $+0.11\pm0.38$ & $-0.70\pm0.03$ & $+0.82\pm0.41$ \\
Tau2 Telecom & 4 & $+0.79\pm0.49$ & $+1.06\pm0.04$ & $-0.27\pm0.50$ \\
Tau2 Telecom & 5 & $+1.64\pm0.54$ & $+0.64\pm0.04$ & $+0.99\pm0.55$ \\
AlpacaEval 2 & 1 & $+0.96\pm0.17$ & $+0.85\pm0.01$ & $+0.11\pm0.18$ \\
AlpacaEval 2 & 2 & $-0.32\pm0.17$ & $+0.17\pm0.01$ & $-0.49\pm0.18$ \\
AlpacaEval 2 & 3 & $-0.06\pm0.18$ & $-0.01\pm0.02$ & $-0.05\pm0.19$ \\
AlpacaEval 2 & 4 & $+0.72\pm0.17$ & $+0.25\pm0.02$ & $+0.47\pm0.18$ \\
AlpacaEval 2 & 5 & $+0.10\pm0.19$ & $+0.41\pm0.02$ & $-0.31\pm0.19$ \\
WildBench v2 & 1 & $+0.14\pm0.05$ & $+0.17\pm0.00$ & $-0.03\pm0.05$ \\
WildBench v2 & 2 & $-0.08\pm0.05$ & $+0.03\pm0.01$ & $-0.10\pm0.06$ \\
WildBench v2 & 3 & $+0.04\pm0.05$ & $+0.11\pm0.01$ & $-0.08\pm0.06$ \\
WildBench v2 & 4 & $+0.33\pm0.05$ & $+0.19\pm0.01$ & $+0.14\pm0.06$ \\
WildBench v2 & 5 & $+0.10\pm0.06$ & $+0.16\pm0.00$ & $-0.06\pm0.06$ \\
LiveBench pooled & 1 & $+0.24\pm0.28$ & $+0.85\pm0.03$ & $-0.61\pm0.30$ \\
LiveBench pooled & 2 & $+3.98\pm0.28$ & $+1.96\pm0.03$ & $+2.02\pm0.29$ \\
LiveBench pooled & 3 & $+1.47\pm0.27$ & $+1.48\pm0.02$ & $-0.01\pm0.28$ \\
LiveBench pooled & 4 & $-0.25\pm0.28$ & $+1.10\pm0.02$ & $-1.35\pm0.28$ \\
LiveBench pooled & 5 & $-0.68\pm0.32$ & $+0.90\pm0.02$ & $-1.58\pm0.33$ \\
LiveBench Coding & 1 & $+1.54\pm0.42$ & $+2.33\pm0.06$ & $-0.79\pm0.45$ \\
LiveBench Coding & 2 & $+0.39\pm0.44$ & $+1.97\pm0.07$ & $-1.58\pm0.47$ \\
LiveBench Coding & 3 & $+4.75\pm0.52$ & $+2.55\pm0.06$ & $+2.20\pm0.54$ \\
LiveBench Coding & 4 & $+3.27\pm0.52$ & $+2.46\pm0.04$ & $+0.82\pm0.54$ \\
LiveBench Coding & 5 & $+2.34\pm0.53$ & $+2.22\pm0.04$ & $+0.12\pm0.54$ \\
LiveBench Instr. & 1 & $+7.83\pm0.49$ & $+4.61\pm0.07$ & $+3.22\pm0.53$ \\
LiveBench Instr. & 2 & $+5.49\pm0.53$ & $+3.84\pm0.05$ & $+1.65\pm0.55$ \\
LiveBench Instr. & 3 & $+4.08\pm0.52$ & $+3.51\pm0.03$ & $+0.57\pm0.54$ \\
LiveBench Instr. & 4 & $+3.92\pm0.55$ & $+3.21\pm0.03$ & $+0.71\pm0.57$ \\
LiveBench Instr. & 5 & $+2.94\pm0.57$ & $+2.90\pm0.03$ & $+0.04\pm0.58$ \\
LiveBench Language & 1 & $+3.16\pm0.45$ & $+2.95\pm0.05$ & $+0.21\pm0.48$ \\
LiveBench Language & 2 & $+0.26\pm0.41$ & $+1.91\pm0.05$ & $-1.65\pm0.44$ \\
LiveBench Language & 3 & $-1.59\pm0.47$ & $+1.43\pm0.06$ & $-3.02\pm0.49$ \\
LiveBench Language & 4 & $+1.44\pm0.50$ & $+1.75\pm0.06$ & $-0.30\pm0.52$ \\
LiveBench Language & 5 & $+0.77\pm0.49$ & $+2.06\pm0.05$ & $-1.29\pm0.50$ \\
\end{longtable}
\floatnote{All 70 benchmark--position combinations use 200 random half-splits. Selection halves contain $N_S=\lfloor N/2\rfloor$ tasks, with $m=\lfloor N_S^{0.9}\rfloor$ and $B=2{,}000$. Error Before RULER is the selected model's selection-half score minus its evaluation-half score. Score Correction is the estimated bias. Error After RULER subtracts that correction within each split. Entries are means and standard errors (SEs) across splits. All scores and errors are in pp. SEs measure Monte Carlo precision over partitions of the same frozen task pool, conditional on that pool.}

\endgroup

\subsection{Calibrated simulation and alternative adjustments}\label{app:alternatives}

Whereas the held-out test uses a second sample to evaluate selected models, the calibrated simulation makes their true marginal means available. We retain dependence across models through a shared task difficulty and generate outcomes as
\begin{equation}
 Y_{ij}\mid T_i\sim\operatorname{Bernoulli}\{\sigma(\alpha_j+T_i)\},\qquad
 T_i\sim\mathcal N(0,\sigma_T^2),\qquad \sigma(x)=(1+e^{-x})^{-1},\label{eq:logistic_dgp}
\end{equation}
with conditionally independent model outcomes given the shared task effect. The common effect produces cross-model dependence, while the marginal model means provide known targets for the selected-score errors.

A Bernoulli GLM with model and task effects is fitted to the 472 nonconstant rows of Verified. The 28 tasks with constant outcomes across models are omitted from fitting, while the empirical audit uses all 500 tasks. Task effects are centered and their mean absorbed into the model intercepts. Their standard deviation is $\hat\sigma_T=2.6481$. Marginal model means are
\begin{equation}
 \tau_j=\int\sigma(\hat\alpha_j+\hat\sigma_T z)\phi(z)\,dz,\label{eq:sim_truth}
\end{equation}
computed with 120-node Gauss--Hermite quadrature. Synthetic benchmark $r=0,\ldots,99$ uses seed $42+r$, and its $m$-out-of-$N$ and standard bootstraps use seeds $42+r+5{,}000{,}000$ and $42+r+7{,}000{,}000$, respectively. Sample splitting, described below, uses seed $42+r+9{,}000{,}000$. Each method uses the same selected model at each raw rank on a given synthetic benchmark. Ties retain the original sort rule.

For the EB baseline, let $x_j=\sum_iY_{ij}$. Fit positive $(a,b)$ by maximizing the beta-binomial marginal composite log-likelihood
\begin{equation}
 \ell(a,b)=\sum_{j=1}^J\left[\log B(x_j+a,N-x_j+b)-\log B(a,b)\right],\label{eq:eb_fit}
\end{equation}
where binomial coefficients are constant in $(a,b)$. Optimize $(\log a,\log b)$ within $[-10,14]^2$ using analytic gradients and two initializations: $(1,1)$ and a moment-based concentration $\max\{\bar p(1-\bar p)/s_p^2-1,0.1\}$ split according to $\bar p$, with numerical clipping. Select the successful fit with the lowest negative log-likelihood. A bounded derivative-free fallback is provided if needed. The posterior mean is $(x_j+a)/(N+a+b)$, so the adjustment for the selected model is $x_j/N$ minus that mean. This composite likelihood ignores cross-model dependence and imposes a beta population model. The comparison concerns this specified EB baseline.

Sample splitting can also be used to correct scores. The held-out test in Section~\ref{sec:verification} estimates the half-sample bias $b_{N_S}(k)$ with the split error $d_k$ in Equation~\ref{eq:split_reference}. Because Proposition~\ref{prop:bias_expansion} shows that the leading term of the bias is proportional to $N^{-1/2}$, averaging $d_k$ over many random partitions and multiplying the average by $\sqrt{N_S/N}$ gives a first-order estimate of $b_N(k)$. Subtracting this estimate from the score at position $k$ of the full leaderboard gives the corrected score. We use 200 random partitions of each synthetic benchmark into 2 halves of $N_S=236$ tasks, so the multiplier is $\sqrt{1/2}$. The correction has 2 limitations. First, a half-sample ranking is noisier than the full ranking and often places different models at a given position. The split error therefore reflects selection among $N_S$ tasks, and rescaling it to $N$ tasks corrects only the leading term of the bias. Second, each half contains only half of the tasks, so the estimate is more variable than the $m$-out-of-$N$ adjustment. At rank~1, the sample-splitting adjustment has a standard deviation of 0.49 pp across synthetic benchmarks, compared with 0.29 pp for the $m$-out-of-$N$ adjustment.

Table~\ref{tab:calib_sim_bias} reports the mean signed errors of all 4 corrections. The $m$-out-of-$N$ correction leaves a mean winner residual of 0.21 pp, compared with 0.46 pp for the standard bootstrap and 1.10 pp for EB. Thus the 3 adjustments remove approximately 85\%, 68\%, and 22\% of mean winner bias, respectively. Resampling $m<N$ tasks therefore removes more bias than resampling all $N$. Sample splitting leaves a mean residual of 0.06 pp, removing nearly all of the winner bias, but its adjustment is more variable and its winner RMSE is higher than that of the $m$-out-of-$N$ correction. At rank~1, the $m$-out-of-$N$ correction has the lowest RMSE, 1.55 pp, compared with 1.62 pp for the standard bootstrap, 1.64 pp for sample splitting, and 1.75 pp for EB (Table~\ref{tab:calib_sim_rmse}). It also has the lowest RMSE at positions 2--5 and ties EB at rank~2 to the displayed precision.

\begin{table}[!t]\centering\small
\caption{The $m$-out-of-$N$ correction and sample splitting leave the smallest mean winner residuals.}\label{tab:calib_sim_bias}
\begin{tabular}{@{}cccccc@{}}
\toprule
& \multicolumn{5}{c}{Mean Signed Error (pp)} \\
\cmidrule(l){2-6}
Rank & No Correction & \shortstack{$m$-out-of-$N$\\bootstrap} & \shortstack{Standard\\bootstrap} & EB & \shortstack{Sample\\splitting} \\
\midrule
1 & $+1.41\pm0.14$ & $+0.21\pm0.15$ & $+0.46\pm0.16$ & $+1.10\pm0.14$ & $+0.06\pm0.16$ \\
2 & $+0.80\pm0.17$ & $-0.06\pm0.17$ & $+0.20\pm0.18$ & $+0.49\pm0.17$ & $-0.02\pm0.18$ \\
3 & $+0.66\pm0.18$ & $-0.12\pm0.18$ & $+0.13\pm0.18$ & $+0.36\pm0.18$ & $-0.07\pm0.18$ \\
4 & $+0.98\pm0.16$ & $+0.23\pm0.16$ & $+0.41\pm0.17$ & $+0.69\pm0.16$ & $+0.16\pm0.17$ \\
5 & $+0.96\pm0.17$ & $+0.24\pm0.18$ & $+0.40\pm0.18$ & $+0.68\pm0.17$ & $+0.13\pm0.19$ \\
\bottomrule
\end{tabular}

\floatnote{Entries are the mean signed error of the selected model's score relative to its known mean, $\pm$ Monte Carlo SE, in pp over the 100 synthetic benchmarks of Table~\ref{tab:calib_sim}. No Correction gives the raw selection bias. The other columns give residuals after each correction. The $m$-out-of-$N$ and standard bootstraps use $m=\lfloor N^{0.9}\rfloor$ and $m=N$, respectively, with 2,000 resamples each. EB denotes empirical Bayes. It replaces each selected model's score with its posterior mean under a beta-binomial model fitted to the benchmark. Sample splitting subtracts the split error averaged over 200 random half-splits and rescaled to $N$ tasks.}
\end{table}

\begin{table}[!t]\centering\small
\caption{The $m$-out-of-$N$ correction gives the winner's score the lowest root mean squared error.}\label{tab:calib_sim_rmse}
\begin{tabular}{@{}cccccc@{}}
\toprule
& \multicolumn{5}{c}{RMSE (pp)} \\
\cmidrule(l){2-6}
Rank & No Correction & \shortstack{$m$-out-of-$N$\\bootstrap} & \shortstack{Standard\\bootstrap} & EB & \shortstack{Sample\\splitting} \\
\midrule
1 & $1.97\pm0.13$ & $1.55\pm0.10$ & $1.62\pm0.10$ & $1.75\pm0.12$ & $1.64\pm0.09$ \\
2 & $1.85\pm0.18$ & $1.74\pm0.14$ & $1.77\pm0.15$ & $1.74\pm0.17$ & $1.77\pm0.14$ \\
3 & $1.87\pm0.16$ & $1.78\pm0.14$ & $1.80\pm0.15$ & $1.79\pm0.15$ & $1.84\pm0.15$ \\
4 & $1.85\pm0.11$ & $1.64\pm0.13$ & $1.71\pm0.12$ & $1.72\pm0.11$ & $1.73\pm0.14$ \\
5 & $1.97\pm0.15$ & $1.79\pm0.13$ & $1.87\pm0.14$ & $1.85\pm0.14$ & $1.87\pm0.14$ \\
\bottomrule
\end{tabular}

\floatnote{Entries are the root mean squared error (RMSE) of the selected model's score relative to its known mean, $\pm$ Monte Carlo SE from the delta method, in pp over the synthetic benchmarks and corrections of Table~\ref{tab:calib_sim_bias}.}
\end{table}

\begin{table}[!t]\centering\small
\caption{At rank~1, every alternative has a higher RMSE than the $m$-out-of-$N$ correction on the same synthetic benchmarks.}\label{tab:calib_sim_rmse_diff}
\begin{tabular}{@{}ccccc@{}}
\toprule
& \multicolumn{4}{c}{RMSE minus $m$-out-of-$N$ RMSE (pp)} \\
\cmidrule(l){2-5}
Rank & No Correction & \shortstack{Standard\\bootstrap} & EB & \shortstack{Sample\\splitting} \\
\midrule
1 & $+0.42\pm0.09$ & $+0.07\pm0.02$ & $+0.20\pm0.07$ & $+0.09\pm0.03$ \\
2 & $+0.11\pm0.09$ & $+0.03\pm0.02$ & $+0.00\pm0.07$ & $+0.03\pm0.03$ \\
3 & $+0.08\pm0.07$ & $+0.01\pm0.03$ & $+0.00\pm0.05$ & $+0.05\pm0.02$ \\
4 & $+0.21\pm0.08$ & $+0.07\pm0.02$ & $+0.07\pm0.06$ & $+0.09\pm0.03$ \\
5 & $+0.18\pm0.06$ & $+0.07\pm0.02$ & $+0.06\pm0.04$ & $+0.08\pm0.03$ \\
\bottomrule
\end{tabular}

\floatnote{Each entry is a method's RMSE minus that of the $m$-out-of-$N$ bootstrap over the same 100 synthetic benchmarks, $\pm$ delta-method Monte Carlo SE of the paired difference, in pp. Positive entries indicate a higher RMSE than the $m$-out-of-$N$ bootstrap. EB denotes empirical Bayes.}
\end{table}

\begin{table}[!t]\centering\small
\caption{The $m$-out-of-$N$ correction has the lowest RMSE across the first 5 ranks.}\label{tab:alternatives_full}
\begin{tabular}{@{}clccc@{}}
\toprule
& & & \multicolumn{2}{c}{After Correction} \\
\cmidrule(l){4-5}
Rank & Correction Method & \shortstack{Mean Score\\Correction (pp)} & \shortstack{Mean Residual\\(pp)} & RMSE (pp) \\
\midrule
1 & $m$-out-of-$N$ bootstrap & 1.20 & +0.21 & 1.55 \\
1 & Standard bootstrap & 0.95 & +0.46 & 1.62 \\
1 & Beta-binomial EB & 0.32 & +1.10 & 1.75 \\
1 & Sample splitting & 1.35 & +0.06 & 1.64 \\
2 & $m$-out-of-$N$ bootstrap & 0.86 & -0.06 & 1.74 \\
2 & Standard bootstrap & 0.60 & +0.20 & 1.77 \\
2 & Beta-binomial EB & 0.31 & +0.49 & 1.74 \\
2 & Sample splitting & 0.82 & -0.02 & 1.77 \\
3 & $m$-out-of-$N$ bootstrap & 0.78 & -0.12 & 1.78 \\
3 & Standard bootstrap & 0.53 & +0.13 & 1.80 \\
3 & Beta-binomial EB & 0.30 & +0.36 & 1.79 \\
3 & Sample splitting & 0.73 & -0.07 & 1.84 \\
4 & $m$-out-of-$N$ bootstrap & 0.75 & +0.23 & 1.64 \\
4 & Standard bootstrap & 0.57 & +0.41 & 1.71 \\
4 & Beta-binomial EB & 0.29 & +0.69 & 1.72 \\
4 & Sample splitting & 0.82 & +0.16 & 1.73 \\
5 & $m$-out-of-$N$ bootstrap & 0.72 & +0.24 & 1.79 \\
5 & Standard bootstrap & 0.56 & +0.40 & 1.87 \\
5 & Beta-binomial EB & 0.28 & +0.68 & 1.85 \\
5 & Sample splitting & 0.84 & +0.13 & 1.87 \\
\bottomrule
\end{tabular}

\floatnote{Methods use the same 100 synthetic benchmarks with 472 tasks and 173 models. The $m$-out-of-$N$ and standard bootstraps use $m=\lfloor N^{0.9}\rfloor$ and $m=N$, respectively, with 2,000 resamples. EB denotes empirical Bayes and uses the fitted beta-binomial posterior mean. Sample splitting uses 200 random half-splits per benchmark. Mean Score Correction is the mean amount subtracted from the selected model's score. Mean Residual is its mean signed error after correction, and RMSE is the root mean squared error after correction. All 3 measures are in pp and use the selected model's known mean as the reference. At rank~2, the $m$-out-of-$N$ and EB RMSEs tie to the displayed precision.}
\end{table}
Because all methods correct the same selected models on the same draws, their squared errors can be compared benchmark by benchmark. At rank~1, the paired RMSE difference from the $m$-out-of-$N$ correction is 0.07 pp (SE 0.02) for the standard bootstrap, 0.09 pp (SE 0.03) for sample splitting, and 0.20 pp (SE 0.07) for EB (Table~\ref{tab:calib_sim_rmse_diff}). At rank~2, EB and the $m$-out-of-$N$ correction have the same RMSE to the displayed precision (SE of the paired difference 0.07).

At rank~1, RMSE differs by at most 0.20 pp across methods, whereas mean residuals differ by up to 1.03 pp (Table~\ref{tab:alternatives_full}). These results distinguish removing a systematic selection effect from eliminating sampling variation.

\section{Robustness Studies}\label{app:robustness}

The robustness studies distinguish three choices that affect a RULER report. First, we examine the i.i.d. tasks assumption. We try alternative bootstrap methods that relaxes the i.i.d. assumption and see whether the estimated selection effect and ranking uncertainty depend on how tasks are grouped. Second, we change the resample size within the range that Appendix~\ref{app:gamma_guidance} recommends and see whether the estimated inflation is robust to the exponent. Finally, we give different rank mass use for model tiering.

\subsection{Beyond i.i.d. tasks}
\label{app:dependence}
The theory in Section~\ref{sec:model} and the $m$-out-of-$N$ bootstrap treat task rows as i.i.d.\ draws from a common task distribution. Real benchmarks can depart from this assumption in different ways. A benchmark may be assembled with a fixed mix of sources, such as the repositories of SWE-bench, so that a new draw of tasks would keep roughly the same composition. Tasks from the same source may also share features, such as a codebase, that move all models' outcomes on those tasks together. Each pattern corresponds to a resampling scheme. Pooled resampling of task rows matches i.i.d.\ tasks, stratified resampling within repositories matches a fixed repository mix, and cluster resampling matches shared repository effects. If the three schemes give close score adjustments and rank summaries, RULER's conclusions are robust to the i.i.d.\ task assumption.

Pooled resampling is the baseline. It draws $m$ independent rows from the empirical task distribution and uses the factor $\sqrt{m/N}$ in Equation~\ref{eq:bias_estimator}. For stratified resampling, repository $g$ with $n_g$ tasks receives $m_g=\max\{1,\operatorname{round}(n_g m/N)\}$. We sample independently with replacement within each repository, concatenate the sampled rows, and use the actual size $m_{strat}=\sum_g m_g$ in the rescaling. This approximately fixes repository composition while continuing to treat issues within a repository as independent. On Multilingual, rounding produces $m_{strat}=174$ rather than the pooled $m=\lfloor N^{0.9}\rfloor=169$.

For cluster resampling, we use a wild-cluster bootstrap. Define the repository residual vector and the perturbed means
\begin{equation}
 \eta_g=\sum_{i\in g}(\mathbf Y_{i\cdot}-\hat{\boldsymbol\tau}),\qquad
 \hat{\boldsymbol\tau}^{*(b)}=\hat{\boldsymbol\tau}+N^{-1}\sum_g W_g^{(b)}\eta_g.\label{eq:wild}
\end{equation}
Each cluster's multiplier is shared across all its tasks and all models, so a draw shifts every task in a repository together. The Webb multipliers take the six values $\{-\sqrt{3/2},-1,-\sqrt{1/2},\sqrt{1/2},1,\sqrt{3/2}\}$ with equal probability. We rank the perturbed means and use the unscaled selection gap, which corresponds to $m=N$ in Equation~\ref{eq:bias_estimator}. These pseudo-means can lie outside $[0,1]$. The independent-task theorem does not establish the calibration of this procedure, and the small cluster counts---12 repositories for Verified and 3 issue subcategories for Tau2 Telecom---make these comparisons stress tests.

\begin{table}[!t]
\centering\small
\setlength{\tabcolsep}{3pt}
\caption{SWE-bench score adjustments remain stable across alternative sampling schemes.}\label{tab:dependence}
\begin{tabular}{@{}lc*{6}{>{\centering\arraybackslash}p{1.4cm}}@{}}
\toprule
& & \multicolumn{3}{c}{Position 1 Correction (pp)} & \multicolumn{3}{c}{Number of Models in Tier 1} \\
\cmidrule(lr){3-5}\cmidrule(l){6-8}
Benchmark & \shortstack{Number of\\Repositories} & Pooled & Stratified & Cluster & Pooled & Stratified & Cluster \\
\midrule
SWE Bash-only & 12 & 1.09 & 1.21 & 0.86 & 6 & 6 & 4 \\
SWE Verified & 12 & 1.12 & 1.20 & 1.16 & 13 & 12 & 4 \\
SWE Multimodal & 12 & 0.94 & 0.82 & 0.84 & 5 & 5 & 5 \\
SWE Multilingual & 41 & 1.50 & 1.23 & 1.72 & 9 & 8 & 7 \\
SWE Pro & 11 & 0.87 & 0.87 & 0.54 & 4 & 4 & 3 \\
\bottomrule
\end{tabular}

\floatnote{Pooled resamples task rows, Stratified resamples within repositories, and Cluster applies a shared random multiplier to all tasks in each repository. Each method reports the position-1 score correction and the number of models in Tier 1 at 80\% rank mass. All schemes use $B=5{,}000$. Pooled resampling uses $m=\lfloor N^{0.9}\rfloor$. Actual stratified resample sizes are 268 for Bash-only, 268 for Verified, 273 for Multimodal, 174 for Multilingual, and 217 for Pro. Cluster results use shared Webb multipliers and the unscaled selection gap, as specified in Appendix~\ref{app:dependence}.}
\end{table}

Across the 5 SWE-bench variants, pooled and stratified resampling give leading-tier sizes that differ by at most 1 model and score adjustments that differ by at most 0.27 pp (Table~\ref{tab:dependence}). Both summaries are therefore robust to a fixed repository mix. The cluster stress test can change rank summaries more substantially. On Verified, the pooled, stratified, and cluster position-1 adjustments are 1.12, 1.20, and 1.16 pp, respectively, while the leading tier contains 13 models under pooled resampling, 12 under stratified resampling, and 4 under clustering. Table~\ref{tab:dependence_ranks} gives the corresponding rank-1 masses. Shared repository effects thus leave Verified's estimated inflation nearly unchanged but can alter the reported ordering. Tau2 Telecom is more sensitive. Clustering its 3 labeled issue subcategories raises the adjustment from 0.18 to 1.05 pp and lowers the leader's rank-1 mass from 94.3\% to 75.5\%. With only 3 clusters, this result serves as a sensitivity diagnostic.

\begin{table}[!t]
\centering\small
\setlength{\tabcolsep}{4pt}
\caption{Repository clustering changes leading tiers in some SWE-bench variants.}\label{tab:dependence_ranks}
\begin{tabular}{@{}llccccc@{}}
\toprule
& & \multicolumn{3}{c}{Rank-1 Probability (\%)} & & \\
\cmidrule(lr){3-5}
Benchmark & \shortstack[l]{Resampling\\Scheme} & Model 1 & Model 2 & Model 3 & \shortstack{Number\\of Tiers} & \shortstack{Number of\\Models in Tier 1} \\
\midrule
SWE Bash-only & Pooled & 41.6 & 16.8 & 18.4 & 10 & 6 \\
SWE Bash-only & Stratified & 43.1 & 15.5 & 18.1 & 10 & 6 \\
SWE Bash-only & Cluster & 57.4 & 18.1 & 15.9 & 12 & 4 \\
SWE Verified & Pooled & 30.5 & 29.0 & 26.4 & 15 & 13 \\
SWE Verified & Stratified & 29.0 & 29.4 & 26.2 & 15 & 12 \\
SWE Verified & Cluster & 30.6 & 29.1 & 33.8 & 24 & 4 \\
SWE Multimodal & Pooled & 34.8 & 34.0 & 22.0 & 3 & 5 \\
SWE Multimodal & Stratified & 34.8 & 34.3 & 22.2 & 3 & 5 \\
SWE Multimodal & Cluster & 31.8 & 37.5 & 27.0 & 3 & 5 \\
SWE Multilingual & Pooled & 44.8 & 33.6 & 9.1 & 4 & 9 \\
SWE Multilingual & Stratified & 45.5 & 35.2 & 9.0 & 4 & 8 \\
SWE Multilingual & Cluster & 51.9 & 34.1 & 7.9 & 4 & 7 \\
SWE Pro & Pooled & 74.9 & 21.1 & 1.9 & 6 & 4 \\
SWE Pro & Stratified & 74.7 & 21.3 & 1.8 & 6 & 4 \\
SWE Pro & Cluster & 87.6 & 10.8 & 0.3 & 6 & 3 \\
\bottomrule
\end{tabular}

\floatnote{Models 1--3 are the first 3 models in the original displayed order, including any observed score ties. Each probability column gives the percentage of bootstrap rank mass assigned to rank 1 for that model. Pooled, Stratified, and Cluster use the sampling schemes and resample sizes defined in Table~\ref{tab:dependence}, with $B=5{,}000$. The last 2 columns report the total number of tiers and the number of models in Tier 1 under the 80\% common-intersection rule.}
\end{table}

\subsection{Choice of resample size}\label{app:mn_exponent_sensitivity}
Appendix~\ref{app:gamma_guidance} shows that, averaged over all leaderboard positions, exponents from 0.4 to 0.9 are nearly equally accurate when models are closely spaced. We therefore recompute the position-1 adjustment of every audited benchmark at $\gamma\in\{0.4,0.5,\ldots,0.9\}$, using $m=\max\{2,\lfloor N^\gamma\rfloor\}$, $B=5{,}000$, and seed 42. Table~\ref{tab:gamma_bias} reports the 84 adjustments, and Table~\ref{tab:gamma_m} gives their exact resample sizes. The default $\gamma=0.9$ was fixed before inspecting outcomes.

Across this range, the correction keeps its direction, and the default is conservative. Every adjustment is positive, and in each of the 14 benchmarks the default gives the smallest adjustment in the range. A smaller exponent in the range would therefore estimate more inflation at the top of each leaderboard than the audit reports. In 12 benchmarks, including the 6 whose 10 leading models lie within 3 standard errors, the adjustment also keeps its order of magnitude across the range. Tau2 Telecom and pooled LiveBench show by far the largest relative changes.

These 2 benchmarks have well-separated leaders, which receive 94.3\% and 99.0\% of the bootstrap rank-1 mass at the default, and their small default adjustments grow several-fold at smaller exponents. For a well-separated leader, the spread design in Appendix~\ref{app:gamma_guidance} shows that smaller exponents are less accurate, and the two-model example in Appendix~\ref{app:toy} shows that they overcorrect. The simulations therefore favor the small default adjustments for these 2 leaders.

\begin{table}[!t]\centering\small
\setlength{\tabcolsep}{3pt}
\caption{Score adjustments stay positive across resample-size exponents from 0.4 to 0.9, and the default gives the smallest.}\label{tab:gamma_bias}
\begin{tabular}{@{}l*{6}{>{\centering\arraybackslash}p{0.7cm}}@{}}
\toprule
& \multicolumn{6}{c}{Resample-Size Exponent $\gamma$} \\
\cmidrule(l){2-7}
Benchmark & 0.4 & 0.5 & 0.6 & 0.7 & 0.8 & 0.9 \\
\midrule
SWE Bash-only & 1.61 & 1.59 & 1.53 & 1.38 & 1.26 & 1.09 \\
SWE Verified & 1.91 & 1.90 & 1.77 & 1.56 & 1.34 & 1.12 \\
SWE Multimodal & 1.13 & 1.13 & 1.13 & 1.04 & 1.02 & 0.94 \\
SWE Multilingual & 2.10 & 2.16 & 2.15 & 1.94 & 1.74 & 1.50 \\
SWE Pro & 2.10 & 1.90 & 1.67 & 1.43 & 1.16 & 0.87 \\
Tau2 Airline & 6.13 & 5.75 & 5.40 & 4.80 & 4.00 & 3.07 \\
Tau2 Retail & 3.01 & 2.93 & 2.78 & 2.60 & 2.45 & 2.20 \\
Tau2 Telecom & 1.23 & 1.23 & 0.93 & 0.58 & 0.36 & 0.18 \\
AlpacaEval 2 & 1.33 & 1.13 & 0.93 & 0.77 & 0.64 & 0.53 \\
WildBench v2 & 0.35 & 0.30 & 0.24 & 0.17 & 0.13 & 0.10 \\
LiveBench pooled & 2.58 & 2.19 & 1.65 & 0.96 & 0.39 & 0.12 \\
LiveBench Coding & 4.12 & 3.42 & 2.77 & 2.13 & 1.53 & 1.11 \\
LiveBench Instr. & 4.02 & 4.09 & 4.15 & 4.17 & 4.09 & 3.86 \\
LiveBench Language & 4.07 & 3.62 & 3.03 & 2.53 & 2.09 & 1.90 \\
\bottomrule
\end{tabular}

\floatnote{Entries are position-1 adjustments in pp for 14 benchmarks and 6 exponents. Each run uses $m=\max\{2,\lfloor N^\gamma\rfloor\}$, $B=5{,}000$, and seed 42. Score ties follow the deterministic sort convention in Appendix~\ref{app:implementation}. The default exponent is $\gamma=0.9$.}
\end{table}
\begin{table}[!t]\centering\small
\setlength{\tabcolsep}{3pt}
\caption{Resample sizes range from 4 to 495 tasks across exponents from 0.4 to 0.9.}\label{tab:gamma_m}
\begin{tabular}{@{}l*{6}{>{\centering\arraybackslash}p{0.7cm}}@{}}
\toprule
& \multicolumn{6}{c}{Resample-Size Exponent $\gamma$} \\
\cmidrule(l){2-7}
Benchmark & 0.4 & 0.5 & 0.6 & 0.7 & 0.8 & 0.9 \\
\midrule
SWE Bash-only & 12 & 22 & 41 & 77 & 144 & 268 \\
SWE Verified & 12 & 22 & 41 & 77 & 144 & 268 \\
SWE Multimodal & 12 & 22 & 42 & 78 & 146 & 273 \\
SWE Multilingual & 9 & 17 & 30 & 54 & 95 & 169 \\
SWE Pro & 10 & 19 & 36 & 65 & 119 & 217 \\
Tau2 Airline & 4 & 7 & 10 & 15 & 22 & 33 \\
Tau2 Retail & 6 & 10 & 17 & 27 & 44 & 70 \\
Tau2 Telecom & 6 & 10 & 17 & 27 & 44 & 70 \\
AlpacaEval 2 & 14 & 28 & 55 & 108 & 211 & 412 \\
WildBench v2 & 15 & 31 & 62 & 124 & 248 & 495 \\
LiveBench pooled & 10 & 17 & 31 & 56 & 100 & 178 \\
LiveBench Coding & 6 & 11 & 18 & 29 & 48 & 78 \\
LiveBench Instr. & 4 & 7 & 10 & 15 & 22 & 33 \\
LiveBench Language & 7 & 11 & 19 & 31 & 52 & 85 \\
\bottomrule
\end{tabular}

\floatnote{Entries are the exact resample sizes $m=\max\{2,\lfloor N^\gamma\rfloor\}$ for the runs in Table~\ref{tab:gamma_bias}. Column headings give $\gamma$, and the default exponent is 0.9.}
\end{table}

\subsection{Choice of rank-interval level}\label{app:alpha}
Retaining more rank mass gives a broader summary of the unresolved ordering. On Verified, increasing central rank mass from 80\% to 90\%, 95\%, and 99\% expands the leading tier from 13 models to 15, 24, and 29, respectively. Table~\ref{tab:alpha} reports the corresponding partitions in every benchmark. The underlying RPM is unchanged. The difference is how much of each model's rank distribution the interval includes.

\begin{table}[!t]\centering\small
\caption{Leading tiers often expand when the retained rank mass increases.}\label{tab:alpha}
\begin{tabular}{@{}l*{8}{>{\centering\arraybackslash}p{1.1cm}}@{}}
\toprule
& \multicolumn{4}{c}{Number of Tiers} & \multicolumn{4}{c}{Number of Models in Tier 1} \\
\cmidrule(lr){2-5}\cmidrule(l){6-9}
Benchmark & 80\% & 90\% & 95\% & 99\% & 80\% & 90\% & 95\% & 99\% \\
\midrule
SWE Bash-only & 10 & 9 & 7 & 6 & 6 & 9 & 13 & 16 \\
SWE Verified & 15 & 13 & 11 & 8 & 13 & 15 & 24 & 29 \\
SWE Multimodal & 3 & 3 & 2 & 2 & 5 & 5 & 8 & 10 \\
SWE Multilingual & 4 & 3 & 3 & 3 & 9 & 11 & 12 & 12 \\
SWE Pro & 6 & 6 & 4 & 4 & 4 & 4 & 5 & 5 \\
Tau2 Airline & 2 & 1 & 1 & 1 & 5 & 7 & 7 & 7 \\
Tau2 Retail & 2 & 2 & 2 & 2 & 5 & 6 & 6 & 6 \\
Tau2 Telecom & 6 & 4 & 3 & 2 & 1 & 2 & 3 & 3 \\
AlpacaEval 2 & 21 & 16 & 14 & 11 & 3 & 4 & 4 & 5 \\
WildBench v2 & 15 & 12 & 12 & 10 & 2 & 3 & 3 & 6 \\
LiveBench pooled & 9 & 7 & 7 & 5 & 1 & 1 & 1 & 6 \\
LiveBench Coding & 8 & 6 & 5 & 4 & 2 & 3 & 5 & 9 \\
LiveBench Language & 7 & 6 & 5 & 4 & 4 & 5 & 7 & 15 \\
LiveBench Instr. & 5 & 4 & 4 & 3 & 19 & 26 & 30 & 42 \\
\bottomrule
\end{tabular}

\floatnote{The 2 column groups separately report the total number of tiers and the number of models in Tier 1. Within each group, the percentage headings specify the central rank mass retained in each model's interval. Partitions are recomputed from the same full-precision RPMs, using $m=\lfloor N^{0.9}\rfloor$, $B=5{,}000$, and the common-intersection rule. Only the retained rank mass changes across columns.}
\end{table}
Wider intervals can change where the common-intersection sweep closes a tier, so the resulting partitions need not form a nested hierarchy. We therefore recompute each level from the full-precision RPM using the same reporting conventions (Appendix~\ref{app:implementation}).

\section{Benchmark Planning and Supplementary Reports}\label{app:tables}
The score adjustment provides a scale for planning larger task pools, and the full RPM reveals how ranking uncertainty extends beyond the models highlighted in the main text. This appendix develops the planning calculation and provides larger rank-distribution displays for Verified and Tau2-Bench.

\subsection{Task-pool sizing for a target level of score inflation}\label{app:sizing_rule}
More tasks reduce the expected selection advantage of the winner. In the exact-tier regime, $b_N(1)\approx\delta/\sqrt N$, so halving expected inflation requires approximately 4 times as many tasks. For a target expected bias $\varepsilon$, the corresponding planning rule is $N\gtrsim(\delta/\varepsilon)^2$.

For Verified, the position-1 adjustment gives $\hat\delta=\sqrt{500}\,\hat b_{500}(1)\approx25.1$ pp$\sqrt{\text{tasks}}$. Table~\ref{tab:sizing} translates this estimate into task counts for different targets. The calculation holds the model pool, task distribution, and covariance structure fixed and uses the exact-tier approximation. Local gaps or changes in these populations can alter the extrapolation, so the table gives approximate planning sizes rather than guaranteed minima.
\begin{table}[!t]\centering\small
\caption{Under the exact-tier approximation, halving expected winner inflation requires about 4 times as many tasks.}\label{tab:sizing}
\begin{tabular}{@{}cc@{}}
\toprule
\shortstack{Target Expected\\Winner Bias (pp)} & \shortstack{Approximate Number\\of Tasks} \\
\midrule
2.00 & 158 \\
1.00 & 631 \\
0.50 & 2,522 \\
0.25 & 10,086 \\
0.10 & 63,035 \\
\bottomrule
\end{tabular}

\floatnote{Planning sizes use $N\gtrsim(\hat\delta/\varepsilon)^2$ with $\hat\delta=\sqrt{500}\,\hat b_{500}(1)\approx25.1$ pp$\sqrt{\text{tasks}}$ from Verified's position-1 estimate. Counts are rounded upward using full precision. The exact-tier approximation holds the model pool, task distribution, and covariance structure fixed. Values are approximate planning sizes.}
\end{table}

\subsection{Additional rank probability matrices}\label{app:rpm_full}

The larger RPM places Verified's leading tier in the context of the first 30 models and rank positions. Uncertainty extends beyond the observed leaders: the models tied at raw positions 9--10 each have an 80\% rank interval of [5,21]. The display shows where rank mass is concentrated and how far it spreads across the leaderboard.
\begin{figure}[!t]\centering
\includegraphics[width=.7\textwidth]{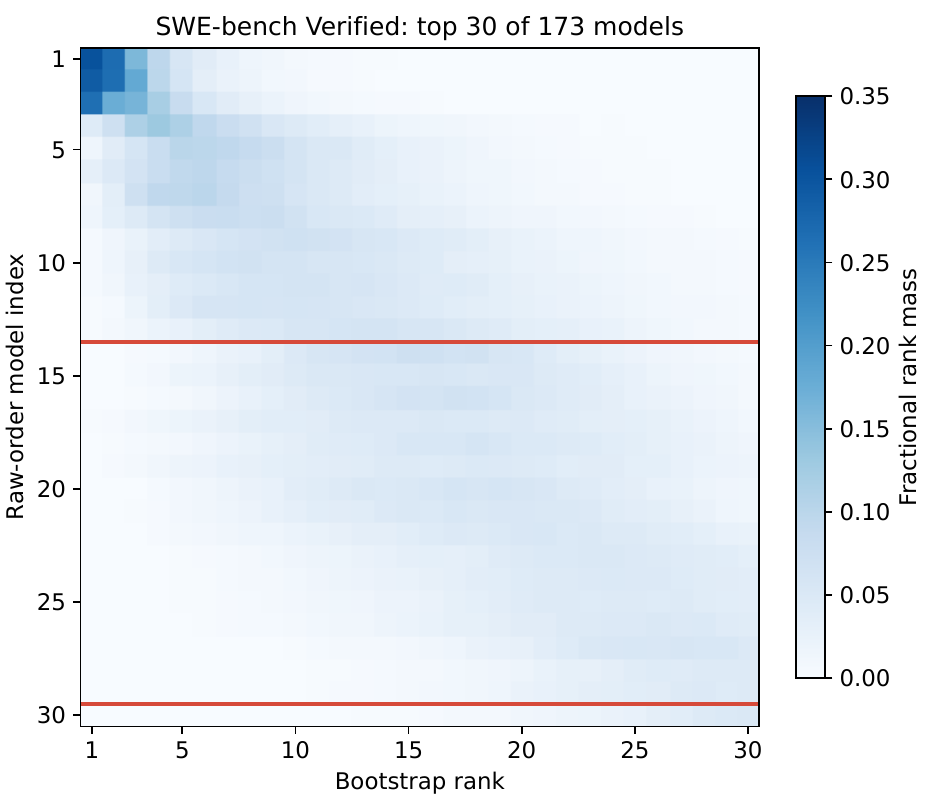}
\caption{Ranking uncertainty extends beyond SWE-bench Verified's leading models.}\label{fig:rpm_full}

\floatnote{The display is the leading $30\times30$ submatrix of the full $173\times173$ RPM, based on 500 tasks, $m=\lfloor N^{0.9}\rfloor$, and $B=5{,}000$. Red horizontal lines mark 80\% tier boundaries in raw model order. Cropped rows need not sum to one, while full RPM rows do. Axis indices identify raw-order model rows without imposing an order within score ties.}
\end{figure}

\begin{figure}[!htb]\centering
\includegraphics[width=\textwidth]{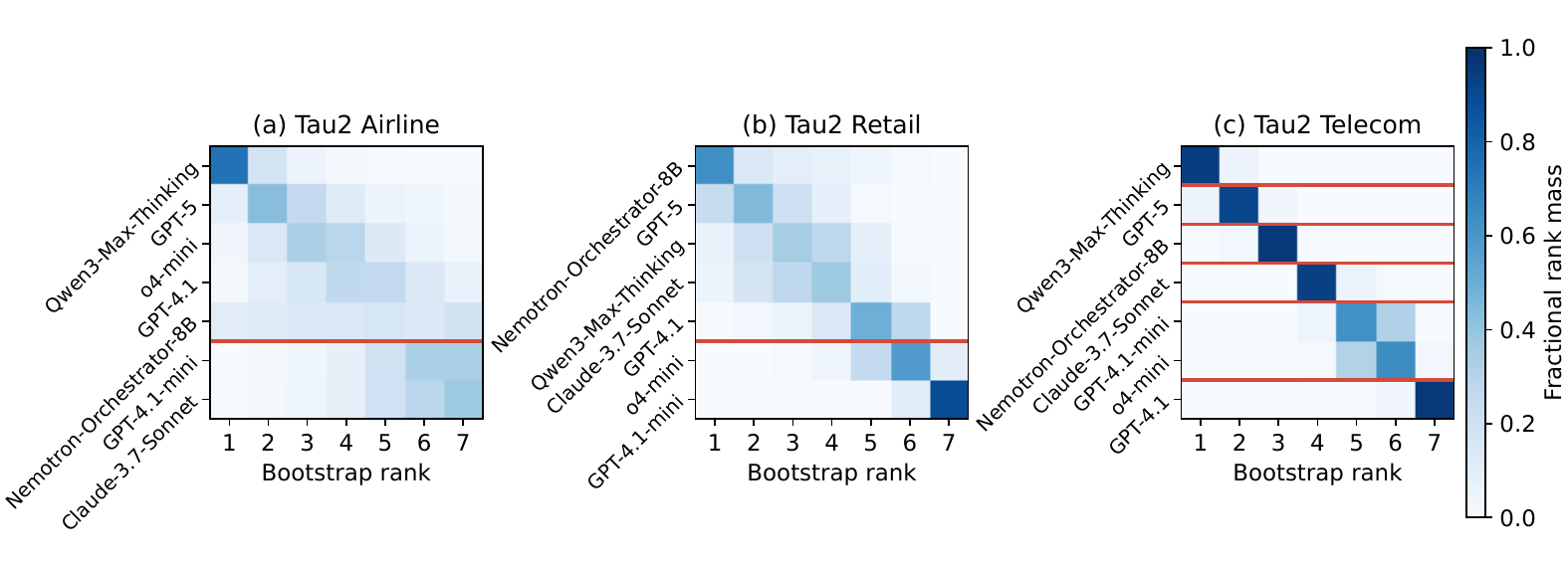}
\caption{Tau2 domains differ in how sharply they resolve the leading tier.}\label{fig:tau2_rpm}

\floatnote{The RPMs use pooled task resampling with $J=7$, $m=\lfloor N^{0.9}\rfloor$, and $B=5{,}000$, with $N=50$ tasks for Airline and $N=114$ for Retail and Telecom. Red lines mark 80\% tier boundaries from the common-intersection rule. In panels (a) and (b), 5 of 7 models share the leading tier. In panel (c), the leading tier contains only the leader. Section~\ref{subsec:dependence} reports Telecom's sensitivity to issue-subcategory clustering.}
\end{figure}

The Tau2 domains illustrate different degrees of ranking resolution. Airline and Retail each place 5 of 7 models in the leading tier, and Telecom's leader forms a tier of its own under pooled resampling. Figure~\ref{fig:tau2_rpm} shows the RPMs behind those tier summaries. Red lines mark the 80\% tier boundaries. Airline has only $N=50$ tasks, and each resample draws $m=\lfloor 50^{0.9}\rfloor=33$ of them. Telecom's sensitivity to clustering is discussed in Section~\ref{subsec:dependence}.
\section{Extended Related Work}\label{app:related}

\paragraph{Selection, comparison, and rank uncertainty.}
Rank-induced bias connects leaderboard interpretation to the broader problem of selection. The WC was documented in oil-lease auctions \citep{capen1971competitive,thaler1988anomalies} and formalized as the optimizer's curse \citep{smith2006optimizer}. Selective inference studies inference after selection \citep{lee2016exact,andrews2024inference}, while empirical Bayes uses information across estimates for shrinkage \citep{efron2011tweedie,gu2023invidious}. RULER builds on classical $m$-out-of-$N$ resampling \citep{bickel1997resampling,bickelsakov2008} and its use for selected-winner correction \citep{xu2025winners}. It extends the correction to every leaderboard position while accounting for cross-model dependence within a task. The RPM and rank intervals complement the score correction by showing how much of the ordering the task pool resolves.

Statistical comparison predates LLM leaderboards. \citet{koehn-2004-statistical} studies significance testing for machine-translation evaluation, and \citet{mathur-etal-2020-tangled} examines how automatic evaluation metrics, outliers, and system comparisons affect conclusions about translation quality. Our target is the mean estimation error of the model \emph{selected by its position after ranking}. Rank confidence sets \citep{mogstad2024inference,mogstad2022ranking} and classifier-comparison procedures \citep{dietterich1998approximate,demsar2006statistical,benavoli2017time} address inferential comparisons. RULER complements these approaches by estimating the score error associated with selection and displaying the resampled ordering. Its rank-mass intervals summarize that distribution rather than provide simultaneous confidence coverage.

\paragraph{Validity and reproducibility.}
\citet{raji2021everything} questions the scope and meaning of general-purpose AI benchmarks. \citet{belz-etal-2022-quantified} develops quantified reproducibility assessment, extended across common result types by \citet{belz2026quantified}. \citet{schmidtova-etal-2024-automatic} surveys automatic metric practices in natural-language generation. These concerns extend beyond task sampling: a precise score can measure the wrong capability, and implementations can vary independently of the task pool (Table~\ref{tab:related_distinctions}).

\begin{table}[!t]
\centering\small
\caption{Selection bias is distinct from statistical comparison, metric validity, and reproducibility.}\label{tab:related_distinctions}
\begin{tabular}{@{}>{\raggedright\arraybackslash}p{3.1cm}>{\raggedright\arraybackslash}p{5.0cm}>{\raggedright\arraybackslash}p{5.0cm}@{}}\toprule
Concern & Question & Relationship to this paper \\\midrule
Statistical comparison & Does a prespecified difference exceed evaluation noise? & Selection after sorting changes which error is averaged. \\
Selection bias & How much does a chosen position benefit from noise? & Our bias target. RPMs additionally describe ranking variability. \\
Construct validity & Does the task pool measure the intended capability? & Our target is conditional on the chosen task population. \\
Reproducibility & How much do independently repeated implementations or studies agree? & Frozen outcomes do not capture all reproduction variability. \\
Metric validity & Does the score agree with the quality notion of interest? & Debiasing a score does not validate its underlying metric. \\\bottomrule
\end{tabular}
\end{table}

\paragraph{Other leaderboard failure modes.}
Leaderboards can also become unreliable through changes in the evaluation process or the models submitted to it. Adaptive holdout reuse \citep{blum2015ladder,dwork2015generalization,roelofs2019meta} and contamination \citep{deng2024investigating} can compromise an evaluation's independence from model development. Private evaluation suites \citep{liang2022holistic} can reduce leakage and adaptation. Benchmark and prompt sensitivity \citep{dehghani2021benchmark,alzahrani2024benchmarks}, run-to-run variance \citep{bouthillier2021accounting}, and efficient or uncertainty-aware evaluation \citep{polo2024efficient,ye2024benchmarking,chiang2024chatbot,ameli2025a} concern further sources of variability and the design of evaluations. Rank-induced bias remains even for a fixed, uncontaminated model pool evaluated once on held-out tasks, because sorting selects favorable noise along with ability.

\end{document}